\pdfoutput=1
\documentclass{these-dbl}
\hypersetup{
   pdfauthor   = {Maxime Bernard},
   pdftitle    = {Th\`{e}se de doctorat de Maxime Bernard},
   pdfsubject  = {Th\`{e}se de doctorat de Maxime Bernard},
   pdfkeywords = {Multi-robot Systems, Distributed Algorithms, Coordination, Graph Theory, Multi-agent Systems, Haptic Teleoperation, Human-Robot Interaction, Haptic},
}

\def\CC{{C\nolinebreak[4]\hspace{-.05em}\raisebox{.4ex}{\tiny\bf ++}}}
\newcommand{\laplacian}{\boldsymbol{L}}
\newcommand{\norm}[1]{\left\lVert#1\right\rVert}
\newcommand{\agentPosj}{\boldsymbol{p}_j}
\newcommand{\agentPos}{\boldsymbol{p}}
\newcommand{\agentYaw}[1]{\psi_{#1}}
\newcommand{\RtoAgent}[1]{\R^T_{#1}}
\newcommand{\RfromAgentj}{\R_j}
\newcommand{\idlePos}{\boldsymbol{p}^0_V}
\newcommand{\deadzone}{\boldsymbol{p}^{min}_V}
\newcommand{\workspaceRange}{\boldsymbol{p}^{max}_V}
\newcommand{\joystickForce}{\boldsymbol{F}_V}
\newcommand{\centeringGain}{k_c}
\newcommand{\deadzoneGain}{k_d}
\newcommand{\limitGain}{k_l}
\newcommand{\centeringForce}{\boldsymbol{F}_C}
\newcommand{\deadzoneForce}{\boldsymbol{F}_D}
\newcommand{\limitForce}{\boldsymbol{F}_L}
\newcommand{\virtuosePos}{\boldsymbol{p}_V}
\newcommand{\virtuoseInput}{\widetilde{\boldsymbol{p}}_V}
\newcommand{\virtuoseInputSub}[1]{V_V^{#1}}
\newcommand{\idleRot}{\boldsymbol{q}^0_V}
\newcommand{\deadzoneR}{\boldsymbol{\eta}^{min}_V}
\newcommand{\workspaceRangeR}{\boldsymbol{\eta}^{max}_V}
\newcommand{\joystickForceR}{\boldsymbol{T}_V}
\newcommand{\centeringGainR}{k_{cr}}
\newcommand{\deadzoneGainR}{k_{dr}}
\newcommand{\limitGainR}{k_{lr}}
\newcommand{\centeringForceR}{\boldsymbol{T}_C}
\newcommand{\deadzoneForceR}{\boldsymbol{T}_D}
\newcommand{\limitForceR}{\boldsymbol{T}_L}
\newcommand{\virtuoseRot}{\boldsymbol{q}_V}
\newcommand{\virtuoseInputR}{\widetilde{\boldsymbol{\eta}}_V}
\newcommand{\virtuoseInputRoll}{\widetilde{\boldsymbol{\eta}}^{\phi}}
\newcommand{\virtuoseInputPitch}{\widetilde{\boldsymbol{\eta}}^{\theta}}
\newcommand{\virtuoseInputYaw}{\widetilde{\boldsymbol{\eta}}^{\psi}}
\newcommand{\pinningPointVelocity}{\vv_p}
\newcommand{\pinningPointYawrate}{\omega_p}
\newcommand{\pinningPointPos}{\boldsymbol{p}_p}
\newcommand{\dotpinningPointPos}{\dot{\boldsymbol{p}}_p}
\newcommand{\pinningPointYaw}{\psi_p}
\newcommand{\dotpinningPointYaw}{\dot{\psi}_p}
\newcommand{\pinningVect}{\boldsymbol{p}_{jp}}
\newcommand{\pinningUnitVect}{\boldsymbol{u}_{jp}}
\newcommand{\centroidVectUnitSub}[1]{\centroidVectUnit^{#1}}
\newcommand{\pinningMinDistj}{d^{min}_{jp}}
\newcommand{\centroidPos}{\boldsymbol{p}_c}
\newcommand{\centroidYaw}{\psi_c}
\newcommand{\centroidUserInputs}{\boldsymbol{\Gamma}}
\newcommand{\centroidUserInputsSub}[1]{\Gamma_{#1}}
\newcommand{\centroidMappingMatrix}{\boldsymbol{H}_c}
\newcommand{\centroidVect}{\boldsymbol{p}_{jc}}
\newcommand{\centroidVectUnit}{\boldsymbol{u}_{jc}}
\newcommand{\centroidVectSub}[1]{\centroidVect^{#1}}
\newcommand{\centroidControlMatrix}{\boldsymbol{Q}}
\newcommand{\centroidControlMatrixj}{\boldsymbol{Q}_j}
\newcommand{\connectivityForceParallell}{\boldsymbol{\tau}_{\parallel}}
\newcommand{\connectivityForceNullSpace}{\boldsymbol{\tau}_{\perp}}
\newcommand{\forceFeedback}{\F_{F}}
\newcommand{\torqueFeedback}{\T_{F}}
\newcommand{\maxForceFeedback}{\F^{max}_{F}}
\newcommand{\maxTorqueFeedback}{\T^{max}_{F}}
\newcommand{\connectivityVelocityj}{\vv^{\lambda}_j}
\newcommand{\connectivityYawRatej}{\omega^{\lambda}_j}
\newcommand{\residualconnectivityVelocityj}{\widetilde{\vv}^{\lambda}_j}
\newcommand{\residualconnectivityYawRatej}{\widetilde{\omega}^{\lambda}_j}
\newcommand{\residualconnectivityVelocity}{\widetilde{\vv}^{\lambda}}
\newcommand{\residualconnectivityYawRate}{\widetilde{\omega}^{\lambda}}
\newcommand{\connectivityVelocity}{\vv^{\lambda}}
\newcommand{\connectivityYawRate}{\omega^{\lambda}}
\newcommand{\connectivityForce}{\F^{\lambda}}
\newcommand{\connectivityForcej}{\F^{\lambda}_j}
\newcommand{\connectivityEigenvector}{\mathbf{v}_2}
\newcommand{\connectivityEigenvectori}{\mathrm{v}_{2_i}}
\newcommand{\connectivityEigenvectorj}{\mathrm{v}_{2_j}}
\newcommand{\controlledAgents}{\calC_u}
\newcommand{\externalInputLinear}{\boldsymbol{v}^E_j}
\newcommand{\externalInputAngular}{\omega^E_j}
\newcommand{\externalInputLinearAll}{\boldsymbol{v}^E}
\newcommand{\externalInputAngularAll}{\bomega^E}
\newcommand{\calbR}{\mathbb{R}}
\newcommand{\vv}{\boldsymbol{v}}
\newcommand{\A}{\boldsymbol{A}}
\newcommand{\R}{\boldsymbol{R}}
\newcommand{\T}{\boldsymbol{T}}
\newcommand{\bbeta}{\boldsymbol{\beta}}
\newcommand{\oo}{\boldsymbol{o}}
\newcommand{\x}{\boldsymbol{x}}

\newcommand{\e}{\boldsymbol{e}}
\renewcommand{\P}{\boldsymbol{P}}
\newcommand{\I}{\boldsymbol{I}}
\newcommand{\p}{\boldsymbol{p}}
\newcommand{\0}{\boldsymbol{0}}
\newcommand{\bomega}{\boldsymbol{\omega}}
\renewcommand{\S}{\boldsymbol{S}}
\newcommand{\Z}{\boldsymbol{Z}}
\newcommand{\calG}{\mathcal{G}}
\newcommand{\calV}{\mathcal{V}}
\newcommand{\calE}{\mathcal{E}}
\newcommand{\calN}{\mathcal{N}}
\newcommand{\calC}{\mathcal{C}}
\newcommand{\diag}{\mathrm{diag}}
\newcommand{\bO}{\boldsymbol{O}}
\newcommand{\X}{\boldsymbol{X}}
\newcommand{\Y}{\boldsymbol{Y}}
\newcommand{\F}{\boldsymbol{F}}
\DeclareMathOperator*{\argmax}{\arg\!\max}
\makeatletter
\let\l@sstar\l@section
\newcommand{\addstarredsection}[1]{%
  \begingroup\let\addcontentsline\@gobblethree\addst@rred{section}{#1}\endgroup
  \addtocontents{toc}{\protect\contentsline{sstar}{#1}{\thepage}{}}}
\makeatother

\theoremstyle{definition}
\newtheorem{remark}{Remark}

\begin{document}
\dominitoc
\selectlanguage{french}

\direcole{MATISSE}
\numeroecole{601}
\definecolor{couleur-ecole-recto}{RGB}{0,112,237}
\definecolor{couleur-ecole-verso}{RGB}{0,84,160}
\nomecoleA{Math\'{e}matiques, T\'{e}l\'{e}communications, Informatique, Signal, Syst\`{e}mes,}
\nomecoleB{\'{E}lectronique}
\hauteurlogoecole{1.8cm}
\hauteurlogoetablissementA{1.8cm}
\hauteurlogoetablissementB{1.8cm}
\couleurpolice{white}
\logoetablissementB{UR}
\nometablissementE{l'Universit\'{e} de Rennes}

\spec{AST : Automatique, Productique et Robotique - Signal,\\Image, Vision - Télécommunications}

\author{Maxime BERNARD}

\title{Haptic Shared Control for Multi-Robot Systems}

\date{30 avril 2025}
\lieu{Rennes}

\uniterecherche{IRISA UMR6074}

\jury{
{\normalTwelve \textbf{Rapporteurs avant soutenance :}}\\ \newline
\footnotesizeTwelve
\begin{tabular}{@{}ll}
Fabio MORBIDI & Maître de conférence, Université de Picardie Jules Verne \\
Francesco CHINELLO & Professeur associé, Université d'Aarhus \\
\end{tabular}

\vspace{\baselineskip}
{\normalTwelve \textbf{Composition du Jury :}}\\
\footnotesizeTwelve
\begin{tabular}{@{}lll}

Pr\'{e}sidente :        & Maud MARCHAL & Professeure, INSA/IRISA Rennes \\
Examinateurs :         & Maud MARCHAL & Professeure, INSA/IRISA Rennes \\
                       & Olivier SIMONIN & Professeur, INSA/Inria Lyon \\
Rapporteurs:           & Fabio MORBIDI & Maître de Conférence, Université de Picardie Jules Verne \\
                       & Francesco CHINELLO & Professeur associé, Université d'Aarhus  \\
Dir. de th\`{e}se :    & Paolo ROBUFFO GIORDANO & Directeur de recherche CNRS, IRISA/Inria Rennes \\
Co-encadrant :         & Claudio PACCHIEROTTI & Charg\'{e} de recherche CNRS, IRISA/Inria Rennes \\
\end{tabular}

\vspace{\baselineskip}
\vspace{\baselineskip}
\vspace{\baselineskip}
\vspace{\baselineskip}
\vspace{\baselineskip}
\vspace{\baselineskip}
\vspace{\baselineskip}
\vspace{\baselineskip}
}

\maketitle

\selectlanguage{english}

\clearemptydoublepage
\chapter*{Acknowledgement}

Faire une thèse, c'est à la fois une aventure scientifique et une expérience humaine intense, et elle n'aurait clairement pas été possible sans l'aide, le soutien et la présence de beaucoup de personnes. Je tiens ici à remercier tous ceux qui ont contribué, de près ou de loin, à l'aboutissement de ce travail.

Je remercie bien sûr mes directeurs de thèse, Paolo Robuffo Giordano et Claudio Pacchierotti, pour leur accompagnement tout au long de ces années, pour leurs conseils, leur disponibilité et la confiance qu'ils m'ont accordée.

Je remercie les rapporteurs de cette thèse, Fabio Morbidi et Francesco Chinello, pour le temps consacré à la relecture attentive du manuscrit et pour leurs remarques constructives, ainsi que les membres du jury, Maud Marchal et Olivier Simonin, pour avoir accepté d'examiner ce travail et de participer à ma soutenance.

Ces années de doctorat n'auraient pas été les mêmes sans l'environnement de travail offert par l'équipe Rainbow. Merci à toutes celles et ceux qui en font partie pour l'accueil, les échanges scientifiques, mais aussi pour les moments plus informels qui ont largement contribué à rendre cette expérience agréable. Un merci tout particulier aux doctorants et post-doctorants avec qui j'ai partagé le quotidien du labo, les discussions, les galères, mais aussi les réussites et les moments de détente.

Merci à mes parents pour leur soutien et leurs encouragements, sans lesquels rien n'aurait été possible. Merci à mes frères pour leur présence, même à distance, et merci à ma tante pour son soutien indéfectible.

Enfin, un immense merci à mes amis pour leur présence, vous avez été essentiels pour garder l'équilibre, surtout dans les périodes plus difficiles.
Aux amis de Rennes, vous rencontrer a été un des plus beaux cadeaux de cette aventure.
Aux amis de longue date, quitter momentanément Rennes pour vous retrouver a toujours été un bol d'air et une source de motivation.

À toutes celles et ceux qui ont, d'une manière ou d'une autre, contribué à rendre ce parcours plus humain et plus agréable : merci sincèrement.

\frontmatter
\clearemptydoublepage
\renewcommand{\contentsname}{Table of Contents}
{\hypersetup{linkcolor=black}
\tableofcontents
}

\clearemptydoublepage
\chapter*{Résumé étendu en Français}
\section*{Contexte et problématique}

Les systèmes multi-robots offrent des avantages significatifs en termes d'efficacité, de robustesse et de scalabilité pour des applications telles que les missions de recherche et sauvetage, la surveillance ou l'exploration d'environnements dangereux. Cependant, le contrôle intuitif d'une flotte de robots par un opérateur humain unique pose des défis majeurs. Les algorithmes de maintenance de connectivité décentralisée permettent de garantir la communication inter-agents tout en distribuant le contrôle, mais les travaux antérieurs reposent sur des hypothèses simplificatrices : capteurs omnidirectionnels et référentiel commun partagé par tous les robots. Cette thèse vise à lever ces limitations en proposant des méthodes de commande partagée avec retour haptique adaptées à des contraintes réalistes de capteurs embarqués.

\section*{Objectifs et contributions}

Cette thèse poursuit trois objectifs principaux :
\begin{itemize}
    \item étendre l'algorithme de maintenance de connectivité décentralisée pour intégrer des capteurs à champ de vision limité (FOV) et des mesures exprimées dans le repère propre de chaque robot,
    \item développer une méthode d'estimation décentralisée de l'orientation relative entre robots,
    \item proposer plusieurs architectures de contrôle partagé à échelle variable avec retour haptique informatif. Les contributions sont validées par des simulations réalistes et une étude utilisateur impliquant 20 participants.
\end{itemize}

\section*{Méthodologie}
La méthodologie repose sur la théorie des graphes algébriques pour modéliser les interactions inter-agents. L'algorithme de maintenance de connectivité consiste en une descente de gradient sur une fonction de coût basée sur $\lambda_2$, la deuxième plus petite valeur propre de la matrice laplacienne du graphe. Cette valeur caractérise la connectivité globale: $\lambda_2 > 0$ garantit un graphe connecté.

\subsection*{Extension pour contraintes de champ de vision (Chapitre 4)}
Les poids d'adjacence sont conçus comme le produit de deux composantes : $\gamma_{ij}$ pour les contraintes de distance (activation progressive entre distances minimale et maximale) et $f_{ij}$ pour le FOV. Le poids $f_{ij}$ est symétrisé de manière à garantir qu'au moins un robot voit l'autre, évitant ainsi des contraintes excessives de visibilité mutuelle. Une méthode novatrice d'estimation décentralisée de l'orientation relative $\psi_{ij}$ est développée. Elle utilise un observateur de perturbation pour estimer la vitesse du voisin dans le repère local, combiné avec l'odométrie et, lorsque disponible, une mesure directe en visibilité mutuelle. Les simulations avec six quadrotors démontrent le maintien de $\lambda_2$ au-dessus du seuil minimal malgré une topologie dynamique et des erreurs d'estimation limitées.

\subsection*{Commande partagée avec retour haptique (Chapitre 5)}
Trois méthodes de contrôle à échelle variable sont proposées. Le contrôle d'agent unique (S) permet à l'opérateur de piloter directement un robot leader, les autres suivant via la maintenance de connectivité. Le contrôle par ancrage (P) introduit un point virtuel contrôlé par l'opérateur, qui exerce une force élastique sur plusieurs robots sélectionnés. Le contrôle par barycentre (C) permet de piloter le centre de masse du groupe avec sept degrés de liberté : translation, rotation, et mise à l'échelle de la formation.

L'interface haptique utilise le dispositif Virtuose 6D (Haption) fournissant retour de force et couple sur six degrés de liberté. Des forces de type joystick (deadzone, centrage, limites d'espace) sont combinées avec un retour haptique proportionnel aux forces de connectivité, informant l'opérateur de l'état du système. Un mécanisme de sélection automatique du leader identifie le robot ayant la plus grande manœuvrabilité en lacet, basé sur la visibilité par ses voisins.

L'implémentation logicielle repose sur une architecture ROS 2 décentralisée, où chaque robot opère comme un processus indépendant, simulant fidèlement le déploiement réel. La visualisation Unity 3D communique via pont TCP pour afficher l'état du système.

\section*{Résultats principaux}
L'étude utilisateur compare les trois méthodes de contrôle avec et sans retour haptique dans des scénarios adaptés. Pour le contrôle d'agent unique (S), le retour haptique réduit significativement le temps de complétion et améliore l'efficacité perçue. Pour le contrôle par ancrage (P), l'efficacité perçue s'améliore avec haptique sans différence significative sur le temps. Pour le contrôle par barycentre (C), l'intuitivité est significativement améliorée et la zone explorée augmente.

Un cas d'usage intégré démontre des transitions fluides entre modes de contrôle durant une mission complexe, avec maintien de la connectivité tout au long de l'exécution. Les résultats confirment que le retour haptique améliore les performances objectives et subjectives, avec des effets variables selon l'échelle de contrôle.

\section*{Discussion et perspectives}
Cette thèse lève des hypothèses simplificatrices des travaux antérieurs en considérant des contraintes réalistes de capteurs (FOV limité, mesures en repère propre) et propose une méthode d'estimation d'orientation décentralisée sans hypothèse de rigidité du graphe. Les architectures de contrôle à échelle variable offrent flexibilité et adaptabilité aux exigences des missions.

Les perspectives incluent l'extension du modèle dynamique pour considérer l'inclinaison des quadrotors affectant le FOV réel, l'intégration de mécanismes de récupération de connectivité après déconnexion temporaire, et l'ajout de capacités d'évitement d'obstacles avec occlusions visuelles. Le retour tactile multimodal pourrait compléter le retour de force, et des études à long terme exploreraient les effets d'apprentissage. Le déploiement sur équipes hétérogènes (aériens et terrestres) et l'intégration de SLAM collaboratif décentralisé pour environnements inconnus constituent des perspectives majeures. La maintenance de connectivité, associée à des interfaces haptiques intuitives, permet des missions autonomes complexes nécessitant intervention humaine ponctuelle.

\mainmatter

\adjustmtc

\clearemptydoublepage
\chapter*{Introduction}
\addcontentsline{toc}{chapter}{Introduction}
\chaptermark{Introduction}

\lettrine{R}{obotics} has emerged as a transformative field, revolutionizing industries, changing how humans interact with technology and opening new societal possibilities, with applications spanning industry, healthcare, agriculture, and exploration. Robots are progressively accepted as an integral part of our society and are being employed both in civil and military applications for decades. They are increasingly utilized to perform tasks that are dangerous, repetitive, or otherwise infeasible for humans. While early robotic systems were confined to structured environments, advances in sensing, computation, and actuation have enabled the deployment of robots in dynamic and unpredictable environments.

Within this broad field, multi-robot systems represent a critical area of research, offering the promise of enhanced efficiency, scalability, and robustness in task execution.
Unlike single-robot systems, multi-robot systems can collaborate to achieve objectives that would be impractical or impossible for a single agent. Applications of such systems include disaster response, environmental monitoring, and cooperative transportation. The ability of multiple robots to operate autonomously while maintaining coordination poses significant challenges in control, communication, and perception, particularly in dynamic environments.

In scenarios where human expertise remains indispensable, shared control paradigms have gained traction as a means to integrate human intuition with robotic precision. Shared control frameworks distribute the decision-making responsibilities between human operators and robotic systems, allowing robots to assist or augment the operator's actions. These frameworks are especially valuable in multi-robot systems, where the complexity of managing multiple agents can exceed the cognitive capacity of a single operator.

To further enhance the operator's ability to interact with multi-robot systems, haptic feedback has been introduced as a critical component of shared control interfaces. By providing kinesthetic or tactile cues, haptic feedback enables operators to perceive the state of the robots and their environment intuitively. This sensory channel not only improves situational awareness but also facilitates precise and responsive control, particularly in complex and dynamic environments.

\section*{Thesis contributions}
In light of these considerations, this thesis explores the intersection of multi-robot systems, shared control, and haptic feedback, with the goal of enabling efficient and intuitive remote operation of multi-robot teams. It seeks to advance the state of the art in shared control for multi-robot systems by addressing critical challenges and proposing innovative solutions. Specifically, we extend a fully decentralized connectivity maintenance scheme to address the limitations posed by body-frame measurements and the restricted field of view (FOV) of onboard sensors. While some techniques for relative orientation measurement exist, we propose a novel method for estimating relative orientations between robots using only local measurements and communication, further enhancing the system's capabilities.

Building on these theoretical contributions, we propose several haptic-enabled control methods with varying control scales over the robot group, designed to facilitate the remote operation of quadrotor UAV teams. These methods maintain connectivity while adapting to diverse task requirements, ranging from single-agent operation to coordinated group movement.

The proposed approaches are implemented in a simulation environment where each robot operates as an individual process, emphasizing the decentralized nature of the algorithms and their closeness to real-world implementation. These methods are validated through a user study conducted across three distinct scenarios and demonstrated in a comprehensive use case, showcasing seamless transitions between control modalities to meet varying task demands.

\section*{Structure of the Thesis}
This thesis is structured into three main parts.
The first part (\textbf{Part \ref{part:preliminaries}}) provides an introduction to robotics with a focus on multi-robot systems, covering both the theoretical foundations and practical tools necessary for their control.
The second part (\textbf{Part \ref{part:contributions}}) presents the core contributions of this work, that led to the author's publication~\cite{bernard2023decentralized} and a submission to the \textit{IEEE Transactions on Human-Machine Systems} journal. Finally, the third part (\textbf{Part \ref{part:conclusions}}) provides the conclusions of the thesis, outlines potential directions for future research, and includes an appendix that offer further technical details to aid the reader's understanding.

Below is a detailed summary of the content of each part.

\subsection*{Outline of Part \ref{part:preliminaries}}

This part provides a comprehensive review of the state of the art in multi-agent systems, localization, and shared control, along with the theoretical foundations of algebraic graph theory and formation control.
\bigskip
\begin{itemize}
    \setlength\itemsep{1em}
    \item[$\bullet$] \textbf{Chapt.~\ref{chap:multirobot}} introduces the fundamental concepts of multi-agent systems. It provides an overview of algebraic graph theory and briefly addresses key challenges in localization and sensing within multi-robot systems.
    \item[$\bullet$] \textbf{Chapt.~\ref{chap:sharedcontrol}} reviews the state of the art in shared control strategies for multi-robot systems, with a particular focus on UAVs. The chapter introduces haptic shared control and explores essential multi-agent control techniques, including consensus and formation control, emphasizing decentralized approaches and their application to systems operating under global constraints.
\end{itemize}

\subsection*{Outline of Part \ref{part:contributions}}

This part presents the core contributions of this thesis, which are organized into two chapters.
\begin{itemize}
    \setlength\itemsep{1em}
    \item[$\bullet$] \textbf{Chapt.~\ref{chap:modeling}} details the models for robot dynamics, measurements, and communication, forming the foundation for the subsequent chapters.
    \item[$\bullet$] \textbf{Chapt.~\ref{chap:dcmFOV}} presents a solution to the problem of \textit{connectivity maintenance} in spite of sensing limitations like (\textit{i}) limited field of view of the onboard cameras, and (\textit{ii}) limited range of the onboard cameras, (\textit{iii}) measurements expressed in individual body frames. The chapter also introduces a novel method for estimating relative orientations between robots using only local measurements and communication.
    \item[$\bullet$] \textbf{Chapt.~\ref{chap:chapitre5}} presents three haptic enabled control methods for the remote operation of quadrotor UAV teams while maintaining connectivity.
    These methods are validated through a user study involving three specific scenarios, and are demonstrated in a final use case, where the operator transitions seamlessly between control modalities to adapt to the task requirements.
\end{itemize}

\subsection*{Outline of Part \ref{part:conclusions}}

\begin{itemize}
    \setlength\itemsep{1em}
    \item[$\bullet$] This part presents the conclusions of the thesis, offering a summary of its key contributions. It also addresses the limitations and unresolved challenges encountered during the course of this work, proposing potential directions for future investigation and research.
\end{itemize}

\subsection*{Outline of Part \nameref{part:appendix}}

\begin{itemize}
    \setlength\itemsep{1em}
    \item[$\bullet$] \textbf{Appendix~\ref{app:software}} provides detailed information on the software architecture used to carry out the experiments described in Chapts.~\ref{chap:dcmFOV} and~\ref{chap:chapitre5}.
\end{itemize}

\clearemptydoublepage
\part{Preliminaries and state of the art}
\label{part:preliminaries}

\clearemptydoublepage
\graphicspath{{Chapitre1/figures/}}
\clearemptydoublepage
\chapter{Multi-agent systems}\label{chap:multirobot}
{\hypersetup{linkcolor=black}
\minitoc
}

\lettrine{T}{his} chapter introduces key concepts for understanding multi-agent systems.
While we first present the foundations of multi-agent systems and their applications before focusing on robotics usages, we then use the terms \textit{robot} and \textit{agent} interchangeably in this work.
We begin by outlining the motivations and reviewing the state of the art in multi-agent systems.
Next, we discuss a central aspect that enables robust and scalable coordination: decentralization.
To model and analyze multi-robot systems and their interactions, we introduce graph theory.
Finally, we explore localization methods for multi-robot systems, examining how they address the unique challenges of such environments.
The use of graph theory to model inter-agent interactions and design decentralized control strategies is a well-established approach in the literature.

\section{Motivations}
Multi-agent systems (MAS) consist of a group of autonomous agents working collaboratively or competitively to achieve complex goals.
These systems are increasingly used in domains such as robotics, distributed sensor networks, autonomous vehicles or distributed computation, where individual agents interact to achieve outcomes beyond the capacity of a single unit.
In MAS, agents can be homogeneous or heterogeneous, they can have different capabilities, functions and roles.
The strength of MAS lies in distributing functionality across multiple simple agents, avoiding the complexity of implementing all capabilities within a single unit, rendering the overall systems scalable, robust (to single agent failure) and flexible~\parencite{cortes2017coordinated}.

The design of MAS is a challenging task, as it requires addressing issues such as coordination, communication, and cooperation among agents.
Each agent must not only perform its local tasks but also respond to the overall system's objectives, requiring coordination mechanisms to ensure the balance between individual autonomy and collective behavior.
Such coordination often lies on communication among agents, which is especially challenging to ensure in dynamic and unstructured environments as agents may have energy constraints, limited communication ranges and bandwidth, as well as being subject to noise and interferences.
Control strategies must account for these constraints and ensure that agents can effectively communicate, coordinate, and adapt to changing conditions, while maintaining system stability and performance.

\begin{figure}[ht]
    \centering
    \subfloat[\label{fig:ferroMicroRobots}]
    {\includegraphics[width=0.40\linewidth]{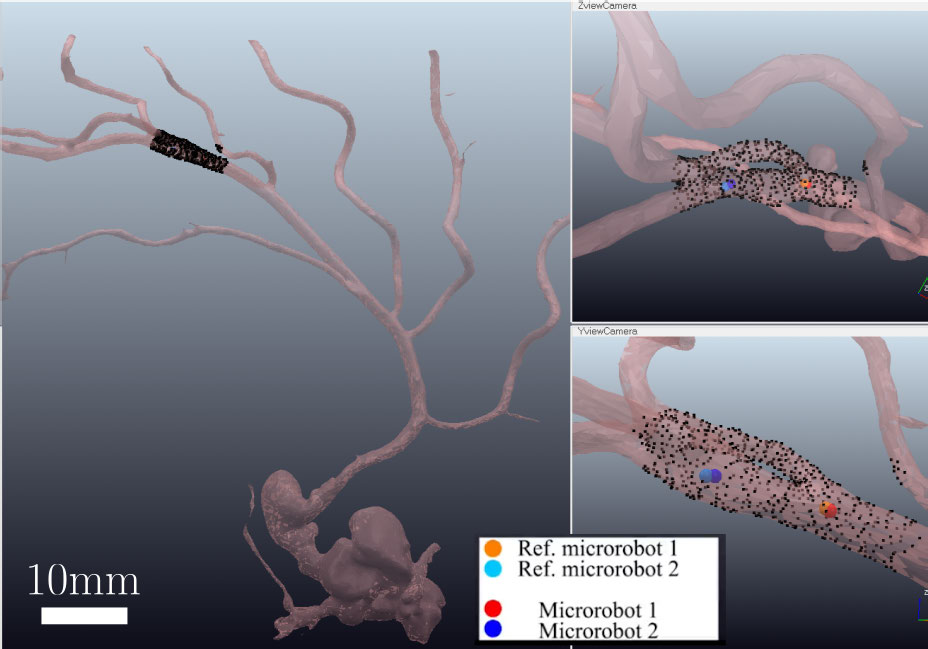}}
    \qquad
    \subfloat[\label{fig:kukaCarAssembly}]{\includegraphics[width=0.50\linewidth]{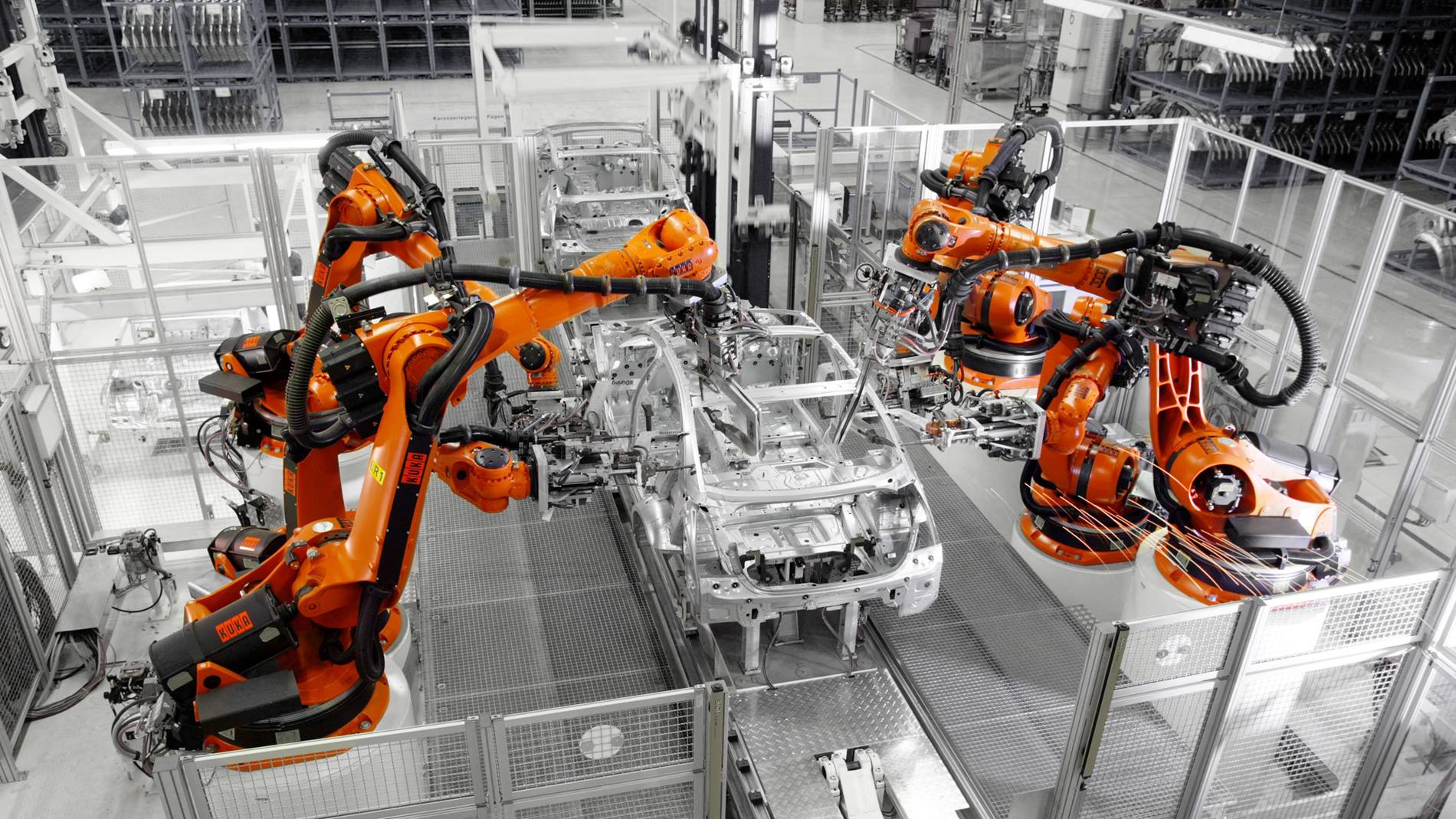}}
    \caption{MAS application examples. Fig.~\ref{fig:ferroMicroRobots}: Simulation of a micro-robot team in a 3D endovascular environment recreated from CT scans. The robots are represented by the red and blue spheres.
    Fig.~\ref{fig:kukaCarAssembly}: KUKA robots working together in the automotive industry.}%
\end{figure}

Multi-agent systems can be applied to robotics at different scales, from micro-robots to large-scale systems, and can be used in various applications such as monitoring, search and rescue, infrastructure inspection, or industrial automation~\parencite{mendoncca2016cooperative, murphy2011use, ota2006multi}.
\cite{ferro2024experimental} explores the use of multiple microrobots for various applications such as microsurgery, drug delivery, microscale manufacturing or environmental remediation, a simulation of a medicinal application is shown in Figure~\ref{fig:ferroMicroRobots}.
Industrial applications, such as car manufacturing, employ multiple robots that work together to perform complex tasks such as assembly, welding, or painting, as shown in Figure~\ref{fig:kukaCarAssembly}.

\begin{figure}[ht]
    \centering
    \subfloat[\label{fig:amazonWheel}]
    {\includegraphics[width=0.46\linewidth]{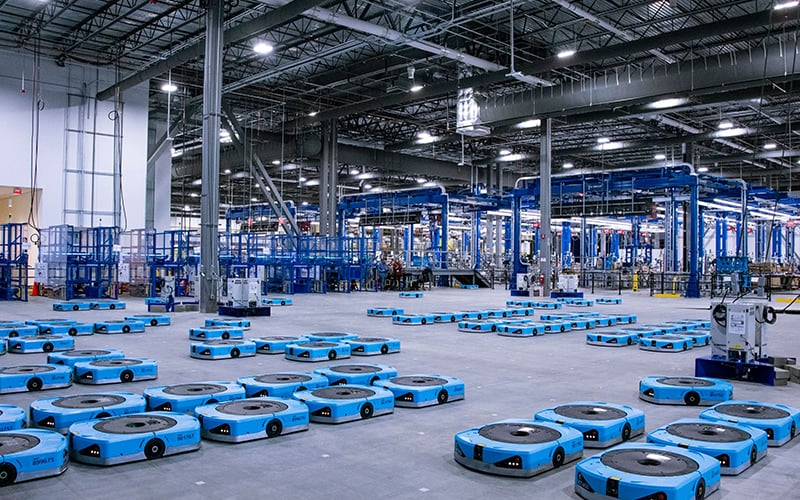}}
    \qquad
    \subfloat[\label{fig:amazonArm}]{\includegraphics[width=0.45\linewidth]{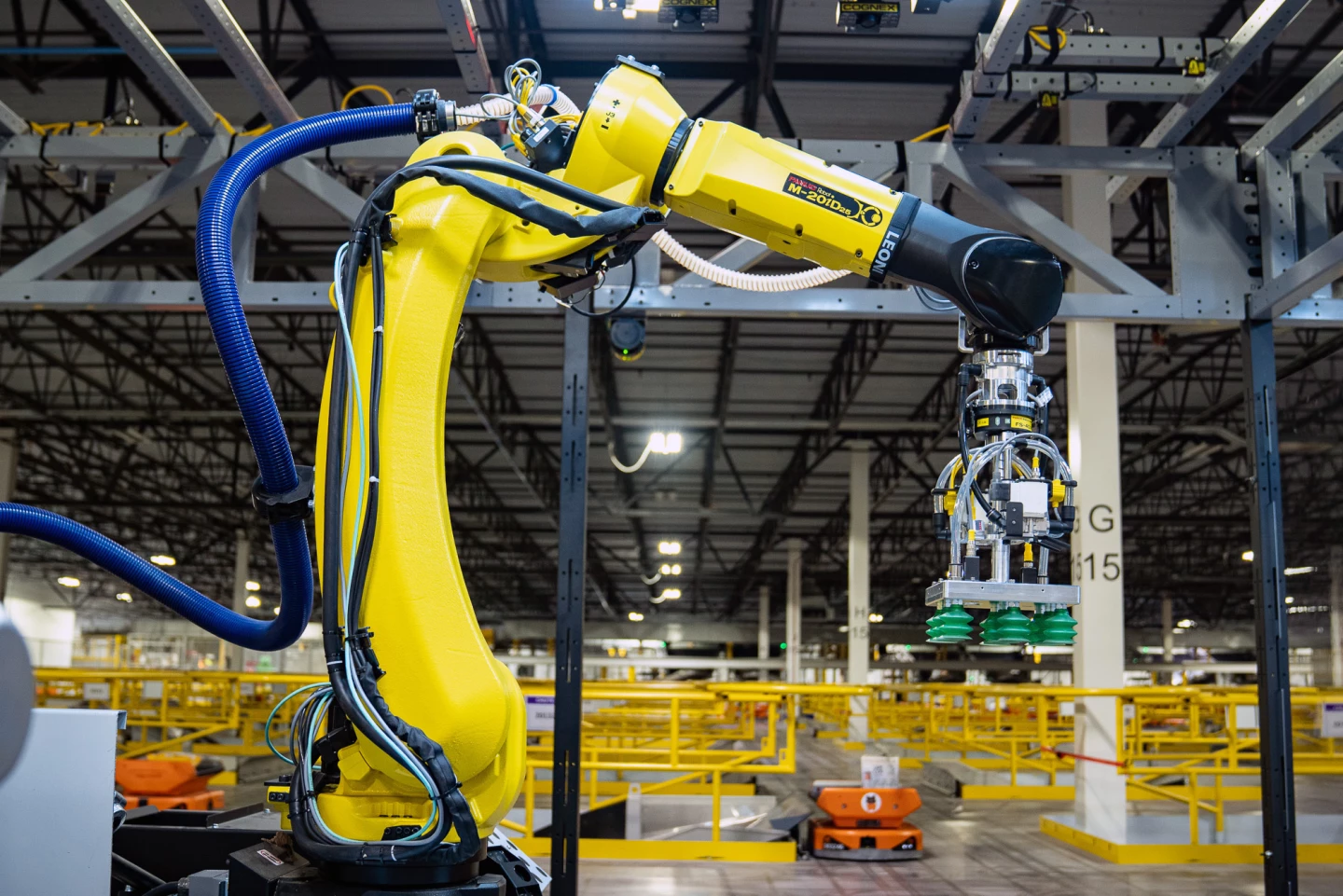}}
    \caption{\label{fig:amazonWarehouse}MRS applications examples. Fig.~\ref{fig:amazonWheel}: Hercules robots being tested at Amazon's BOS27 Robotics Innovation Hub \parencite{fmt:groundedRobots}.
    Fig.~\ref{fig:amazonArm}: Stationary robot arm inside Amazon's facility, designed to grasp and handle packages on conveyors and shelves \parencite{amazon:robotArm}.}%
\end{figure}

Another key domain for the development and application of multi-robot systems (MRS) in industry is warehouses automation.
Robots can efficiently perform repetitive tasks such as handling, sorting, and transporting goods, improving productivity and reducing human labor.
As illustrated in Figure~\ref{fig:amazonWarehouse}, Amazon's warehouses employ wheeled robots to transport goods and stationary robotic arms to grasp and handle packages on conveyors and shelves.
Other companies, such as Otto Motors (\url{https://ottomotors.com/}), offer solutions for material handling in the industry.
In such applications, robots can overcome the physical limitations of human workers, including heavy lifting, repetitive motion, and exposure to hazardous environments.
However, developing robots capable of handling boxes and bags without causing damage remains a significant challenge.
Moreover, controlling and coordinating multiple robots to operate collaboratively within a dynamic warehouse environment introduces additional complexity, requiring advanced algorithms and real-time communication systems.

A key interest in MAS lies in the development of systems capable of operating without a centralized control unit overseeing the entire system.
These systems rely on individual agents making autonomous decisions based on local information and interactions.
This decentralized approach leverages the advantages of MAS, such as scalability, robustness, and flexibility.
The following section will explore this concept of decentralization, discussing how it enables efficient coordination and cooperation within MAS.

\section{Decentralization}\label{sec:decentralization}
As discussed in~\cite{antonelli2013interconnected}, decentralized systems stand in contrast to centralized systems, where a single controller is responsible for making decisions for all agents.
Decentralization plays a fundamental role in various domains such as economics, finance, politics, and society, and it is also present in many daily activities.
For example, modern computing systems rely on multiprocessing, where tasks run concurrently across multithreaded, multicore processors. This distribution of tasks improves efficiency and precision, ensuring a seamless user experience, as discussed in~\cite{akhter2006multi}.
Real-time route optimization applications such as \textit{Google Maps} or \textit{Waze} coordinate with other agents to guide drivers, optimizing routes, managing congestion and reducing travel times, see~\cite{dresner2004multiagent}.

\begin{figure}[ht]
    \centering
    \subfloat[\label{fig:antBridge}]
    {\includegraphics[width=0.45\linewidth]{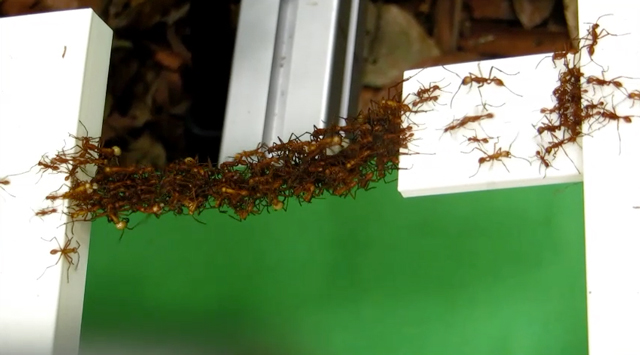}}
    \qquad
    \subfloat[\label{fig:japanMikoshi}]{\includegraphics[width=0.45\linewidth]{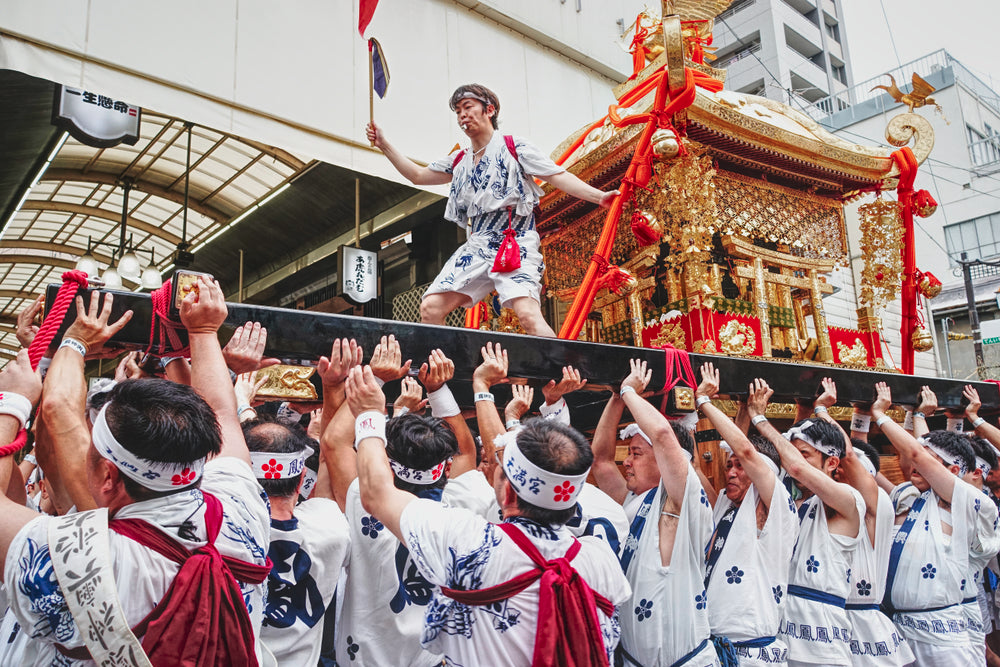}}
    \caption{Examples of decentralization. Fig.~\ref{fig:antBridge}: a group of ants collaboratively forming a bridge to reach a distant location, studied in~\cite{reid2015army}.
    Fig.~\ref{fig:japanMikoshi}: Osaka's Festival of Fire and Water, contingent collectively carrying a Mikoshi (portable shrine).}%
\end{figure}

Beyond technological applications, natural systems provide compelling evidence of decentralized behavior.
Social insects, such as ants, bees, and termites, demonstrate complex collective behaviors without centralized control.
These behaviors, including nest building, foraging, and bridge formation, emerge from local interactions and adaptive responses within the colony, as depicted in Figure~\ref{fig:antBridge} and analyzed in~\cite{sumpter2006principles, reid2015army}.

Decentralization is also prevalent in biological systems.
For instance, the immune system operates through the coordinated actions of distributed cells that detect and eliminate pathogens without a central controller.
In social contexts, decentralization can be observed when groups of individuals collaboratively perform tasks requiring collective effort.
A notable example is the lifting and transport of a Mikoshi (portable shrine) during Osaka's Festival of Fire and Water, where participants work together without a central directive to achieve a shared goal, as illustrated in Figure~\ref{fig:japanMikoshi}.
In such cases, communication between individuals typically occurs only when necessary, such as when maneuvering through obstacles.
Similarly, agents in decentralized systems activate their sensing capabilities only at critical moments, rather than maintaining continuous awareness throughout the entire task.
For instance, during the collaborative transportation of a load, individuals do not continuously observe one another.
Instead, they rely on intermittent cues to adjust their actions when required, ensuring efficient coordination with minimal communication overhead.
This selective interaction strategy contributes to the strength and adaptability of decentralized systems.

Robotics can significantly benefit from decentralized approaches, as they enable the development of scalable, resilient, and flexible multi-robot systems.
In decentralized systems, each agent operates independently, making decisions bagsed on local information.
In contrast, a centralized robotic system relies on a single controller to oversee the entire fleet, requiring communication with each individual agent \parencite{belta2004abstraction,antonelli2006kinematic}.
This approach significantly increase both computational and communication loads.
Decentralization keeps algorithmic complexity manageable regardless of system size while enhancing robustness by eliminating single points of failure.
Furthermore, it reduces the information burden on individual agents as the number of agents in the system grows by limiting inter-agent interactions to local neighborhoods, promoting greater flexibility.
These qualities are essential for the effective operation of large multi-robot systems.

Decentralized multi-robot control strategies have been widely studied in the literature, with applications ranging from collision avoidance \parencite{arul2021v} and coverage \parencite{breitenmoser2010voronoi} to search and rescue missions \parencite{horyna2023decentralized} and patrolling \parencite{aggravi2021decentralized}.

\section{Graph theory}%
\label{sec:graphTheory}
As discussed earlier, multi-agent systems operate under the assumption that each agent has access only to partial information, typically gathered through local interactions with neighboring agents.
Graph theory offers a powerful framework to model these interactions, with agents represented as nodes and their communication or sensing capabilities as edges.
This approach is fundamental for analyzing system connectivity, information flow, and coordination, making graph theory an essential tool in designing control algorithms for multi-robot systems.

\subsection{Notation}
Here, we introduce the notations commonly used in robotic multi-agent systems.
\medskip
\begin{itemize}
    \setlength\itemsep{1em}
    \item[$\bullet$] \textbf{Scalar quantities}: we usually represent scalar quantities by lowercase symbols such as $\omega, d$ or $\lambda$.
    \item[$\bullet$] \textbf{Vectors}: Elements of $\mathbb{R}^N$ are expressed as bold lowercase letters such as $\boldsymbol{v}, \boldsymbol{p}$.
    \item[$\bullet$] \textbf{All-ones and all-zeros}: $\boldsymbol{1}_N$ and $\boldsymbol{0}_N$ denote vectors of all ones and all zeros of dimension $N$.
    \item[$\bullet$] \textbf{Matrices}: matrices are represented by bold uppercase letters such as $\boldsymbol{A} \in \mathbb{R}^{n\times m}$, and scalar elements of matrices are denoted as uppercase letters such that $A_{ij}$ with ${0<i\leq n}$, $0<j\leq m$ is the element of $\boldsymbol{A}$ on row $i$ and column $j$.
    \item[$\bullet$] \textbf{Hadamard product}: We represent the element-wise product between two matrices, the so-called Hadamard procuct, with $\circ$.
    \item[$\bullet$] \textbf{Kronecker product}: We represent the Kronecker product between two matrices with $\otimes$.
    \item[$\bullet$] \textbf{Set cardinality}: $|\mathcal{V}|$ represents the cardinality of set $\mathcal{V}$.
    \item[$\bullet$] \textbf{Matrix transpose}: $\boldsymbol{A}^T$ denotes the transpose of matrix $\boldsymbol{A}$.
    \item[$\bullet$] \textbf{Matrix inverse}: $\boldsymbol{A}^{-1}$ denotes the inverse of matrix $\boldsymbol{A}$.
    \item[$\bullet$] \textbf{Matrix pseudo-inverse}: $\boldsymbol{A}^{\dagger}$ denotes the pseudo-inverse of matrix $\boldsymbol{A}$.
    \item[$\bullet$] \textbf{Identity matrix}: $\boldsymbol{I}_N$ denotes the identity matrix of dimension $N$.
\end{itemize}

We now introduce the notation for multi-robot systems in the context of graph-based modeling:
\medskip
\begin{itemize}
    \setlength\itemsep{1em}
    \item[$\bullet$] \textbf{Agents as Vertices}: Each agent is represented as a vertex $i$ in a graph $\mathcal{G} = (\mathcal{V}, \mathcal{E})$, where $\mathcal{V} = \{1...N\}$ is the set of vertices (agents) in the system and $\mathcal{E}\subseteq\mathcal{V}\times\mathcal{V}$ denotes the set of \emph{edges} in a graph.
    \item[$\bullet$] \textbf{Edges as Connections}: An edge ${e_k := (i,j) \in \mathcal{E}}$ is an ordered pair representing a direct connection from vertex $i$ to vertex $j$. The presence of an edge ${e_k := (i,j)}$ implies that agent $i$ can sense and communicate with agent $j$. An edge is \emph{undirected} if the connection is reciprocal, i.e. $(i,j) \in\calE  \iff (j,i) \in\calE $, and \emph{directed} otherwise. Throughout this thesis, we primarily focus on undirected graphs.
    \item[$\bullet$] \textbf{Neighbor Set}: For a given node $i$, the set of neighbors $\mathcal{N}_i$ is defined as the collection of vertices connected to $i$ by an edge. Formally, $\calN_i=\{j\in\calV|\,(i,\,j)\in\calE \}$.
    \item[$\bullet$] \textbf{Edge Weights}: Each edge $(i,j)$ can be associated with a weight $w_{ij}$. Here, we consider weights ranging between 0 and 1, and symmetric, i.e. $w_{ij} = w_{ji}$, in line with the assumption of undirected graphs.
\end{itemize}

\subsection{Graph Properties}
\begin{figure}[htp]
    \centering
    \subfloat[\label{fig:completeGraph}]
    {\includegraphics[width=0.21\linewidth]{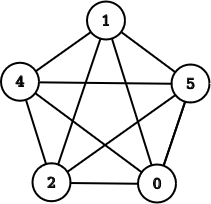}}
    \qquad
    \subfloat[\label{fig:directedGraph}]
    {\includegraphics[width=0.21\linewidth]{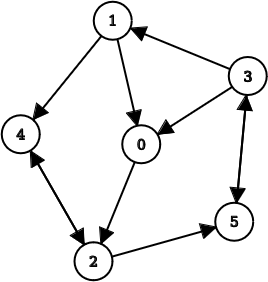}}
    \qquad
    \subfloat[\label{fig:connectedGraph}]
    {\includegraphics[width=0.21\linewidth]{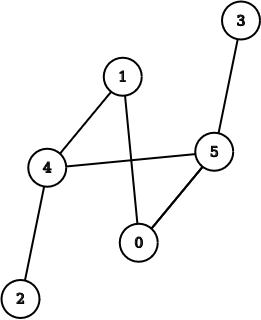}}
    \qquad
    \subfloat[\label{fig:disconnectedGraph}]
    {\includegraphics[width=0.21\linewidth]{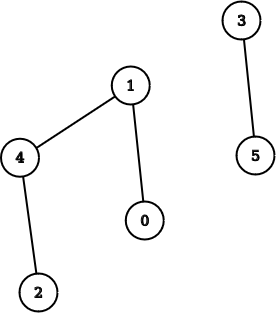}}
    \caption{Fig.~\ref{fig:completeGraph}: complete graph, Fig.~\ref{fig:directedGraph}: strongly connected graph, Fig.~\ref{fig:connectedGraph}: connected graph, Fig.~\ref{fig:disconnectedGraph}: disconnected graph.}%
\end{figure}

Here are key concepts in graph theory relevant to multi-robot systems:
\begin{itemize}
    \setlength\itemsep{1em}
    \item[$\bullet$] \textbf{Completeness}: A graph is considered \emph{complete} if every pair of vertices is connected by an undirected edge, see Figure~\ref{fig:completeGraph} for an illustration.
    \item[$\bullet$] \textbf{Paths and Connectivity}: \begin{itemize}
        \item \emph{Directed Path}: A directed path exists from vertex $i$ to vertex $j$ if there is a sequence of directed edges connecting $i$ to $j$ within a directed graph. For example, in Figure~\ref{fig:directedGraph}, there is a directed path from vertex 5 to vertex 4, passing through nodes 3 and 1.
        \item \emph{Undirected Path}: An undirected path exists between two vertices $i$ and $j$ if there is a sequence of undirected edges connecting $i$ to $j$ within an undirected graph. For example, in Figure~\ref{fig:connectedGraph}, there is a path between vertices 5 and 2 passing through node 4.
        \item \emph{Strong Connectivity}: A directed graph is \emph{strongly connected} if there is a directed path between every pair of vertices, as shown in Figure~\ref{fig:directedGraph}.
        \item \emph{Connectivity}: An undirected graph is \emph{connected} if there is an undirected path between every pair of vertices, as shown in Figure~\ref{fig:connectedGraph}.
    \end{itemize}
\end{itemize}

\subsection{Algebraic Graph Theory}
\label{sec:algebraicGraphTheory}
Graphs are not only effective for visually representing multi-agent systems, but they also offer a robust suite of mathematical tools for analysis.
These algebraic tools can be employed to assess various properties within a multi-agent system such as connectivity.
In the following section, we introduce some of these key algebraic concepts.

\begin{figure}[htp]
    \captionsetup[subfigure]{labelformat=empty}
    \centering
    \subfloat[$\lambda_2 = 4$]
    {\includegraphics[width=0.21\linewidth]{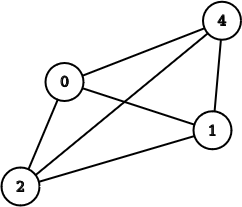}}
    \qquad
    \subfloat[$\lambda_2 = 2$]
    {\includegraphics[width=0.21\linewidth]{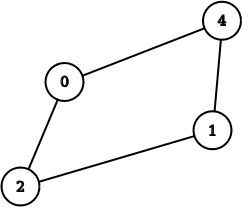}}
    \qquad
    \subfloat[$\lambda_2 = 0.58$]
    {\includegraphics[width=0.21\linewidth]{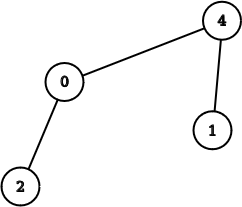}}
    \qquad
    \subfloat[$\lambda_2 = 0$]
    {\includegraphics[width=0.21\linewidth]{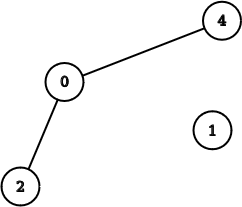}}
    \caption{\label{fig:lambdagraph}Graphs with different connectivity eigenvalues $\lambda_2$}%
\end{figure}

To represent and manipulate graphs using linear algebra, we encode graphs in matrix form.
For a graph $\mathcal{G} = (\mathcal{V}, \mathcal{E})$ with $N$ vertices and $M$ edges, we define several matrices that encapsulate the structure and relationships within the graphs:

\begin{itemize}
    \setlength\itemsep{1em}
    \item[$\bullet$] \textbf{Adjacency matrix}: This $N \times N$ matrix represents the connections between vertices. Each element $A_{ij}$ is nonzero if there is an edge between vertex $i$ and vertex $j$.
    The adjacency matrix can be binary or weighted:
        \item \emph{Binary adjacency matrix}: $\boldsymbol{A} \in \{0,1\}^{N\times N}$ is defined as
        \begin{equation}
            \boldsymbol{A}_{ij} = \begin{cases}
                1 & \text{if } (i,j) \in \mathcal{E},\\
                0 & \text{otherwise}.
            \end{cases}
        \end{equation}
        \item \emph{Weighted adjacency matrix}: $\boldsymbol{A} \in \mathbb{R}^{N\times N}$ is defined as
        \begin{equation}
            \boldsymbol{A}_{ij} = \begin{cases}
                w_{ij} & \text{if } (i,j) \in \mathcal{E},\\
                0 & \text{otherwise}.
            \end{cases}
        \end{equation}

    \item[$\bullet$] \textbf{Degree matrix}: This diagonal $N \times N$ matrix holds the degree of each vertex along the diagonal. Each diagonal element $D_{ii}$ represents the number of edges connected to vertex $i$:
    \begin{equation}
        \boldsymbol{D}_{ii} = \sum_{j=1}^{N} \boldsymbol{A}_{ij}.
    \end{equation}
    This definition applies to both binary and weighted adjacency matrices.

    \item[$\bullet$] \textbf{Incidence matrix}: This $N \times M$ matrix represent the relationship between vertices and edges in a graph. Each column $\boldsymbol{B}_k$ corresponds to an edge $e_k$ and contains two nonzero entries, $-1$ and $1$, indicating the start and end vertices of the edge:
    \begin{equation}
        \boldsymbol{B}_{ij} = \begin{cases}
            -1 & \text{if } i \text{ is the start vertex of edge } j,\\
            1 & \text{if } i \text{ is the end vertex of edge } j,\\
            0 & \text{otherwise}.
        \end{cases}
    \end{equation}
    For \emph{undirected graphs}, the incidence matrix can be constructed by assigning arbitrary directions to each edge, even though no true directionality exists in undirected edges.
    \item[$\bullet$] \textbf{Laplacian matrix}: This $N \times N$ matrix is defined as the difference between the degree matrix and the adjacency matrix:
    \begin{equation}
        \laplacian = \boldsymbol{D} - \boldsymbol{A}.
    \end{equation}
    Alternatively, the Laplacian can also be expressed as:
    \begin{equation}
        \laplacian = \boldsymbol{B}\boldsymbol{B}^T.
    \end{equation}
    where $\boldsymbol{B}$ is the incidence matrix.
    For an \emph{undirected graph}, the Laplacian matrix is \emph{positive semi-definite}, meaning all its eigenvalues $\lambda_i$ are real and non-negative.
    The smallest eigenvalue is zero.
    Ordering the eigenvalues in ascending order $0 = \lambda_1 \leq \lambda_2 \leq \ldots \leq \lambda_N$, an undirected graph $\mathcal{G}$ is connected if and only if $\lambda_2 > 0$~\parencite{fiedler1973algebraic, fiedler1989laplacian}.
    The eigenvalue $\lambda_2$ is known as the \emph{Fiedler value} or the \emph{connectivity eigenvalue}.
    Refer to Figure~\ref{fig:lambdagraph} for examples of graphs with different values for $\lambda_2$.
\end{itemize}

\section{Localization and sensing}

\begin{figure}[ht]
    \centering
    \includegraphics[width=0.7\linewidth]{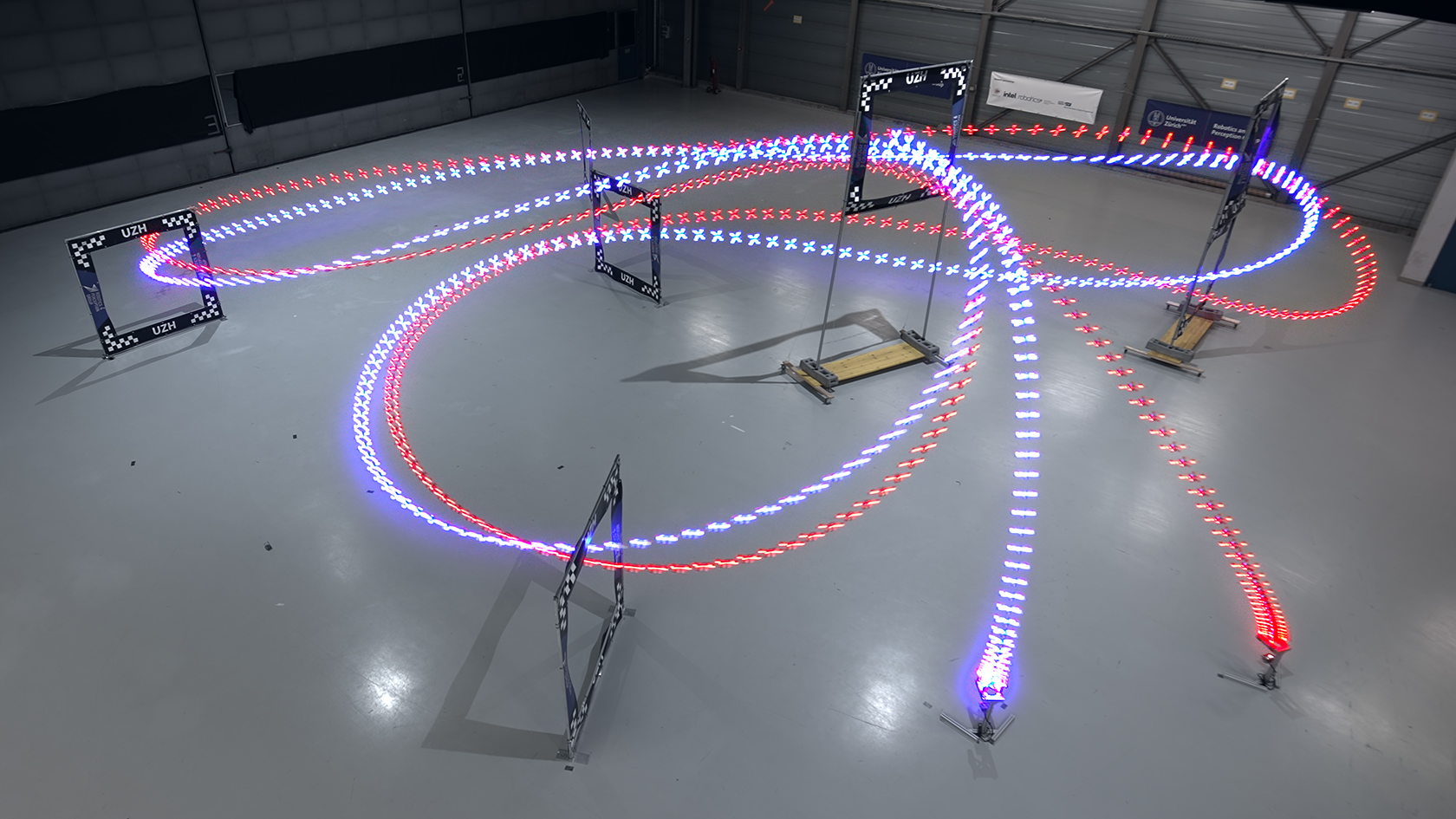}
    \caption{\label{fig:drone_race}Drone race opposing human pilots and an autonomous drone, presented in~\cite{kaufmann2023champion}. The autonomous drone was trained using a trajectory dataset captured via global sensing techniques (laser tracker) and dynamically adapts its flight-path mid-race by detecting racing gates with local sensing techniques (onboard camera).}%
\end{figure}

Sensing mechanisms are broadly categorized into two types: \textit{global sensing} and \textit{local sensing}.
\begin{itemize}
    \setlength\itemsep{1em}
    \item[$\bullet$] \textbf{Global Sensing}: In global sensing, each agent has access to information about the entire system, regardless of its proximity to other agents.
    Global sensing is often used in centralized systems, where a single controller has access to all agents' states and can make decisions based on the complete system information.
    Examples of global sensing systems include GPS or centralized motion capture systems such as Vicon, OptiTrack or Qualisys.
    Different global sensing applications offer distinct advantages and limitations.
    For instance, the Global Navigation Satellite System (GNSS) provides precise positioning information, with accuracies reaching the centimeter level~\parencite{guo2018multi}.
    However, it is prone to signal loss in urban canyons or indoor environments, limiting its reliability in such settings.

    Motion capture systems  are renowned for their exceptional accuracy in position and orientation measurements, often achieving millimeter-level precision, as highlighted in~\cite{merriaux2017study}.
    For example,~\cite{antonini2018blackbird} present a dataset where ground-truth positions were recorded using a Vicon motion capture system to support autonomous aggressive and agile flight of unmanned aerial vehicles.
    Similarly,~\cite{delmerico2019we} provides a dataset for the same application, utilizing a laser tracker device based on static close-range reference points to capture ground-truth positions.
    The outcomes of training a racing drone using such datasets are demonstrated in~\cite{kaufmann2023champion}, where an autonomous quadrotor competes against human pilots in a drone race, see Figure~\ref{fig:drone_race}.
    Despite their accuracy, these systems are constrained to controlled environments, require an unobstructed line-of-sight between cameras and agents, and necessitate specific markers or tags on the agents for effective tracking.

    \item[$\bullet$] \textbf{Local Sensing}: In local sensing, each agent can only perceive information from its immediate surroundings, typically within a limited sensing radius.
    Local sensing is often used in decentralized systems, where agents must rely on their neighbors to gather information about the global system.
    Examples of local sensing systems include onboard cameras, LiDARs, or ultrasonic sensors.
    Local sensing systems, when compared to global sensing methods, exhibit greater robustness against signal loss and environmental disturbances as they rely on local information.
    This makes them particularly well-suited for operation in dynamic and unstructured environments.

    For example,~\cite{kaufmann2023champion} describes a drone race between human pilots and an autonomous drone, as shown in Fig.~\ref{fig:drone_race}.
    The autonomous drone was trained using datasets obtained through global sensing techniques but adjusts its velocity during flight using the video stream from its onboard camera, demonstrating its reliance on local sensing capabilities.
\end{itemize}

The combination of global and local sensing is considered a best practice for achieving accurate and reliable localization.
Autonomous vehicles, such as self-driving cars, commonly integrate LiDARs, cameras, GNSS, and inertial measurement units (IMUs) to achieve precise localization within their environment~\parencite{geiger2012we, bresson2017simultaneous}.

\subsection{Relative sensing}\label{sec:relativeSensing}
In algebraic graph theory, the weights $w_{ij}$ in the weighted adjacency matrix $\boldsymbol{A}$, discussed in Sect.~\ref{sec:algebraicGraphTheory}, are often determined by the \textit{relative configuration} between agent $i$ and $j$.
This \textit{relative configuration} refers to geometric and positional parameters such as relative positions, orientations, and distances between agents.
Consequently, relative sensing becomes a pivotal component in multi-robot systems, enabling agents to ascertain their positions and orientations relative to their neighbors.
We introduce the following relative sensing measurements commonly used in multi-robot systems:

\boitemagique{Bearing}{
    The bearing indicates the direction of a neighboring agent relative to the local reference frame of the observing agent.
    It is represented as a unit vector pointing from agent $i$ to agent $j$ in the reference frame of agent $i$:
    \begin{equation*}
    \boldsymbol{\beta}_{ij} = \frac{\boldsymbol{p}_j - \boldsymbol{p}_i}{||\boldsymbol{p}_j - \boldsymbol{p}_i||},
    \end{equation*}
    where $\boldsymbol{p}_i$ and $\boldsymbol{p}_j$ denote the positions of agents $i$ and $j$, respectively.
}
Bearing measurements are typically obtained using onboard cameras or other vision-based sensors.
They have been widely used in tasks such as localization and formation control.
For example,~\cite{mariottini2009vision} propose a vision-based localization system that employs bearing measurements to estimate the relative positions of neighboring robots, implementing a formation control strategy.
Similarly,~\cite{schiano2016rigidity} explores the use of bearing rigidity to maintain graph structures in formation control.

\boitemagique{Distance}{
    The distance measurement provides the range between two agents and is computed as the Euclidean distance between their positions:
    \begin{equation*}
    d_{ij} = ||\boldsymbol{p}_j - \boldsymbol{p}_i||,
    \end{equation*}
    where $\boldsymbol{p}_i$ and $\boldsymbol{p}_j$ denote the positions of agents $i$ and $j$, respectively.
}
Distance measurements can be obtained using various sensors, including stereo cameras, ultrasonic sensors, LiDARs, or Ultra-Wideband radios.
These measurements are essential for maintaining formations in multi-robot systems.
For instance,~\cite{anderson2008rigid} present a formation control algorithm based on graph rigidity principles that relies on distance measurements to ensure that agents maintain the desired relative positions.

\boitemagique{Relative position}{
    The relative position measurement is the vector from agent $i$ to agent $j$ in a common reference frame $\mathcal{W}$:
    \begin{equation*}
    \boldsymbol{p}_{ij}^{\mathcal{W}} = \boldsymbol{p}_j - \boldsymbol{p}_i,
    \end{equation*}
    where $\boldsymbol{p}_i$ and $\boldsymbol{p}_j$ denote the positions of agents $i$ and $j$, respectively.
    In this work, we consider bodyframe relative positions, where the reference frame is the body frame of the measuring agent.
    For instance, the relative position measurement from agent $i$ to agent $j$ expressed in agent $i$'s body frame is expressed as:
    \begin{equation*}
        \boldsymbol{p}_{ij} = \RtoAgent{i}(\boldsymbol{p}_j - \boldsymbol{p}_i).
    \end{equation*}
    where $\RtoAgent{i}$ is the rotation matrix from the global reference frame $\mathcal{W}$ to agent $i$'s body frame.
}

Relative position measurements are more informative than distance or bearing alone.
They are often used in systems that require precise positioning for tasks such as collaborative manipulation or tight formation control.
However, they are also harder to obtain, often requiring more complex sensors.

\boitemagique{Relative orientation}{
    The relative orientation measurement represents the angular difference between the orientations of two agents.
    Specifically, the relative yaw considers the angular difference along the Z-axis:
    \begin{equation*}
    \psi_{ij} = \psi_j - \psi_i,
    \end{equation*}
    where $\psi_i$ and $\psi_j$ are the yaw angles (rotations around the Z-axis) of agents $i$ and $j$, respectively.
}
In this thesis, we consider relative orientation specifically along the Z-axis (yaw) for both simplicity and practical relevance.
This focus on yaw is common in multi-robot systems, particularly in aerial robotics, where controlling roll and pitch can be challenging.
For underactuated robots, such as quadrotors with fixed cameras, adjusting roll or pitch to aim the camera typically induces undesired linear motions due to the vehicle's dynamics.
Therefore, yaw-based orientation is often prioritized for multi-robot applications~\parencite{franchi2012shared,decarli2021online}

\begin{figure}[htp]
    \centering
    \subfloat[\label{fig:uvdarHardware}]
    {\includegraphics[width=0.35\linewidth]{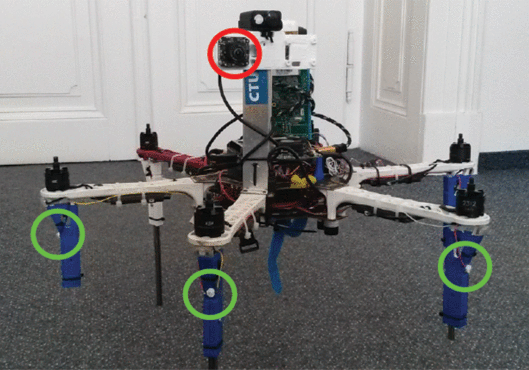}}
    \qquad
    \subfloat[\label{fig:uvdarOr}]{\includegraphics[width=0.58\linewidth]{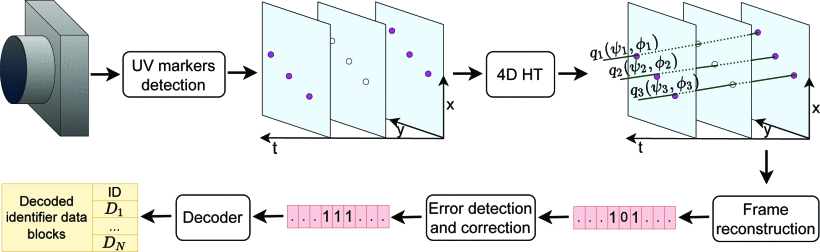}}
    \caption{\label{fig:uvdarSystem}UVDAR and UVDAR-COM systems. Fig.~\ref{fig:uvdarHardware}: Configuration of the UVDAR system, featuring UV LEDs circled in green and cameras circled in red.
    Fig.~\ref{fig:uvdarOr}: UVDAR-COM system workflow, including time tracking of UV markers using 4D Hough Transformation, frame reconstruction, error detection, and data decoding.}%
\end{figure}

It is essential to emphasize that for relative measurements to hold meaning, agents must not only detect their neighbors but also identify them.
Often overlooked in the literature, this identification process ensures that agents can accurately associate the relative measurements they perceive with the correct neighbors.
Identification can be achieved through various methods, including unique identifiers, visual markers, or communication protocols.
\cite{walter2019uvdar} address this challenge by introducing an innovative relative localization system for drones.
This system utilizes blinking ultraviolet (UV) light-emitting diodes (LEDs), acting as active markers, paired with UV cameras, detecting neighbors' LEDs, see Fig~\ref{fig:uvdarSystem}.
The use of the UV spectrum provides robustness against ambient light interference, making it effective in diverse environments, including indoor and outdoor settings.
By employing blinking LEDs at distinct frequencies, the system not only identifies individual drones but also determines the observed side of a sensed neighbor, facilitating accurate relative orientation measurements.
This approach has been further enhanced in~\cite{horyna2022uvdar} to enable short-range inter-agent communication, expanding its functionality and applicability.

In summary, relative sensing is a critical component in multi-robot systems, allowing agents to measure their positions and orientations relative to their neighbors.

\subsection{Collaborative localization}
\label{sec:collaborativeLocalization}
\textit{Collaborative localization}, also known as \textit{cooperative localization}, is another fundamental challenge in multi-robot systems.
In the context of this thesis, the term refers to the process of estimating the configuration of multiple agents within a shared reference frame using a sparse set of relative measurements between robots.
The primary objective is to address the localization problem without depending on centralized localization system, such as a GNSS or motion capture system.

First introduced by~\cite{kurazume1994cooperative}, the problem of cooperative localization has been extensively studied through various approaches to robot localization, including probabilistic methods~\parencite{fox1999collaborative, fox2000probabilistic}, Kalman filtering techniques~\parencite{roumeliotis2002distributed, roumeliotis2000collective}, and observer-based frameworks~\parencite{decarli2024adaptive, schiano2016rigidity}.
A critical requirement for collaborative localization algorithms is the ability to associate relative measurements with the corresponding identities of neighboring agents.
For effective operation, an agent must reliably link each measurement to a specific neighbor.
The sensing system described in~\cite{walter2019uvdar} and introduced in the previous section achieves this capability.

Throughout this thesis, all proposed algorithms assume that the identities of neighboring agents are known.
Therefore, the issue of processing anonymous measurements is beyond the scope of this work.

\section{Conclusions}

This chapter introduced key concepts in multi-agent systems (MAS), covering decentralization and graph theory as foundations for agent coordination.
These tools are essential for understanding the dynamics of multi-robot systems and designing effective control strategies.
We also examined localization and sensing techniques which enables agents to estimate their positions relative to one another without centralized systems.
Finally, we discussed collaborative localization and the importance of identifying neighboring agents to associate relative measurements accurately.
This critical challenge grants agents the ability to estimate their positions within a shared reference frame, facilitating effective multi-robot coordination and supervision by a human operator.

Graph theory can be used to encode and enforce sensing and communication limitations in multi-agent formations, allowing us to design high-level control strategies directly related to these physical limitations.
By leveraging the local nature of these interactions, we can achieve decentralized applications allowing for scalability, versatility, and resilience in multi-robot systems.
The next chapter will build on this by introducing formation control strategies, shared control approaches combining human and autonomous inputs, and the role of haptic feedback in enhancing human-robot interaction.

\clearemptydoublepage
\graphicspath{{Chapitre2/figures/}}
\clearemptydoublepage
\chapter{Shared Control of UAVs}\label{chap:sharedcontrol}
{\hypersetup{linkcolor=black}
\minitoc
}

\lettrine{I}{n} this chapter, we examine a specific scenario for multi-robot applications: the remote operation of a fleet of Unmanned Aerial Vehicles (UAVs) by a human operator.
This scenario presents unique challenges due to the physical distance from the robots and the complexity of simultaneously controlling multiple agents in dynamic environments.
To address these challenges, researchers have explored various methods to provide the remote operator with effective information feedback, which we discuss in the first section of this chapter.
In the second section, we explore different control strategies designed to enable a single operator to oversee multiple agents, leveraging the concepts introduced in the first chapter.
Finally, we present the formation control method designed to preserve global constraint that forms the foundation of this thesis: decentralized global connectivity maintenance.

The use of Unmanned Aerial Vehicles (UAVs) has demonstrated potential in a variety of applications, ranging from mountain rescue operations~\parencite{silvagni2017multipurpose} to surveillance and humanitarian missions~\parencite{custers2016flying}.
Given that these scenarios are often dynamic and unstructured, research into intuitive human-robot interfaces for effective control has become increasingly critical.
These interfaces must balance between ease of control and informative feedback to allow the user to effectively execute tasks while receiving clear, intuitive information from the robots.
Typically, such feedback is provided through a visual interface displaying real-time data on robot positions, states, and environmental context.
However, haptic feedback systems are increasingly recognized for their potential to guide users more dynamically, as examined in~\cite{ferro2021online} and~\cite{aggravi2020connectivitymaintenance}.
By integrating multimodal feedback, user awareness and control can be significantly enhanced, particularly in complex or visually demanding situations.

\section{Telerobotics}
\begin{figure}[ht]
    \centering
    \subfloat[\label{fig:mechanicalTelepoeration}]
    {\includegraphics[width=0.42\linewidth]{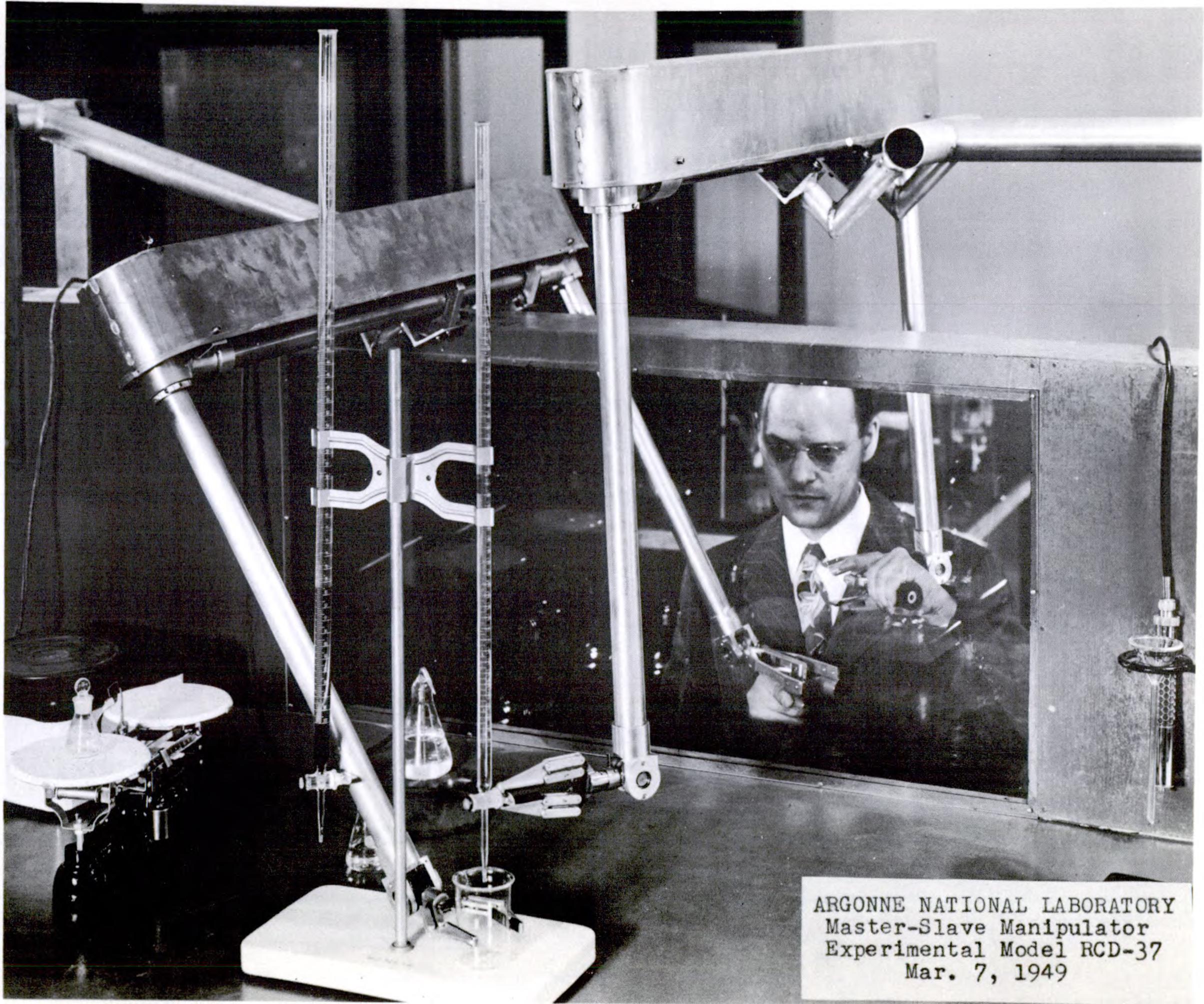}}
    \qquad
    \subfloat[\label{fig:phantomDevice}]{\includegraphics[width=0.42\linewidth]{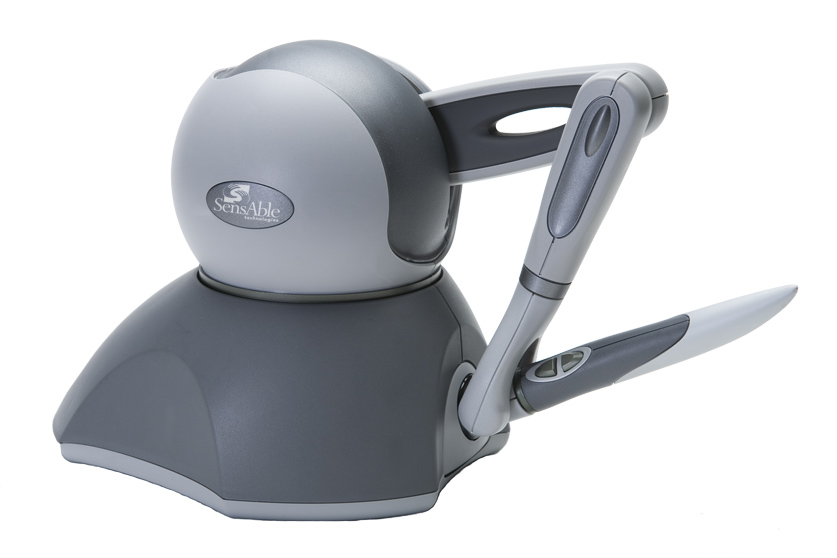}}
    \caption{Fig.~\ref{fig:mechanicalTelepoeration}: Mechanical teleoperation system developed by Goertz in 1954. The system allowed operators to manipulate radioactive materials from a safe location.
    Fig.~\ref{fig:phantomDevice}: PHANToM device, 6 Degreed of Freedom haptic device developed by Sensable Technologies.}%
\end{figure}

The term \textit{telerobotics}, also known as \textit{teleoperation} or \textit{telemanipulation}, is derived from the Greek word \textit{tele}, meaning ``distant''~\parencite{niemeyer2016telerobotics}.
Teleoperation systems were first introduced in the 1940s to meet the demands of nuclear research, see Fig.~\ref{fig:mechanicalTelepoeration}, introduced in~\cite{goertz1954mechanical}.
These early systems allowed human operators to manipulate radioactive materials from a safe distance while providing haptic feedback from the remote environment.
Initially, teleoperation systems relied on mechanical linkages, which limited their range of operation.
To overcome this constraint, electrically coupled manipulators were developed in subsequent decades to support nuclear applications and enable remote handling over greater distances~\parencite{vertut2013teleoperation}.

As nuclear-related activities declined, interest in telerobotics shifted toward other domains, including space exploration, medical applications, and underwater robotics.
This transition was facilitated by advancements in computing power and the development of haptic devices such as the PHANToM device shown in \ref{fig:phantomDevice} and presented in~\cite{massie1994phantom}.
Similarly to nuclear applications, the adoption of telerobotics in these fields was driven by the need to perform tasks in environments where human operators could not be physically present, either due to safety concerns or the high costs associated with direct human involvement.

\subsection{Teleoperation of UAVs}

Commercial aerial robots typically rely on proprietary remote controllers or smartphone applications. However, researchers have been exploring more intuitive control methods to enhance user experience and effectiveness. Gesture-based control has captured significant attention, with several studies employing cameras to recognize natural gestures for drone navigation~\parencite{ng2011collocated,sanna2013kinect,pfeil2013exploring}. Gaze control, which leverages eye-tracking technology for UAV navigation, is another promising approach, particularly beneficial for users with physical impairments~\parencite{hansen2014use}.
In addition, brain-computer interfaces (BCIs) have been investigated, enabling UAV control through the interpretation of EEG signals~\parencite{yu2012flyingbuddy2}.
Wearable interfaces, including custom-made suits, exoskeletons, and gloves, have also been developed to provide immersive control experiences~\parencite{rognon2018flyjacket,sandru2016glove}. More recently, infrared hand trackers such as the Leap Motion device have gained popularity for gesture-based control, offering precise detection of hand movements for drone navigation~\parencite{sarkar2016gesture,fernandez2016natural,chandarana2017natural}.

\subsection{Visual Feedback}

Effective remote operation of robots typically requires providing the operator with visual feedback about the surrounding environment.
For single-robot operations, the most common approach is streaming video from the robot onboard cameras~\parencite{fattah2016r3diver}.
However, this approach becomes quickly impractical for multi-robot applications, where it is challenging for a single user to simultaneously monitor multiple video feeds.
Streaming multiple videos is also costly in terms of communication bandwidth and energy consumption.

\begin{figure}[ht]
    \centering
    \subfloat[\label{fig:orbSlam2}]{\includegraphics[width=0.45\linewidth]{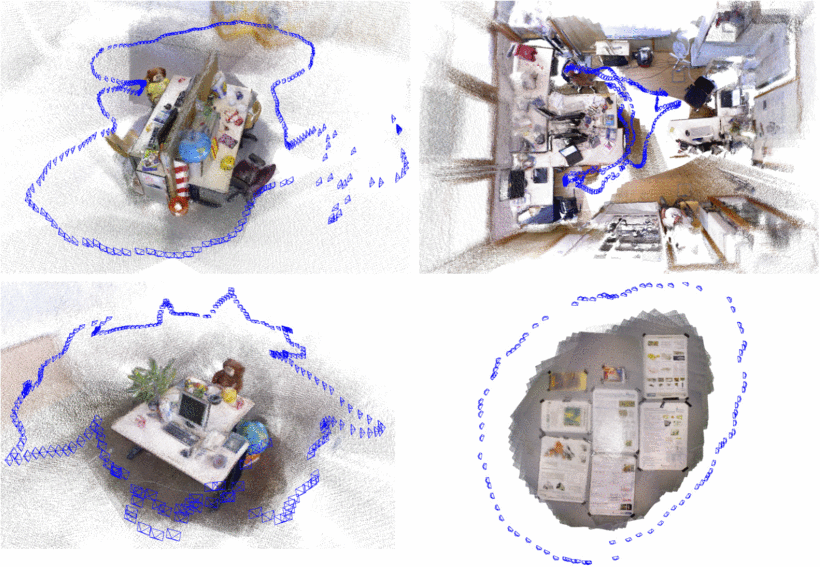}}
    \qquad
    \subfloat[\label{fig:swarmSlam}]{\includegraphics[width=0.45\linewidth]{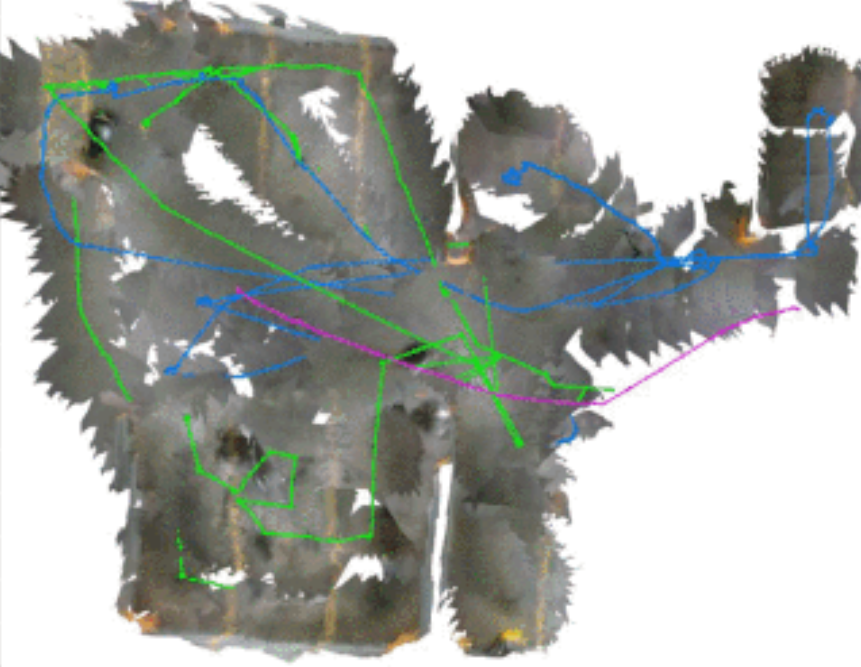}}
    \caption{Fig.~\ref{fig:orbSlam2}: dense point cloud reconstructions generated by ORB-SLAM2, utilizing estimated keyframe poses and sensor depth maps, presented in~\cite{murartal2017orbslam2}.
    Fig~\ref{fig:swarmSlam}: map and pose graph from a Swarm-SLAM experiment involving three robots, detailed in~\cite{lajoie2023swarm}.}
\end{figure}

To address these challenges, visualization techniques often leverage data fusion, aggregation, and abstraction to reconstruct the environment and track the robots' positions and states.
A common approach is the creation of a "digital twin" of the remotely operated system in a virtual environment, offering a comprehensive view of the robot's state~\parencite{Tsokalo2019remote, barresi2022beyond}.
The environment may be pre-defined or dynamically reconstructed using Simultaneous Localization and Mapping (SLAM) techniques~\parencite{engel2014lsdslam, murartal2017orbslam2}.
For example, Fig.~\ref{fig:orbSlam2} shows a dense point cloud reconstruction generated by ORB-SLAM2, a widely-used SLAM algorithm.
Alternatively, other methods, such as the Next-Best-View path planning algorithm, are employed to model the environment.
A cluster-based approach for this algorithm is proposed in~\cite{hardouin2020next}.
The approach discussed here aims to achieve accurate 3D reconstruction of large structures with minimal runtime by utilizing a fleet of UAVs.

For multi-robot systems, visualizing the entire group requires knowledge of each robot's position and state within a shared reference frame. In controlled environments or simulations with motion capture systems~\parencite{aggravi2021hapticenabled, aggravi2021decentralized}, this information is readily available. However, in real-world applications, it must be estimated through sensor data and collaborative localization algorithms, as discussed in Section~\ref{sec:collaborativeLocalization}.

Collaborative SLAM (C-SLAM) extends the SLAM framework to multi-robot systems, allowing robots to share sensor data, construct a unified map, and localize themselves within it.
Both centralized~\parencite{ahmed2023entropy} and decentralized~\parencite{dubois2022sharing, lajoie2023swarm} C-SLAM approaches have been explored using various sensors, such as LiDAR, stereo, and RGB-D. Fig.~\ref{fig:swarmSlam} presents a map and pose graph generated by a Swarm-SLAM experiment with three robots, as described in~\cite{lajoie2023swarm}.

\subsection{Haptic Feedback}

Traditional teleoperation systems primarily relied on visual feedback to guide the operator. However, visual feedback alone can impose a high cognitive load on the operator~\parencite{abdi2020haptics} and may degrade performance when the view is partially occluded.
To address these limitations, researchers have investigated the integration of haptic feedback as a means to enhance the operator's situational awareness and improve control accuracy~\parencite{wildenbeest2012impact, pacchierotti2023cutaneous}.

The broad definition of \textit{haptics} refers to the human sense of touch, which relies on both kinesthetic (force and position) and cutaneous (tactile) receptors.
These sensory mechanisms enable humans to perceive, interact with, and manipulate objects in their environment.
Everyday tasks highlight the importance of this sense, for instance, writing, reaching for an object in the dark, or buttoning a shirt all require haptic feedback

More fundamentally, even basic motor functions would be impossible without haptic information.
Humans rely on cutaneous sensations in their feet and fingertips to walk, move their limbs, or grasp objects, even in well-lit environments.
Without these tactile inputs, precise movement control would be significantly impaired.

In robotics and virtual reality literature, \textit{haptics} refers to the various combinations of real and simulated touch interactions between humans, robots, and their environments, whether real, remote, or virtual. These interactions are facilitated by \textit{haptic interfaces}, which allow human operators to experience the sense of touch in teleoperated or simulated environments~\parencite{hannaford2016haptics}.

\begin{figure}[ht]
    \centering
    \subfloat[\label{fig:chinelloThree}]{\includegraphics[width=0.4\linewidth]{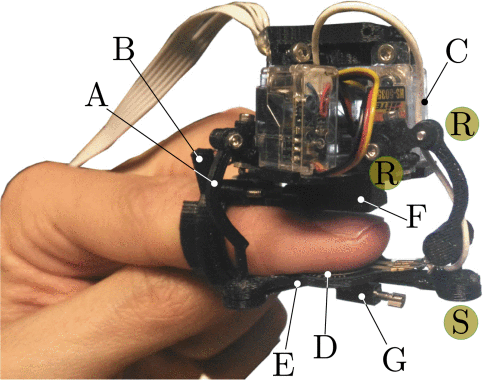}}
    \qquad
    \subfloat[\label{fig:electrotactile}]{\includegraphics[width=0.5\linewidth]{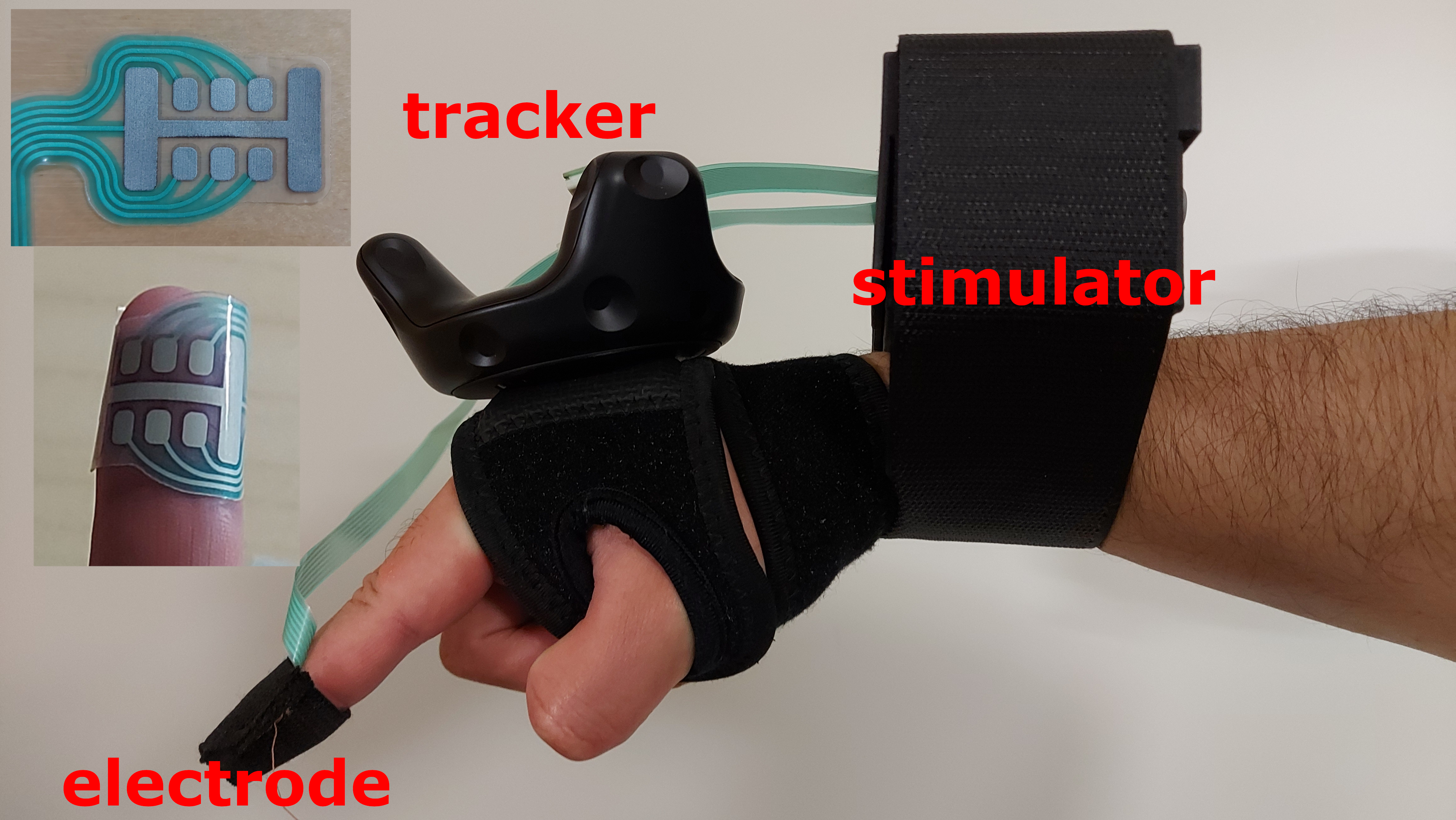}}
    \caption{\label{fig:wearableDevices}Fig.~\ref{fig:chinelloThree}: The Revolute-Revolute-Spherical fingertip device presented in~\cite{chinello2017three}.
    Fig~\ref{fig:electrotactile}: An electrode, placed in contact with the finger pad, is controlled by an electrical stimulator attached to the forearm~\parencite{vizcay2021electrotactile}}
\end{figure}

Haptic feedback systems can be broadly categorized into two types: \textit{grounded} and \textit{wearable} interfaces.
Wearable devices, designed to be worn by the user, provide tactile feedback through mechanisms embedded in gloves, suits, or vests.
Given the high density of nerve receptors in the fingers, research has focused on designing haptic devices specifically dedicated to fingertip use.
Through contact simulation~\parencite{music2019human}, vibration~\parencite{chinello2017three}, pressure~\parencite{chinello2012three}, or electric stimulation~\parencite{vizcay2021electrotactile}, these devices aim to provide precise tactile feedback, capitalizing on the sensitivity of the fingertips to enhance the user's ability to perceive texture, pressure, and other environmental cues during interactions with remote or virtual environments. Fig.~\ref{fig:wearableDevices} illustrates two examples of wearable haptic devices.
Tactile haptic feedback is less commonly employed in drone teleoperation compared to other haptic modalities~\parencite{aggravi2018design, aggravi2021decentralized, aggravi2016haptic}.

To address the need for effective haptic feedback in drone teleoperation, researchers have explored various wearable devices, including cable-driven systems~\parencite{rognon2018haptic}, vibrotactile armbands~\parencite{cacace2016multimodal}, and wearable tactile gloves~\parencite{tsykunov2019swarmtouch}.
These devices aim to convey information about drone trajectories, formation states, and environmental interactions, enhancing the operator's ability to control and perceive the system's behavior intuitively.

\begin{wrapfigure}{L}{0.35\linewidth}
    \centering
    \includegraphics[width=0.9\linewidth]{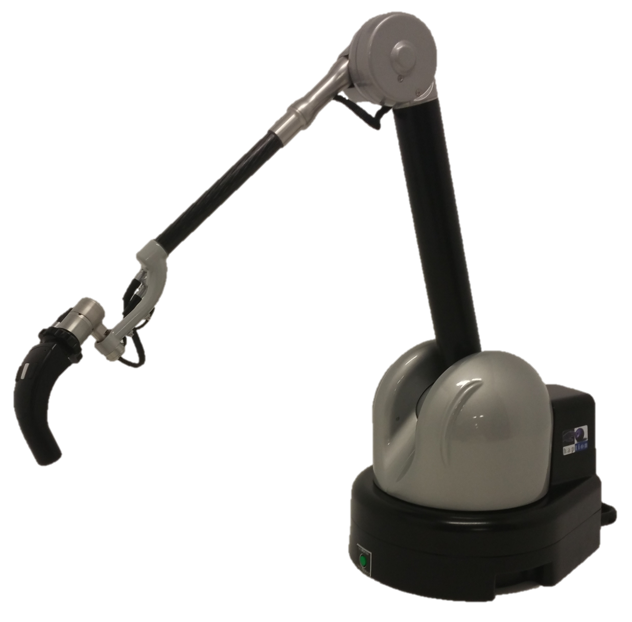}
    \caption{\label{fig:VirtuoseDevice1}Virtuose 6D haptic device, Haption.}
\end{wrapfigure}
Compared to \textit{wearable} devices, \textit{grounded} haptic interfaces can deliver kinesthetic feedback, providing the human operator with information about position, velocity, forces, and torques. These interfaces have been employed in various applications, including force-feedback devices used to control fleet leaders and convey team status updates~\parencite{lee2013semiautonomous, robuffo2011iros, franchi2012shared}. Additionally, some researchers have explored hybrid approaches that combine haptic feedback with visual or auditory cues to further enhance the operator's situational awareness~\parencite{aggravi2020connectivitymaintenance}.
This thesis specifically utilizes the Virtuose 6D device, shown in Fig.~\ref{fig:VirtuoseDevice1}, a high-precision haptic interface developed by Haption.
The Virtuose 6D delivers 6 degrees of freedom (DoF) force feedback, enabling intuitive 6 DoF motion control by the operator.

As the complexity of multi-robot systems continues to grow, providing rich and intuitive feedback to human operators becomes increasingly important. These feedback mechanisms are designed to improve the operator's understanding of the robots' state and the surrounding environment, thereby enhancing control accuracy and mission success rates. The nature of the haptic feedback information is often determined by the type of control method employed for operating the drones and typically is linked to aspects that the user can directly influence.

\section{Multi-agent control}
Beyond addressing the challenge of a user's physical absence from the remote environment, telerobotics is also motivated by the complementary capabilities of humans and robots.
Humans contribute cognitive flexibility, high-level decision-making, and adaptability to complex tasks.
In contrast, robots offer precision, strength, and motion scaling, which are critical for scenarios requiring fine manipulation.
Additionally, robots provide autonomy, data aggregation, and low-level control, enabling the management of tasks involving multiple robots that would be beyond the capabilities of a single human operator. This synergy between human intelligence and robotic capabilities enhances the effectiveness and versatility of telerobotic systems.

In this context, \textit{consensus algorithms} enable agents to align their states by exchanging local information, forming the basis of distributed decision-making.
\textit{Formation control} focuses on maintaining desired spatial configurations among agents, which is essential for tasks like exploration, transportation, or surveillance.
Lastly, \textit{shared control} integrates human input with autonomous decision-making, allowing an operator to guide the system while leveraging the collective intelligence of the agents while providing the user with information feedback and control assistance.

\subsection{Consensus}
\label{sec:consensus}

Consensus-based algorithms, often referred to as \emph{agreement algorithms} in the literature~\parencite{bullo2009distributed}, originated in the field of computer science~\parencite{lynch1996distributed} and have since become foundational in multi-robot research~\parencite{kuriki2014consensus}.
These algorithms address the consensus problem, which aims to ensure all agents within a system converge to a common value for a specific variable, relying solely on information exchanged with a local subset of neighboring agents.

Continuous-time consensus algorithms are inherently tied to the Laplacian matrix of the communication graph.
Consider a system of $N$ agents, where each agent has an internal state $\agentPos_i \in \mathbb{R}^m$ and simple integrator dynamics defined as $\dot{\agentPos}_i = \vv_i$, with $\vv_i$ representing the control input for agent $i$.
With $\mathcal{G}$ representing the communication topology, the continuous-time consensus algorithm is expressed as:
\begin{equation}
    \dot{\agentPos} = -\laplacian\agentPos,
\end{equation}
where $\laplacian$ is the Laplacian matrix of graph $\mathcal{G}$.
If $\mathcal{G}$ is connected, the system will converge to a common state, i.e. $\lim_{t\to\infty} \agentPos_i = \lim_{t\to\infty} \agentPos_j$ for all $i,j\in\mathcal{V}$.

For individual agents, convergence to a common state is guaranteed by applying the following control input:
\begin{equation}
    \vv_i = -\sum_{j\in\calN_i}w_{ij}(\agentPos_i - \agentPos_j),
\end{equation}
with $\calN_i$ the set of neighbors of agent $i$ and $w_{ij}$ the weight of the edge connecting agents $i$ and $j$ (i.e. the $(i,j)$-th element of the adjacency matrix $\boldsymbol{A}$).

Let us assess the stability of the consensus algorithm.
We recall that for a connected undirected graph $\mathcal{G}$, the Laplacian matrix $\laplacian$ is symmetric, positive semi-definite (i.e. all its eigenvalues are non-negative), and its smallest eigenvalue is zero and corresponds to the eigenvector $\boldsymbol{1}_N$ (a vector of all ones)~\parencite{bullo2018lectures}.
Let us define a Lyapunov candidate function based on the disagreement among agents:
\begin{equation}
    V(\agentPos) = \frac{1}{2}\agentPos^T\laplacian\agentPos.
\end{equation}
Since $\laplacian$ is positive semi-definite, $V(\agentPos) \geq 0$ for all $\agentPos$ and $V(\agentPos) = 0$ if and only if $\agentPos$ lies in the null space of $\laplacian$, i.e. $\agentPos = \alpha\boldsymbol{1}_N$ for some scalar $\alpha$.
From this definition, we can derive the time derivative of $V(\agentPos)$:
\begin{equation}
    \dot{V}(\agentPos) = \frac{d}{dt}\left(\frac{1}{2}\agentPos^T\laplacian\agentPos\right)
\end{equation}
Using the product rule:
\begin{equation}
    \dot{V}(\agentPos) = \frac{1}{2}\left(\dot{\agentPos}^T\laplacian\agentPos + \agentPos^T\laplacian\dot{\agentPos}\right)
\end{equation}
Since $\laplacian$ is symmetric ($\laplacian = \laplacian^T$):
\begin{equation}
    \dot{V}(\agentPos) = \agentPos^T\laplacian\dot{\agentPos}
\end{equation}
From the consensus algorithm, $\dot{\agentPos} = -\laplacian\agentPos$, so:
\begin{equation}
    \dot{V}(\agentPos) = \agentPos^T\laplacian(-\laplacian\agentPos) = -\agentPos^T\laplacian^2\agentPos
\end{equation}
Since $\laplacian$ is positive semi-definite, $\laplacian^2$ is also positive semi-definite, so $\dot{V}(\agentPos) \leq 0$ for all $\agentPos$, implying that $V(\agentPos)$ is non-increasing over time.
Therefore, the disagreement among agents' states decreases over time.
Furthermore, $\dot{V}(\agentPos) = 0$ if and only if $\agentPos$ lies in the null space of $\laplacian$, i.e. $\agentPos = \alpha\boldsymbol{1}_N$ for some scalar $\alpha$.
This implies that the agents' states converge to a common value.

From~\cite{bullo2018lectures}, $\dot{V}(\agentPos) < -\lambda_2V$, where $\lambda_2$ is the second smallest eigenvalue of $\laplacian$.
Thus, the consensus algorithm is globally exponentially stable if $\lambda_2 > 0$ (i.e. if the graph is connected).
This proof holds for undirected graphs; proofs can be found in the literature for directed graphs, see~\cite{moreau2004stability}.

The consensus protocol can be applied at various levels and applied to multi-agent systems control, serving different objectives:
\begin{itemize}
\setlength\itemsep{1em}
\item[$\bullet$] Rendezvous: Agents aim to converge to a common position in space.
\item[$\bullet$] Alignment: Agents adjust their orientations to achieve uniform alignment.
\item[$\bullet$] Synchronization: Agents synchronize their internal clocks to operate cohesively.
\item[$\bullet$] Distributed Estimation: Agents collaboratively estimate and agree upon the value of a shared parameter.
\end{itemize}

Many decentralized control and estimation laws are based on the consensus protocol.
Examples include the distributed average consensus algorithm introduced in~\cite{olfati2004consensus}, the distributed Kalman filter detailed in~\cite{olfati2007distributed}, and the distributed optimization algorithms discussed in~\cite{nedic2018distributed}.

\subsection{Formation Control}

\begin{figure}[ht]
    \centering
    \includegraphics[width=0.6\linewidth]{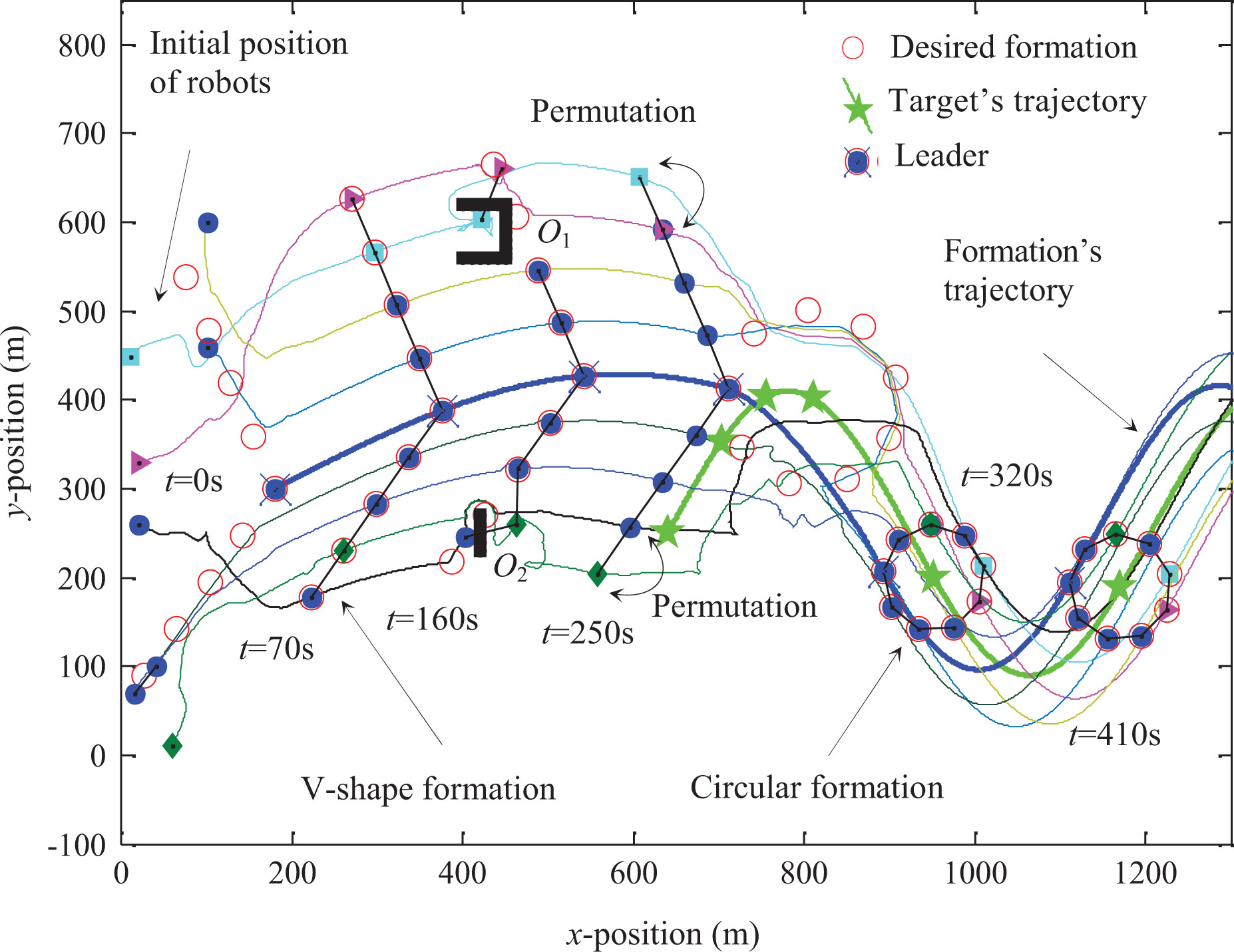}
    \caption{Formation control of a group of robots maintaining a desired shape while tracking a moving target and avoiding obstacles, see~\cite{dang2019formation}.}
    \label{fig:formationControl}
\end{figure}

Controlling multiple robots simultaneously presents unique challenges, and various strategies have been proposed to address these issues.
Formation control typically involves imposing geometric constraints on the relative state of agents.
Depending on the robots' states and their sensing capabilities, specific constraints can be applied to achieve different behaviors.
These constraints may include inter-agent distances~\parencite{anderson2008rigid, lee2013semiautonomous} or angles between agents~\parencite{basiri2010distributed, bishop2011stabilization, franchi2012modeling}.
This method is particularly valuable in scenarios that require precise positioning of multiple drones.
The example provided in~\cite{dang2019formation} illustrates a formation control algorithm that enables a group of autonomous robots to track a moving target while navigating through environments with obstacles.
The algorithm ensures that the robots maintain various predefined formations, defined by a set of inter-agent distances and relative angles, adjusting their positions dynamically to avoid collisions, as depicted in Fig.~\ref{fig:formationControl}.

In contrast to rigid formation control, leader-follower strategies enable dynamic formation control by having follower agents maintain relative positions to a leader using virtual spring-like couplings.
This approach supports both autonomous navigation and operator guidance, ensuring flexible and cohesive multi-agent coordination~\parencite{franchi2012bilateral, ren2008distributed, consolini2008leader}.
Dynamic leader selection has been explored by some researchers, optimizing tracking performance by changing the leader identity in real-time~\parencite{franchi2016online,franchi2018online}. This approach adapts to changing environmental conditions or mission requirements.

Formation control methods has been further developed to maintain \textit{global properties} such as \textit{connectivity} or \textit{rigidity} across the entire formation~\parencite{robuffo2013passivity,zelazo2015decentralized,schiano2017bearing}.

\subsection{Shared Control}
Unlike applications where robotic systems are fully teleoperated~\parencite{zucker2015general} or fully autonomous~\parencite{wurman2008coordinating}, shared control systems combine human and autonomous control to leverage the strengths of both approaches~\parencite{queralta2020collaborative, selvaggio2018haptic}.
In sush systems, the human operator and the autonomous control system work together to achieve a shared goal \parencite{abi2018user, ferro2024experimental}.

\begin{figure}[ht]
    \centering
    \includegraphics[width=0.7\linewidth]{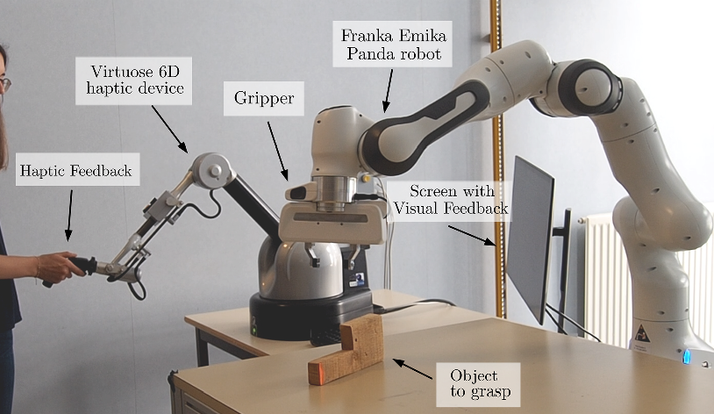}
    \caption{Teleoperation setup. The human operator uses a Virtuose 6D grounded haptic interface (Haption, France) to control a 7 degree of freedom Panda robotic manipulator (Franka Emika, Germany),
    used as a 6-DoF manipulator. The user receives haptic feedback through the haptic interface and visual feedback from an LCD screen~\parencite{rahal2023haptic}.}
    \label{fig:hapticGuidanceRahal}
\end{figure}

For instance, in~\cite{ferro2021online}, a human operator controls a robotic arm to insert a needle into soft tissues.
While the operator guides the arm toward the target, an autonomous system assists by ensuring a safe trajectory and providing haptic feedback to indicate the needle's penetration depth.
Similarly,~\cite{rahal2023haptic} investigates haptic feedback to reduce manipulator torque during remote manipulation tasks. The system also provides visual and haptic guidance to help the operator select an appropriate grasping pose for displacing an object, as shown in Fig.~\ref{fig:hapticGuidanceRahal}.

Shared control systems are particularly advantageous in multi-robot systems, where the complexities of coordinating multiple robots can quickly overwhelm a single human operator \parencite{selvaggio2021shared, aggravi2021hapticenabled, rahal2020caring}
In these systems, the human operator can focus on high-level decision-making, while the autonomous system assists with coordination, obstacle avoidance, or maintaining formation or connectivity~\parencite{turco2024reducing, zelazo2012rigidity, franchi2012shared}.

\section{Formation Control under Global Constraints: Connectivity Maintenance}\label{section:connectivityMaintenanceIntro}
This section addresses the connectivity maintenance challenge in multi-robot systems.
Properly modeling each robot's ability to sense and communicate with surrounding robots is fundamental to achieving autonomy in such systems.
As introduced in Section~\ref{sec:algebraicGraphTheory}, algebraic graph theory offers a powerful framework for representing these interactions.
In this context, an edge between two agents signifies their capability to exchange information, either through communication or sensing.
Among the key properties of a graph is its connectivity, defined as the ability to traverse between any two nodes via direct or indirect paths and quantified by the connectivity eigenvalue $\lambda_2$.

In multi-robot systems, ensuring global connectivity within the robotic group is essential for enabling decentralized tasks such as consensus~\parencite{olfati2007distributed}, rendezvous~\parencite{dimarogonas2007rendezvous}, flocking~\parencite{olfati2006flocking}, leader-follower strategies~\parencite{mariottini2009vision}, and other collaborative operations.
Consequently, maintaining connectivity throughout the system's operation has become a central research area, with two predominant approaches emerging:
\begin{itemize}
    \setlength\itemsep{1em}
    \item[$\bullet$] \textbf{Conservative methods}: These methods aim to preserve the initial connected graph topology by maintaining existing edges. For example,~\cite{ji2007distributed} considers an inter-robot sensing model based on maximum range with a fixed graph topology. in~\cite{dimarogonas2008connectedness}, the possibility to permanently add edges over time is included.
    \item[$\bullet$] \textbf{Flexible methods}: In contrast, flexible approaches allow the graph topology to evolve dynamically, permitting the addition and removal of edges as long as global connectivity is maintained. For instance,~\cite{kim2006maximizing} proposed a centralized approach to achieve a specific $\lambda_2$ value by placing a set of robots with maximum-range constraints in an obstacle-free environment.~\cite{de2006decentralized} pursued a similar objective with a decentralized solution.
\end{itemize}

\medskip
This thesis focuses on flexible and decentralized strategies for solving the connectivity maintenance problem in multi-robot systems.
In particular,~\cite{robuffo2013passivity} introduced a fully decentralized approach capable of handling complex sensing models, dynamically forming or dissolving inter-agent links while preserving global connectivity, and executing additional tasks beyond sole connectivity maintenance.

\subsection{Notations}
We now recall the definitions introduced in Sect.\ref{sec:graphTheory}.
Let $\mathcal{G} = (\mathcal{V}, \mathcal{E})$ represent the graph associated with the multi-robot system, where $\mathcal{V}$ is the set of vertices corresponding the robots, and $\mathcal{E}$ is the set of edges representing the inter-robot communication links.

Let $\boldsymbol{A} \in \mathbb{R}^{N \times N}$ denote the adjacency matrix of $\mathcal{G}$, where $N = |\mathcal{V}|$ is the number of robots.
The matrix $\boldsymbol{A}$ is symmetric and has elements $A_{ij} \geq 0$, with $A_{ij} = 0$ if $(i,j) \notin \mathcal{E}$, and $A_{ij} > 0$ otherwise.

The Laplacian matrix, $\laplacian \in \mathbb{R}^{N \times N}$, is defined as $\laplacian = \boldsymbol{D} - \boldsymbol{A}$, where $\boldsymbol{D}$ is the degree matrix of the graph, a diagonal matrix with elements $D_{ii} = \sum_{j=1}^{N} A_{ij}$.
Because of its symmetry and positive semi-definiteness, all $N$ eigenvalues of $\laplacian$ are real and non-negative.

When the eigenvalues are ordered in ascending order, $0 \leq \lambda_1 \leq \lambda_2 \leq \ldots \leq \lambda_N$, we have $\lambda_1 = 0$, and the second smallest eigenvalue, $\lambda_2$, known as the \textit{connectivity eigenvalue} or \textit{Fiedler eigenvalue}, is strictly positive if and only if the graph is globally connected.

Finally, let $\mathbf{v}_i \in \mathbb{R}^N$ represent the \textit{normalized eigenvector} corresponding to the eigenvalue $\lambda_i$ of the Laplacian matrix, i.e., a vector satisfying $\mathbf{v}_i^T\mathbf{v}_i = 1$ and $\lambda_i = \mathbf{v}_i^T\laplacian\mathbf{v}_i$.
The eigenvector $\connectivityEigenvector$, associated with the second smallest eigenvalue $\lambda_2$, is referred to as the ``connectivity eigenvector''.

\subsection{Gradient descent}
\begin{figure}[ht]
    \centering
    \includegraphics[width=0.9\linewidth]{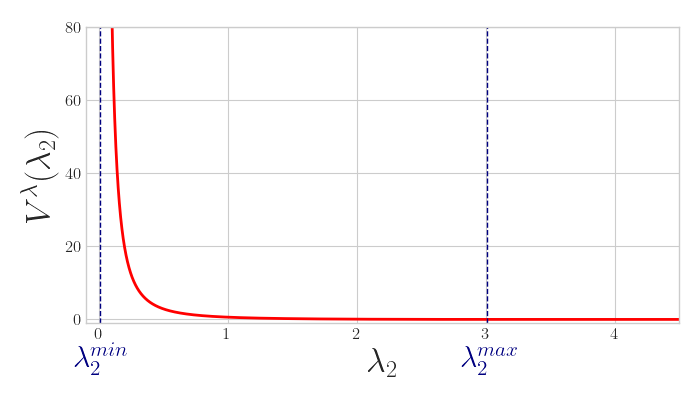}
    \caption{The shape of $V^{\lambda}(\lambda_2)$ with $\lambda^{min}_2 = 0.01$ and $\lambda^{max}_2 = 3.01$.}
    \label{fig:vlambda}
\end{figure}

The connectivity maintenance presented in~\cite{robuffo2013passivity} involves a decentralized gradient descent method applied to a cost function $V^{\lambda}(\lambda_2)\geq 0$ of $\lambda_2$, which is designed to maintain $\lambda_2$ above a positive threshold.
The cost function $V^{\lambda}(\lambda_2)$ is designed to have a vertical asymptote for a minimum connectivity value $\lambda^{min}_2>0$ and to smoothly decrease towards $0$ for $\lambda_2>\lambda^{min}_2$ (see Fig.~\ref{fig:vlambda}).

From a conceptual perspective, minimizing the cost function $V^{\lambda}(\lambda_2(\agentPos))$, where $\agentPos = [\agentPos_1^T, \agentPos_2^T, \dots, \agentPos_N^T]^T$ represents the vector of positions of all agents, with $\agentPos_i \in \mathbb{R}^3$ being the position of the $i$-th agent, can be achieved by having each agent $i$ implement a gradient-based controller:
\begin{equation}
    \connectivityForce_i(\agentPos) = -\dfrac{\partial V^{\lambda}(\lambda_2(\agentPos))}{\partial \agentPos_i}
\end{equation}
This force will be referred to as the \textit{Generalized Connectivity Force}.
It is important to note that, in general, $\connectivityForce_i(\agentPos)$ depends on the state of \textit{all agents} in the system.
This would typically require some form of centralization for its evaluation by agent $i$.
However, as we will demonstrate in the following sections, the design of $\connectivityForce_i(\agentPos)$ exhibits a decentralized structure.
Specifically, its evaluation by agent $i$ can be carried out using only local and 1-hop neighbor information.
This crucial property forms the foundation for the \textit{fully decentralized} implementation of this approach.

It is also important to note that, while following the gradient force $\connectivityForce_i(\agentPos)$, the agents are not required to maintain a \textit{fixed topology}, meaning that the edge set $\mathcal{E}$ of the interaction graph $\mathcal{G}$ does not need to remain constant. The creation and deletion of individual or multiple links, even concurrently, are fully permissible as long as the value of $\lambda_2$ is kept above the minimum threshold, $\lambda_2^{min}$.

The stability issues arising when controlling the agents' motion using the proposed force are analyzed and discussed in~\cite{robuffo2013passivity}.

\subsection{Control action}

The connectivity control action for a single integrator model $\dot{\agentPos}_i = \vv_i$ is given by:
\begin{equation}\label{eq:connectivity_control_linear}
    \connectivityVelocity_i = -\dfrac{\partial V^{\lambda}}{\partial \agentPos_i} = -\dfrac{\partial V^{\lambda}}{\partial \lambda_2}\sum_{j\in\calN_i}\dfrac{\partial A_{ij}}{\partial \p_i}(\connectivityEigenvectori-\connectivityEigenvectorj)^2
\end{equation}
where $\connectivityEigenvectori$ is the $i$-th component of the connectivity eigenvector $\connectivityEigenvector$, and $\calN_i$ is the set of neighbors of agent $i$.
If the weights $A_{ij}$ are properly designed so that they (and their gradients) can be evaluated from locally available quantities, then the connectivity maintenance law~(\ref{eq:connectivity_control_linear}) can be implemented by robot $i$ in a decentralized way.
Evaluation of~(\ref{eq:connectivity_control_linear}) requires knowledge of $\lambda_2$ and of the local components of $\connectivityEigenvector$ for robot $i$ and its neighbors.

\subsection{Decentralized estimation of \texorpdfstring{$\connectivityEigenvector$}{v2} and \texorpdfstring{$\lambda_2$}{lambda2}}
For the sake of decentralization, the estimation of $\lambda_2$ and $\connectivityEigenvector$ is performed as proposed by~\cite{yang2010decentralized} and refined in~\cite{sabattini2011decentralized, sabattini2012decentralized}.
This strategy relies on local estimation of $\lambda_2$, i.e. $\tilde{\lambda}_2$, and of the $i$-th component of $\connectivityEigenvector$, i.e. $\tilde{\mathrm{v}}_{2_i}$, for each robot $i$.
From~\cite{sabattini2011decentralized}, the estimation of $\connectivityEigenvector$ is obtained through a power iteration procedure with the following update law:
\begin{equation}\label{eq:power_iteration_centralized}
    \dot{\tilde{\mathbf{v}}}_{2} = -k_1\text{Ave}\left(\left\{\tilde{\mathrm{v}}_{2_i}\right\}\right) - k_2\laplacian\tilde{\mathbf{v}}_2  - k_3 \left(\text{Ave}\left(\left\{\tilde{\mathrm{v}}_{2_i}^2\right\}\right) - 1\right)\tilde{\mathbf{v}}_{2}
\end{equation}

where $k_1, k_2, k_3 > 0$ are the control gains, and $\text{Ave}\left(.\right)$ is the averaging operation.

To implement the update law~(\ref{eq:power_iteration_centralized}) in a decentralized way, the averaging operation is performed by each robot $i$ using a proportional integral average consensus estimator (PI-ACE), as presented in~\cite{freeman2006stability}.

\boitemagique{PI-ACE}{
Consensus estimators allow $N$ agents, each of which measures some time-varying scalar $\alpha_i(t)$, to compute an approximation of $\tilde{\alpha}(t) = \frac{1}{N}\sum_{i=1}^{N}\alpha_i(t)$ using only local communication with connected neighbors $j \in \calN_i$.
Let $\hat{y}_i$ be the average estimate of agent $i$, the PI-ACE estimator has the form:
\begin{equation}
    \dot{\tilde{y}}_i = \gamma(\alpha_i - y_i) - k_P \sum_{j\in\calN_i}(\tilde{y}_i - \tilde{y}_j) + k_I \sum_{j\in\calN_i}(\tilde{z}_i - \tilde{z}_j)
\end{equation}
\begin{equation}
    \dot{\tilde{z}}_i = - k_I \sum_{j\in\calN_i}(\tilde{y}_i - \tilde{y}_j).
\end{equation}
where $\gamma > 0$ is the rate new information replaces old information, $\calN_i$ is the set of neighbors of agent $i$, and $k_P, k_I$ are estimator gains.
}

With inputs $\alpha_{i_1} = \tilde{\mathrm{v}}_{2_i}$ and $\alpha_{i_2} = \tilde{\mathrm{v}}_{2_i}^2$, the PI average consensus estimator provides $(\tilde{y}_{i_1}, \tilde{y}_{i_2})$ as the $i$-th agent's estimate of $\left(\text{Ave}\left(\left\{\tilde{\mathrm{v}}_{2_i}\right\}\right), \text{Ave}\left(\left\{\tilde{\mathrm{v}}_{2_i}^2\right\}\right) \right)$.
Each agent receives from communication its neighbors estimates $\left(\tilde{v}_{2_j}, \tilde{y}_{i_1}, \tilde{z}_{i_1}, \tilde{y}_{i_2}, \tilde{z}_{i_2} \right)$ and implements, alongside the PI-ACE, the following decentralized update law:
\begin{equation}\label{eq:power_iteration}
    \dot{\tilde{\mathrm{v}}}_{2_i} = -k_1\tilde{y}_{i_1} - k_2\laplacian\tilde{\mathbf{v}}_2  - k_3 \left(\tilde{y}_{i_2} - 1\right)\tilde{\mathrm{v}}_{2_i}
\end{equation}
As detailed in~\cite{yang2010decentralized}, when the gain condition $k_3 > k_2\lambda_N$ is satisfied, for almost all initial condition the system \ref{eq:power_iteration_centralized} converges to an eigenvector $\tilde{\mathbf{v}}_{2}$ of $\laplacian$ corresponding to the eigenvalue $-\lambda_2$ satisfying $||\tilde{\mathbf{v}}_{2}|| = \sqrt{N\left(\frac{k_3-k_2\lambda_2}{k_3}\right)}$.
Thus, each agent can compute its own estimation of $\lambda_2$ as
\begin{equation}
    \tilde{\lambda}_{2_i} = \frac{k_3}{k_2}(1-\tilde{y}_{i_2}).
\end{equation}

\section{Conclusions}

This chapter discussed key aspects of telerobotics and multi-robot systems control.
Shared control systems leverage the complementary strengths of humans and robots, with humans providing cognitive flexibility and robots offering precision.
Haptic feedback enhances teleoperation, guiding operators with real-time force and positional cues.
Decentralized connectivity maintenance in multi-robot systems was addressed through algebraic graph theory.

Together, these concepts form the foundation for effective collaboration in remote multi-robot environments, integrating human oversight, autonomous compliance with global constraints, and efficient, intuitive feedback mechanisms

\part{Haptic Shared Control for Multi-Robot Systems}
\label{part:contributions}
\clearemptydoublepage
\graphicspath{{Chapitre3/figures/}}
\clearemptydoublepage
\chapter{Main modeling assumptions}\label{chap:modeling}
{\hypersetup{linkcolor=black}
\minitoc
}

\lettrine{I}{n} this chapter, we introduce the core modeling assumptions and definitions that form the foundation for the analysis presented in the following chapters of this thesis.
The multi-robot system under consideration consists of quadrotors operating in $\mathbb{R}^3 \times \mathbb{S}^1$, with their dynamics expressed in the body frame of each individual robot.
The chapter first outlines the dynamic model of the quadrotors, followed by a detailed description of the measurement model used to obtain relative position information. Finally, we define the sensing and communication model that governs the interaction between agents in the system.

\section{Agent model}
\label{sec:agent_model}
We consider a group of $N$ robots, focusing specifically on quadrotor UAVs due to their extensive use and popularity within the robotics community.
Nonetheless, the proposed methodology remains general and can be adapted to other types of mobile robots with minimal adjustments to account for differences in locomotion and sensing capabilities.

No global common frame is assumed directly available to the robots since they can only obtain \textit{relative measurements} w.r.t. each other and their velocity commands are in the robot body frames. A common frame must, however, be implicitly or explicitly selected for expressing the simulated robot poses. Further details on the simulation can be found in the appendix.

Robots are not assumed to have direct access to a \textit{global common reference frame}; instead, they rely solely on \textit{relative measurements} with respect to one another.
Their velocity commands are expressed in their respective body frames.
However, for simulation purposes, it is necessary to define a \textit{global fixed frame} to express the robots' poses consistently across the system.

In this thesis, we consider a world frame $\mathcal{W} : \{ \bO_{\mathcal{W}}, \X_{\mathcal{W}}, \Y_{\mathcal{W}}, \Z_{\mathcal{W}} \}$ where $\mathbf{\emph{Z}}_{\mathcal{W}}$ is aligned with the vertical (gravity) direction. The specific details of how the simulation framework handles this reference frame are provided in the appendix.

Similarly to~\cite{franchi2012modeling, schiano2016rigidity, schiano2017bearing} and other related works, we consider a group of \emph{N} ``velocity-controlled'' quadrotor UAVs with the following kinematic model
\begin{equation} \label{eq:model}
    \left( \begin{array}{c}
        \dot{{\p}}_i \\
        \dot{\psi}_i
    \end{array} \right)
    =
    \left( \begin{array}{cc}
        {\R}_i & 0 \\
        \0 & 1
    \end{array} \right)
    \left( \begin{array}{c}
        {\vv}_i \\
        \omega_i
    \end{array}
    \right),
\end{equation}
where ${\p}_i \in \mathbb{R}^3$ is the robot 3D position in $\mathcal{W}$, $\psi_i \in \textit{S}^1$ the yaw angle about $\mathbf{\emph{Z}}_{\mathcal{W}}$ with ${\R}_i = {\R}_z(\psi_i) \in \textit{SO}(3)$ the associated rotation matrix, and ${\vv}_i \in \mathbb{R}^3$ and $\omega_i \in \mathbb{R}$ represents the \emph{body-frame} linear velocity and yaw rate, which are assumed to be known and controllable.

\section{Measurement model}\label{sec:measurement_model}
\begin{figure}[htp]
    \centering
    \subfloat[\label{fig:drone_onboard_camera}Camera-like sensor with a limited FOV.]{\includegraphics[width=0.45\linewidth]{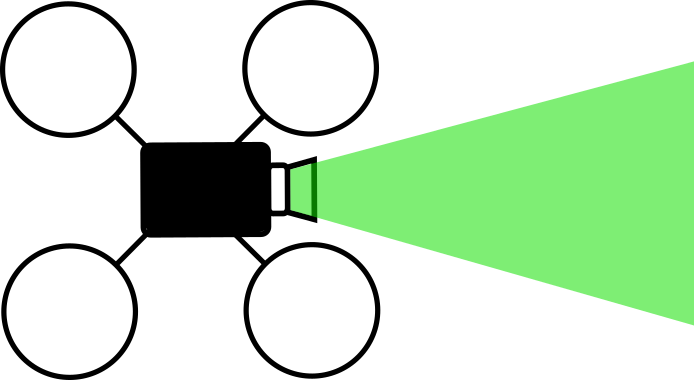}}
    \qquad
    \vrule
    \qquad
    \subfloat[\label{fig:omnidirection_sens}Distance sensor.]{\includegraphics[width=0.30\linewidth]{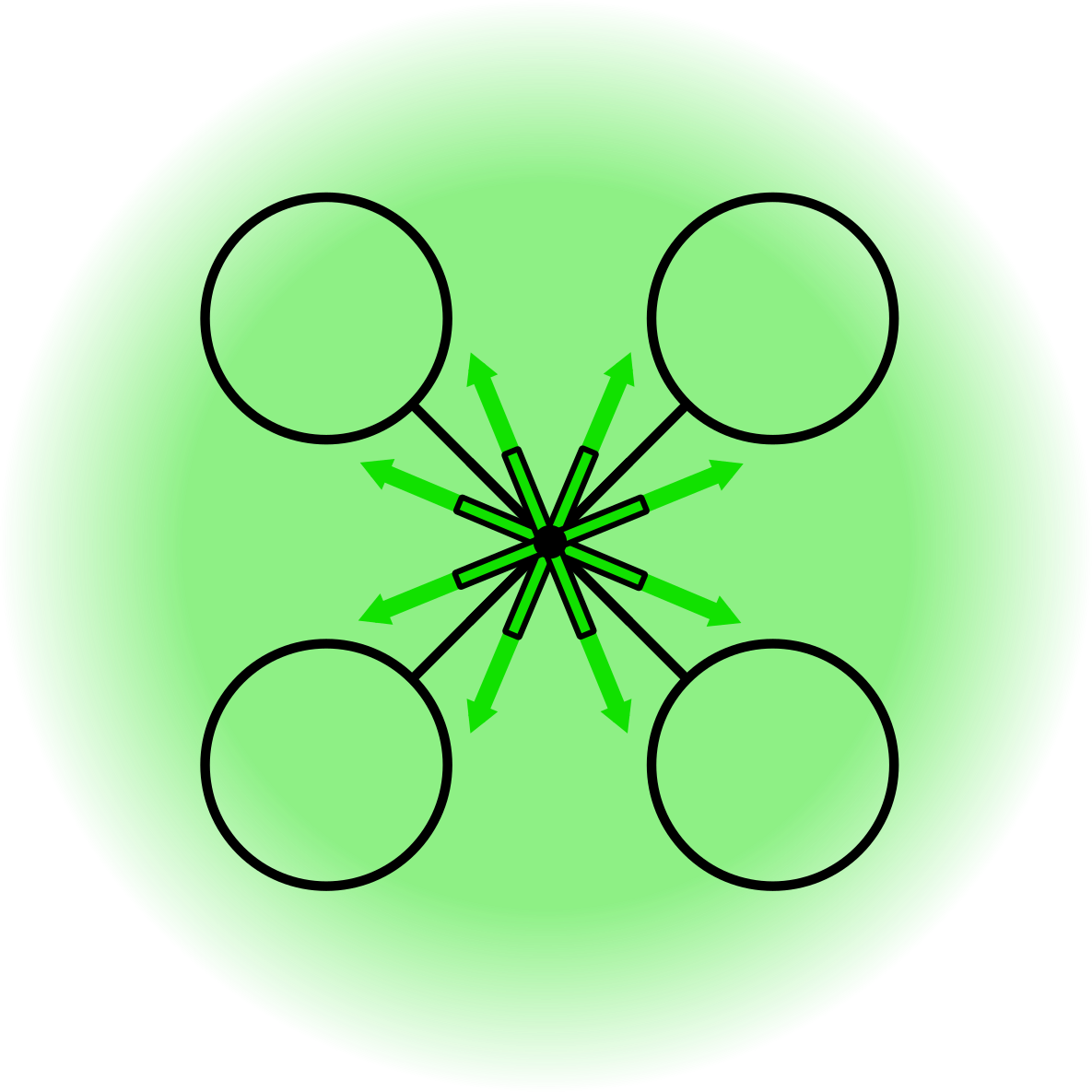}}
    \caption{\label{fig:onboard_sensors} Onboard sensors: The green areas in the figures represent the sensing directions of the respective sensors.}
\end{figure}

The robots are assumed to be equipped with two kinds of sensors: a camera-like sensor with limited FOV and an omnidirectional distance sensor (Fig.~\ref{fig:onboard_sensors}). The camera is used for detecting other UAVs in the group. In particular, when robot $j$ is in visibility of robot $i$, we assume that robot $i$ can obtain from its onboard camera a measurement of
\begin{equation}\label{eq:meas_cam}
\p_{ij} = \mathbf{R}^T_i(\p_j - \p_i),
\end{equation}
that is, the relative position of $j$ w.r.t.~$i$ expressed in the body-frame of $i$. The distance sensor is instead used for detecting any ``obstacle point'' $\p_{obs,i}$ within a given sensing range, including other robots as well.

Contrarily to most of the existing literature on connectivity-maintenance algorithms (as a small selection, see~\cite{robuffo2013passivity, gasparri2017bounded,capelli2021decentralized,aggravi2020connectivitymaintenance}), we note the following facts: we \emph{do not} assume the presence of a common frame shared by all robots (as in all previous work) where all the measurements and commands are expressed. This is evident from model~(\ref{eq:model}) (body-frame input commands) and the measurement~(\ref{eq:meas_cam}) (body-frame relative position instead of a world-frame relative position).
Furthermore, in this work, the onboard camera is modeled with a limited FOV: among others, this prevents the use of the camera for \emph{obstacle avoidance} whenever an obstacle (or another UAV) is outside the FOV. This is why a second (typically much simpler) omnidirectional distance sensor is also assumed onboard the robots. This distance sensor is assumed able to only detect ``non-free space'' within a sensing range, but with no ability to discern actual obstacles from other robots (that are then just treated as obstacles for collision avoidance purposes, see Sect.~\ref{sec:collision_detection}). The limited camera FOV also differentiates this work from the existing literature on connectivity maintenance, which instead simply assumes presence of an omnidirectional sensor able to measure relative positions w.r.t.~robots and obstacles in all directions.

\section{Communication model}\label{sec:sensing_model}

Let $\calG=(\calV,\,\calE)$ be an undirected graph with $\calV=\{1\ldots N\}$ being the vertex set and $\calE\subseteq\calV\times\calV$ the edge set. We also let $\calN_i=\{j\in\calV|\,(i,\,j)\in\calE\}$ be the set of neighbors of robot $i$. Presence of an edge $e_k=(i,\,j)$ in $\calE$ models the possibility of robots $i$ and $j$ to sense and communicate among themselves. As usual, this is obtained by associating each edge with a state-dependent weight $A_{ij}\geq 0$ such that $A_{ij}>0$ if the robots can sense/communicate and $A_{ij}=0$ otherwise. The adjacency matrix $\A\in\mathbb{R}^{N\times N}$ associated to $\calG$ is then built out of the weights $A_{ij}$.

In the context of connectivity-maintenance algorithms, in general, we consider an undirected graph with $A_{ij}=A_{ji}$ and, thus, a symmetric adjacency matrix $\A=\A^T$. This is straightforward when assuming omnidirectional (and equal) sensing capabilities for all robots as in, e.g.,~\cite{robuffo2013passivity, gasparri2017bounded,capelli2021decentralized,aggravi2020connectivitymaintenance} and similar works. In our context, however, presence of individual (and not aligned) body-frames and of the limited camera FOV  would in general result in non-symmetric weights: if $A_{ij}>0$ (robot $i$ measures robot $j$), $A_{ji}$ could be $0$ (robot $j$ does not measure $i$ because of the limited FOV). In the next Chapter we discuss how we address this issue for recovering in all cases the symmetry of the weights $A_{ij}$.

\bigskip
We are now prepared to present the \textit{core contributions} of this thesis, organized into two chapters:

\begin{itemize}
    \setlength\itemsep{1em}
    \item[$\bullet$] \textbf{Chapt.~\ref{chap:dcmFOV}} presents a solution to the problem of \textit{connectivity maintenance} in spite of sensing limitations like (\textit{i}) limited field of view of the onboard cameras, and (\textit{ii}) limited range of the onboard cameras, (\textit{iii}) measurements expressed in individual body frames. The chapter also introduces a novel method for estimating relative orientations between robots using only local measurements and communication.
    \item[$\bullet$] \textbf{Chapt.~\ref{chap:chapitre5}} presents three haptic enabled control methods for the remote operation of quadrotor UAV teams while maintaining connectivity.
    These methods are validated through a user study involving three specific scenarios, and are demonstrated in a final use case, where the operator transitions seamlessly between control modalities to adapt to the task requirements.
\end{itemize}

\clearemptydoublepage
\graphicspath{{Chapitre4/figures/}}
\clearemptydoublepage
\chapter{Decentralized Connectivity Maintenance with Field of View Constraints}\label{chap:dcmFOV}
{\hypersetup{linkcolor=black}
\minitoc
}

\lettrine{I}{n} this chapter we consider the problem of controlling a formation of quadrotor UAVs equipped with onboard cameras able to measure relative position in their local body frames with respect to neighboring UAVs.
Distributed connectivity-maintenance algorithms such as~\cite{robuffo2013passivity} provide a flexible yet intuitive solution for enabling a human user the control of a multi-robot systems while guaranteeing inter-agent communication throughout the considered mission. Connectivity maintenance has been successfully employed in applications of cooperative surveillance~\parencite{aggravi2021decentralized}, search-and-rescue~\parencite{aggravi2021hapticenabled}, and guidance~\parencite{aggravi2018design} using teams of multiple drones controlled by a human operator.
The maintenance of the team connectivity is enforced in presence of a series of sensing and interaction constraints, aiming at keeping the agents safe and able to communicate with the rest of the team.
However, despite their success and broad applicability, the vast majority of previous works on connectivity maintenance makes several strong \emph{simplifying assumptions} about the robot sensing capabilities and shared knowledge. For instance, they typically assume presence of an omnidirectional sensor for retrieving relative positions of neighboring robots, which are also assumed expressed in a \emph{single common frame} shared by the group. These assumptions simplify the control design but are not very realistic, since most onboard sensors have a limited field of view (cameras being a prominent example) and, in addition, every robot collects measurements and expresses command actions in its own \emph{body-frame}, which in general does not need to be aligned with the body frames of the other robots.

The goal of this chapter is to then propose a decentralized connectivity-maintenance algorithm able to overcome the two aforementioned issues and explicitly take into account presence of a limited FOV and lack of a common shared frame among the robots.
For implementing the connectivity-maintenance action under these new constraints, we also develop a novel decentralized estimator of the relative orientation among neighboring robots, which is needed by the proposed scheme. Finally, we validate the proposed approach in a realistic simulation scenario employing a team of $6$ quadrotors UAVs.

This work has been published in~\cite{bernard2023decentralized}.

The rest of the chapter is structured as follows.
Section~\ref{sec:conn_maint} details the proposed decentralized connectivity maintenance strategy by taking into account the FOV constraints.
Section~\ref{sec:psi_estimation} illustrates the decentralized estimator of the relative orientation among the robot body-frames.
Section~\ref{sec:conn_maint_results} then validates the approach in a realistic simulation involving $6$ quadrotor UAVS, and Section~\ref{sec:fov_conclusions} finally concludes the chapter.

\section{Decentralized connectivity maintenance}\label{sec:conn_maint}

We briefly recall the connectivity-maintenance algorithm presented in~\cite{robuffo2013passivity} and introduced in Sect.~\ref{section:connectivityMaintenanceIntro} and then discuss its extension to the setting considered in this work.
The connectivity maintenance presented in~\cite{robuffo2013passivity} consists of a decentralized gradient descent of a cost function $V^{\lambda}(\lambda_2)\geq 0$ of the so-called \emph{connectivity eigenvalue} $\lambda_2$, the second smallest eigenvalue of the Laplacian matrix $\boldsymbol{L}=\diag(\delta_i)-\A$ associated to the graph, where $\delta_i=\sum^N_{j=1}A_{ij}$.
The cost function $V^{\lambda}(\lambda_2)$ is designed to have a vertical asymptote for a minimum connectivity value $\lambda^{min}_2$ and to smoothly decrease towards $0$ for $\lambda_2>\lambda^{min}_2$ (see Fig.~\ref{fig:vlambda}). Assume that the edge weights $A_{ij}$ are smooth enough functions of the state with some (mild) locality properties (in short, a weight $A_{ij}$ may only depend on the state of robot $i$ and of its neighbors).

The connectivity control action for a single integrator model $\dot \x_i=\boldsymbol{u}_i$ is the gradient descent
\begin{equation}\label{eq:conn_ctrl}
\boldsymbol{u}_i = -\dfrac{\partial V^{\lambda}}{\partial \x_i} = -\dfrac{\partial V^{\lambda}}{\partial \lambda_2}\sum_{j\in\calN_i}\dfrac{\partial A_{ij}}{\partial \x_i}(\connectivityEigenvectori-\connectivityEigenvectorj)^2
\end{equation}
where $\connectivityEigenvectori$ is the $i$-th component of the eigenvector $\connectivityEigenvector$ associated to the eigenvalue $\lambda_2$. Evaluation of~(\ref{eq:conn_ctrl}) requires knowledge of $\lambda_2$ and of the local components of $\connectivityEigenvector$ (of $i$ and its neighbors). These can be estimated online in a decentralized way as explained in~\cite{robuffo2013passivity}. If the weights $A_{ij}$ are properly designed so that they (and their gradients) can be evaluated from locally available quantities, then the law~(\ref{eq:conn_ctrl}) can be implemented by robot $i$ in a decentralized way.

The robot model~(\ref{eq:model}) considered in this work is, however, not a simple single integrator. Therefore, the gradient descent~(\ref{eq:conn_ctrl}) needs to be adjusted accordingly. To this end, let us consider the time derivative of the cost $V^{\lambda}$:
\begin{equation}\label{eq:dot_V_lambda}
\begin{split}
\dot{V}^{\lambda}&=\dfrac{\partial V^{\lambda}}{\partial \lambda_2}\dot{\lambda}_2\\
&=\dfrac{\partial V^{\lambda}}{\partial \lambda_2}\sum^N_{i=1}\sum_{j\in\calN_i}\left(\dfrac{\partial^T A_{ij}}{\partial \p_i}\dot\p_i+\dfrac{\partial A_{ij}}{\partial \psi_i}\dot\psi_i\right)(\connectivityEigenvectori-\connectivityEigenvectorj)^2\\
&=\dfrac{\partial V^{\lambda}}{\partial \lambda_2}\sum^N_{i=1}\sum_{j\in\calN_i}\left(\dfrac{\partial^T A_{ij}}{\partial \p_i}{\R}_i\vv_i+\dfrac{\partial A_{ij}}{\partial \psi_i} \omega_i\right)(\connectivityEigenvectori-\connectivityEigenvectorj)^2
\end{split}
\end{equation}
where we used~(\ref{eq:model}).
By inspection, the $i$-th robot body-frame velocity commands for implementing a gradient descent of $V^{\lambda}$ are then
\begin{equation}\label{eq:grad_body}
\left\{
\begin{array}{rcl}
\vv_i &=& -\dfrac{\partial V^{\lambda}}{\partial \lambda_2}\sum_{j\in\calN_i}\R^T_i\dfrac{\partial A_{ij}}{\partial \p_i}(\connectivityEigenvectori-\connectivityEigenvectorj)^2 \vspace{5pt} \\
\omega_i&=& -\dfrac{\partial V^{\lambda}}{\partial \lambda_2}\sum_{j\in\calN_i}\dfrac{\partial A_{ij}}{\partial \psi_i}(\connectivityEigenvectori-\connectivityEigenvectorj)^2
\end{array}
\right..
\end{equation}

The upcoming sections detail the design of the interaction weights $A_{ij}$ based on the chosen sensing model and assess the feasibility of implementing the gradient control law~(\ref{eq:grad_body}) using only the quantities assumed available to robot $i$.

\subsection{Design of the weights}\label{sec:weight_design}

\begin{figure}[htp]
\centering
\subfloat[][\centering $w^0_1(x,\,x^d,\,x^D)$\par with $x^d=0.2$ and $x^D=0.5$]{\includegraphics[width=0.45\linewidth]{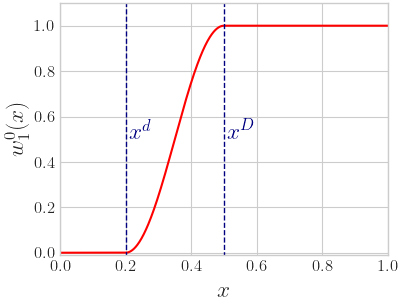}}%
\qquad
\subfloat[][\centering $w^1_0(x,\,x^d,\,x^D)$\par with $x^d=0.5$ and $x^D=0.8$]{\includegraphics[width=0.45\linewidth]{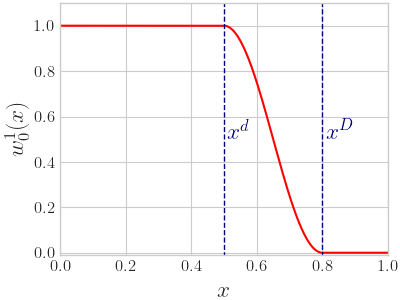}}
\caption{The shape of the two considered rising/falling activation functions.}%
\label{fig:w01w10}%
\end{figure}

As discussed in Sect.~\ref{sec:agent_model}, we consider the presence of an onboard camera with limited FOV and of a (simpler) omnidirectional distance sensor.
The camera has a minimum and maximum detection range, while the distance sensor has only a maximum range value to consider. For simplicity, we assume that the maximum range of the camera equals the maximum communication range of the robots. The weight $A_{ij}$ is then designed as the product of two sub-weights
\begin{equation}\label{eq:A_ij}
A_{ij}=\gamma_{ij}f_{ij},
\end{equation}
where $\gamma_{ij}$ accounts for the min/max detection range and $f_{ij}$ for the camera FOV. Similarly to~\cite{aggravi2020connectivitymaintenance}, let us introduce two (classical) rising/falling activation functions
\begin{equation}\label{eq:rising_fun}
w^0_1(x,\,x^d,\,x^D)=
\left\{
\begin{array}{cc}
 0 & x\leq x^d  \\
\dfrac{1}{2}(1-\cos(ax+b))&  x^d < x \leq x^D \\
1 & x^D<x
\end{array}
\right.
\end{equation}
and
\begin{equation}\label{eq:descending_fun}
w^1_0(x,\,x^d,\,x^D)=
\left\{
\begin{array}{cc}
 1 & x\leq x^d \\
\dfrac{1}{2}(1+\cos(ax+b))&  x^d < x \leq x^D \\
0 & x^D<x
\end{array}
\right.
\end{equation}
with $a=\pi/(x^D-x^d)$ and $b=-ax^d$, and lower/upper thresholds set at $x^d$ and $x^D$ (see Fig.~\ref{fig:w01w10} for an example).

Weight $\gamma_{ij}$ is then defined as
\begin{equation}\label{eq:gamma_ij}
\gamma_{ij}(d_{ij})=w^0_1(d_{ij},\,d_{min},\,D_1)w^1_0(d_{ij},\,D_2,\,d_{max}),
\end{equation}
where $d_{ij}=\|\p_{ij}\|$ is the distance among the robots and $d_{min}<D_1<D_2<d_{max}$ two suitable thresholds representing the min/max detection ranges.

As for the FOV weight, we define $\bbeta_{ij} = \frac{\p_{ij}}{d_{ij}}$ as the (unit-norm) bearing vector from $i$ to $j$ in the body-frame of robot $i$ (note that $\bbeta_{ij}$ can be directly computed from the measured $\p_{ij}$). Let also $\mathbf{o}_C$ be the fixed body-frame direction of the camera optical axis. The quantity $a_{ij}=\mathbf{o}^T_C\bbeta_{ij}$ (the cosine of the angle between $\bbeta_{ij}$ and $\mathbf{o}_C$) can therefore be leveraged for defining the FOV weight. Towards this end, we start by defining the function
\begin{equation}\label{eq:f_ij^*}
f^*_{ij}(a_{ij})=w^0_1(a_{ij},\,a_{min},\,a_{max}),
\end{equation}
where $a_{max}$ is the cosine of the FOV limit (expressed as a maximum angular displacement from $\mathbf{o}_C$). Although tempting, function $f^*_{ij}$ cannot be directly used as the sought FOV weight, as it is not symmetric, i.e., in general, $f^*_{ij}\neq f^*_{ji}$. We then propose to define the FOV weight as
\begin{equation}\label{eq:f_ij}
f_{ij}=f^*_{ij} + f^*_{ji} - f^*_{ij}f^*_{ji}.
\end{equation}
This definition has two advantages: first of all, $f_{ij}$ is now symmetric, as it can be easily verified. Furthermore, $f_{ij}=1$ whenever $f^*_{ij}=1$ or $f^*_{ji}=1$ (or both), and $f_{ij}\to 0$ only when both $f^*_{ij}\to0$ and $f^*_{ji}\to 0$. In other words, the FOV weight $f_{ij}$ \emph{will not force} robots $i$ and $j$ to simultaneously look at each other; instead, it will only enforce that at least one of the robots keeps the other one in sight. This is, in our opinion, an important feature for not \emph{overconstraining} the robot group during motion.

\subsection{Weight gradient}\label{sec:weight_gradients}

We detail here the computation of the weight gradients.
From~(\ref{eq:A_ij}), one clearly has
\begin{equation}\label{eq:grad_terms_v_i_nd}
\frac{\partial A_{ij}}{\partial \p_i}=f_{ij}\frac{\partial \gamma_{ij}}{\partial \p_i}+\gamma_{ij}\frac{\partial f_{ij}}{\partial \p_i}
\end{equation}
and
\begin{equation}\label{eq:grad_terms_w_i_nd}
\frac{\partial A_{ij}}{\partial \psi_i}=f_{ij}\frac{\partial \gamma_{ij}}{\partial \psi_i}+\gamma_{ij}\frac{\partial f_{ij}}{\partial \psi_i} ,
\end{equation}
since $\gamma_{ij}$ is only a function of the distance $d_{ij}$, it follows that
\begin{equation}\label{eq:grad_gamma_ij}
\dfrac{\partial \gamma_{ij}}{\partial \p_i}=-\dfrac{\partial \gamma_{ij}}{\partial d_{ij}}\dfrac{\p_j-\p_i}{d_{ij}}
\end{equation}
and ${\partial \gamma_{ij}}/{\partial \psi_i}=0$.

As for ${\partial f_{ij}}/{\partial \p_i}$, from~(\ref{eq:f_ij}) one has
\begin{equation}
\dfrac{\partial f_{ij}}{\partial \p_i}=(1-f^*_{ji})\dfrac{\partial f^*_{ij}}{\partial \p_i}+(1-f^*_{ij})\dfrac{\partial f^*_{ji}}{\partial \p_i}.
\end{equation}
and analogously for ${\partial f_{ij}}/{\partial \psi_i}$.

Recalling the definition~(\ref{eq:f_ij^*}) and that $a_{ij}=\mathbf{o}^T_C\bbeta_{ij}$, one obtains
\begin{equation}
\dfrac{\partial f^*_{ij}}{\partial \p_i}=\dfrac{\partial f^*_{ij}}{\partial a_{ij}}\dfrac{\partial^T \bbeta_{ij}}{\partial \p_i}\mathbf{o}_C=-\dfrac{\partial f^*_{ij}}{\partial a_{ij}}\dfrac{\R_i\P_{ij}}{d_{ij}}\mathbf{o}_C
\end{equation}
and
\begin{equation}
\dfrac{\partial f^*_{ji}}{\partial \p_i}=\dfrac{\partial f^*_{ji}}{\partial a_{ji}}\dfrac{\partial^T \bbeta_{ji}}{\partial \p_i}\mathbf{o}_C=\dfrac{\partial f^*_{ji}}{\partial a_{ji}}\dfrac{\R_j\P_{ji}}{d_{ij}}\mathbf{o}_C,
\end{equation}
where $\P_{ij}=\I_3-\bbeta_{ij}\bbeta^T_{ij}$. Further details about this derivation can be found in~\cite{zelazo2015bearing,schiano2016rigidity}. Similar considerations also yield
\begin{equation}
\dfrac{\partial f^*_{ij}}{\partial \psi_i}=-\dfrac{\partial f^*_{ij}}{\partial a_{ij}}\mathbf{o}^T_C\S\bbeta_{ij}
\end{equation}
and ${\partial f^*_{ji}}/{\partial \psi_i}=0$ with $\S=[[0\;0\;1]^T]_{\times}$ and $[\cdot]_{\times}$ being the usual skew-symmetric matrix operator.
Therefore, we can conclude that the structure of the gradient of weights $A_{ij}$~(\ref{eq:A_ij}) needed for implementing~(\ref{eq:grad_body}) is as follows

\begin{equation}\label{eq:grad_terms_v_i}
    \R^T_i\dfrac{\partial A_{ij}}{\partial\p_i} =-f_{ij}\dfrac{\partial \gamma_{ij}}{\partial d_{ij}}\bbeta_{ij}+\gamma_{ij}\left( -(1-f^*_{ji})\dfrac{\partial f^*_{ij}}{\partial a_{ij}}\dfrac{\P_{ij}}{d_{ij}}\mathbf{o}_C \right. \left. + (1-f^*_{ij})\dfrac{\partial f^*_{ji}}{\partial a_{ji}} {^{i}\R_j}\dfrac{\P_{ji}}{d_{ij}}\mathbf{o}_C   \right)
\end{equation}
and
\begin{equation}\label{eq:grad_terms_w_i}
\dfrac{\partial A_{ij}}{\partial\psi_i}  =-\gamma_{ij}(1-f^*_{ji})\dfrac{\partial f^*_{ij}}{\partial a_{ij}}\mathbf{o}^T_C\S\bbeta_{ij},
\end{equation}
where ${^{i}\R_j}=\R^T_i\R_j=\R_z(\psi_j-\psi_i)$ depends on the relative rotation among the body frames of robot $j$ and $i$.

\subsection{Discussion}

We now discuss the maintenance controller~(\ref{eq:grad_body})--(\ref{eq:grad_terms_v_i}--\ref{eq:grad_terms_w_i}) in terms of the quantities actually needed by robot $i$ for implementing it. From inspection of the various terms, one can conclude that evaluation of~(\ref{eq:grad_terms_v_i}) requires knowledge of $d_{ij}$, $\bbeta_{ij}$, the relative angle $\psi_{ij}=\psi_j-\psi_i$ (for evaluating ${^{i}\R_j}$), and $\bbeta_{ji}=-{^{i}\R_j}\bbeta_{ij}$. Evaluation of~(\ref{eq:grad_terms_w_i}) requires again knowledge of $d_{ij}$, $\bbeta_{ij}$, and $\bbeta_{ji}$.
Sect.~\ref{sec:psi_estimation} will discuss how an estimation of the relative angle $\psi_{ij}$ can be maintained by robot $i$ during motion. %

We then distinguish the following cases:
\begin{enumerate}
    \item robot $i$ can measure robot $j$: in this case, robot $i$ has access to all the needed quantities, since $d_{ij}$, $\bbeta_{ij}$ and $\bbeta_{ji}$ can be obtained from the measured $\p_{ij}$ and the estimated angle $\psi_{ij}$;
    \item robot $i$ cannot measure robot $j$ but it is measured by robot $j$: in this case, robot $j$ needs to communicate to robot $i$ the measured $\p_{ji}$ and estimated $\psi_{ji}$ from which robot $i$ can again compute all the needed quantities;
    \item robot $i$ and robot $j$ do not measure themselves: in this case the terms $\partial A_{ij}/\partial \p_i=0$ and $\partial A_{ij}/\partial \psi_i=0$ (and alike for robot $j$).
\end{enumerate}
Concerning the symmetry of weights $A_{ij}$ in~(\ref{eq:A_ij}), we note that the weight $f_{ij}=f_{ji}$ is symmetric by construction, thanks to the definition~(\ref{eq:f_ij}), and that $\gamma_{ij}=\gamma_{ji}$ since $d_{ij}=d_{ji}$.
The only remark is that, in the case 2. listed above, robot $i$ still needs to compute the weight $\gamma_{ij}$, even if robot $j$ is not in visibility, so as to ensure symmetry of the total weight $A_{ij}$.
This can of course be done using the quantity $\p_{ji}$ received from robot $j$.

In summary, the weights $A_{ij}$ can be made symmetric and the maintenance controller~(\ref{eq:grad_body})--(\ref{eq:grad_terms_v_i}--\ref{eq:grad_terms_w_i}) can be implemented by the robots in the group with the requirement that, if a robot $i$ measures another robot $j$, then it needs to communicate its measured relative position $\p_{ij}$ and current estimation of $\psi_{ij}$ to robot $j$. Since this amount of information is constant per neighbor, we can conclude that the overall maintenance scheme is decentralized and scalable.

\subsection{Collision Avoidance}\label{sec:collision_detection}

Collision avoidance with obstacles or other robots is dealt with in a simple yet practical way. Let $\oo^k_{i}$ be the position of a $k$-th obstacle point in the frame of robot $i$, detected by the onboard sensor. For each obstacle point inside the sensing range $d_{max}$, we define a repulsive velocity command
$
\vv^k_{i,obs} = -k_{obs}(\|\oo^k_{i}\|)\oo^k_{i}
$
where the gain $k_{obs}(\|\oo^k_{i}\|)\geq 0$ is any monotonic function with a vertical asymptote at $\|\oo^k_{i}\|=d^{obs}_{min}$ and vanishing (with vanishing slope) for $\|\oo^k_{i}\|\to d^{obs}_{max}$, where $0<d^{obs}_{min}<d^{obs}_{max}$ is a safety distance. The $i$-th robot velocity command $\vv_i$ from~(\ref{eq:grad_body}) is then augmented with the additional term $\vv_{i,obs}=\sum_{k}\vv^k_{i,obs}$ for all detected obstacle points $\oo^k_{i}$. This  simple reactive strategy is sufficient to handle collision avoidance in most practical situations. Stability of the coupling between connectivity maintenance and obstacle repulsion can be shown by following the same arguments as in~\cite{lee2013semiautonomous}: in short, for any bounded external action (in our case, a bounded external command from the human operator), the closed-loop system is stable besides practically improbable (zero-measure) situations of perfect (and unlikely) alignments among the robots.

\section{Estimation of the relative orientation} \label{sec:psi_estimation}

As explained in the previous section, for implementing the maintenance controller, each robot $i$ needs to know the relative orientation $\psi_{ij}=\psi_j-\psi_i$ w.r.t.~the other robots it is interacting with. This is a typical requirement common to previous works sharing the same robot model~(\ref{eq:model}), see, e.g.,~\cite{franchi2012modeling,schiano2016rigidity,schiano2017bearing} and references therein. %
In general, in these prior works the problem of obtaining $\psi_{ij}$ is addressed by assuming special structures of the interaction graph $\calG$. in~\cite{franchi2012modeling}, any interacting pair of robots is always assumed in mutual visibility so that both $\bbeta_{ij}$ and $\bbeta_{ji}$ are available. This, as well-known, allows for an algebraic computation of $\psi_{ij}$. in~\cite{schiano2016rigidity,schiano2017bearing}, constant mutual visibility is not required, but the graph is assumed to be \emph{(bearing) rigid}. The rigidity condition allows implementing a simple decentralized localization algorithm that allows estimating $\psi_{ij}$ from the measured bearings. These solutions however cannot be used in the setting considered in this work as we \emph{do not} require mutual visibility nor maintenance of graph rigidity (which is a much stronger requirement than just graph connectivity). Therefore, we detail here an alternative (and, to our knowledge, novel) scheme for retrieving an estimation of $\psi_{ij}$.

\boitesimple{
Note that we do not (purposely) assume the relative yaw angle $\psi_{ij}=\psi_j-\psi_i$ to be directly measurable from the onboard camera: indeed, while extracting a relative position is (relatively) feasible, obtaining a relative orientation information from vision is in practice much harder because of the typical rotational symmetries of the UAVs. Therefore we consider $\p_{ij}$ as the only information that can be reliably extracted from the images. Also, while from the onboard IMUs one could obtain a (noisy) measurement of $\dot{\psi}_{ij}$, a plain forward integration of $\dot{\psi}_{ij}$ would clearly quickly drift over time.
These considerations then motivate the estimation algorithm described hereafter.
}

\begin{figure}[htp]
    \centering
    \subfloat[\label{fig:relatVjobs}]{\includegraphics[width=0.50\linewidth]{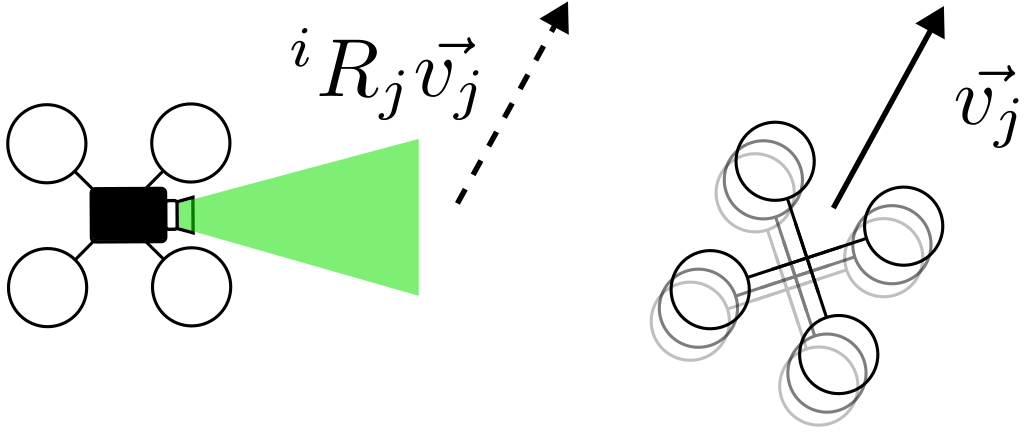}}
    \qquad
    \vrule
    \qquad
    \subfloat[\label{fig:relatVjRepair}]{\includegraphics[width=0.30\linewidth]{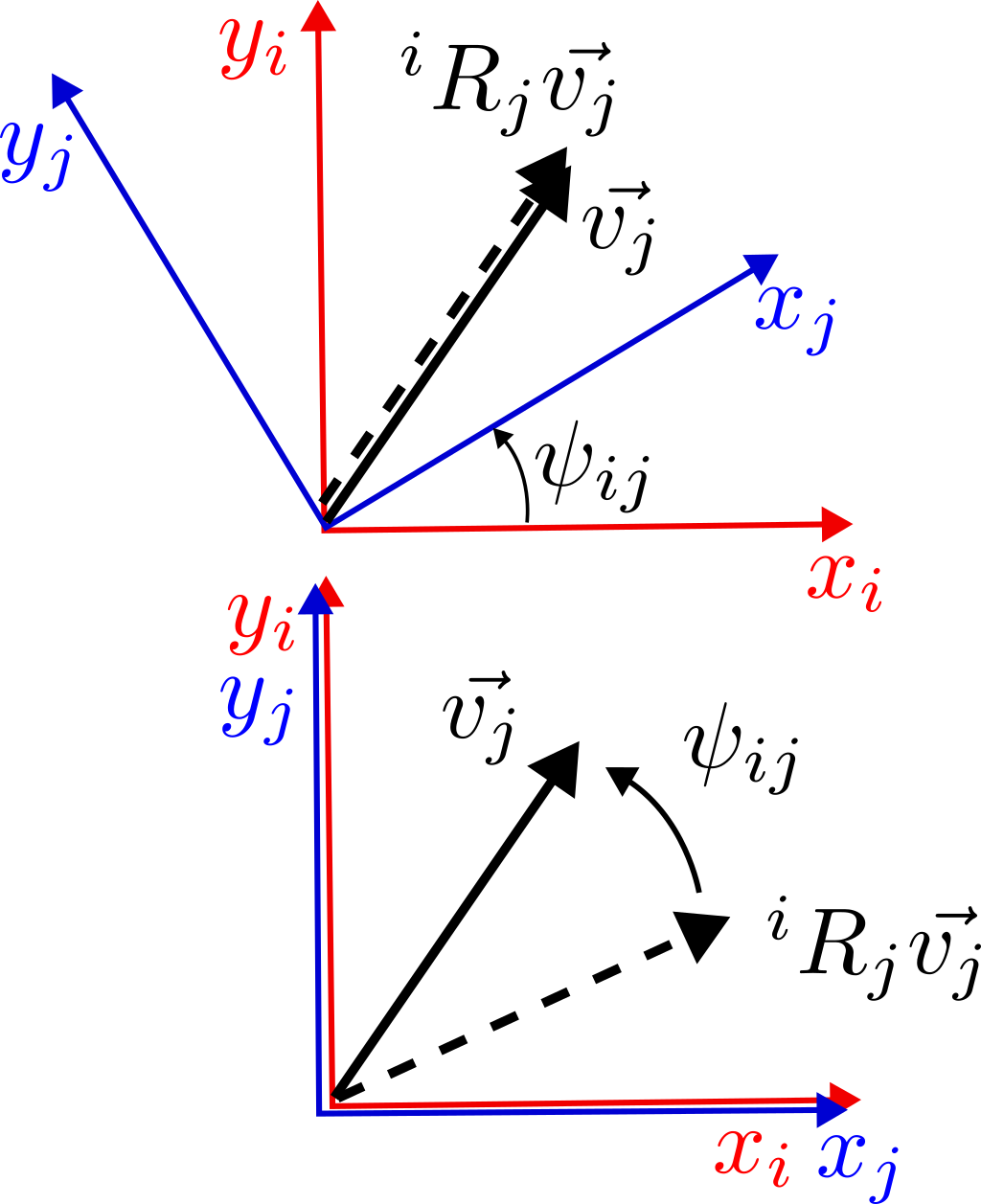}}
    \caption{\label{fig:relateVj} Relative orientation estimation. Fig.~\ref{fig:relatVjobs} Robot $i$ senses robot $j$ and measures its velocity. Fig.~\ref{fig:relatVjRepair} The angle between ${}^i\textbf{R}_j{\vv}_j$ and $\vv_j$ on the $XY$-plane is the relative orientation $\psi_{ij}$.}%
\end{figure}

Our initial approach involves observing the velocity of a neighboring robot $j$ in the body-frame of robot $i$ and comparing it to the velocity of robot $j$ in its own body frame.
As illustrated in Fig.~\ref{fig:relateVj}, the relative orientation is determined as the angle between these two velocity vectors when projected onto the $XY$-plane.

We start by considering the dynamics of the measurement~(\ref{eq:meas_cam})
\begin{equation}\label{eq:bearing_dynamics}
    \dot{\p}_{ij} =
    - \vv_i
    - \textbf{S}\p_{ij} \omega_i %
    + \textbf{}^i\textbf{R}_j{\vv}_j.
\end{equation}
We note that in~(\ref{eq:bearing_dynamics}) the only unknown term is $\textbf{}^i\textbf{R}_j{\vv}_j$, since $\vv_i$, $\omega_i$, and $\p_{ij}$ are assumed available to robot $i$. This motivates to consider the use of a ``disturbance observer'' for estimating the unknown $\textbf{}^i\textbf{R}_j{\vv}_j$, seen as a disturbance acting on the dynamics of $\p_{ij}$. In particular, let $\hat{\p}_{ij}$ be an estimation of $\p_{ij}$ maintained by robot $i$ and updated with the following law
\begin{equation}\label{eq:hat_bearing_dynamics}
    \dot{\hat{\p}}_{ij} =
    - {\vv}_i
    - \textbf{S}\p_{ij} \omega_i %
    + k_1 \left( \p_{ij} - \hat{\p}_{ij} \right)
\end{equation}
with $k_1>0$. Let us also define the quantity
\begin{equation}\label{eq:p_ij_error}
    \e = k_1(\p_{ij} - \hat{\p}_{ij})
\end{equation}
which, using~(\ref{eq:bearing_dynamics}--\ref{eq:hat_bearing_dynamics}), has dynamics
\begin{equation}\label{eq:p_ij_error_dynamics}
    \dot{\e} = k_1(\textbf{}^i\textbf{R}_j{\vv}_j -  \e).
\end{equation}
In Laplace domain, this is equivalent to
\begin{equation}\label{eq:p_ij_error_dynamics_Laplace}
    {\e}(s) = \dfrac{k_1}{s+k_1}\textbf{}^i\textbf{R}_j{\vv}_j.
\end{equation}
Therefore, the quantity ${\e}(t)$ (which can be evaluated by robot $i$) acts as a first-order low-pass filtered version of the unknown ${}^i\textbf{R}_j{\vv}_j$ with a cut-off frequency given by gain $k_1$. Robot $i$ can then (algebraically) obtain an estimation $\hat\psi_{ij}$ by computing the angle around the vertical axis among the vectors $\e$ (approximating $\textbf{}^i\textbf{R}_j{\vv}_j$) and $\vv_j$ (communicated by robot $j$).

This estimation strategy works reasonably well, but it has some shortcomings. First of all, the estimated $\hat\psi_{ij}$ is reliable only if $\vv_j\neq \0$ and if $\vv_j$ is not aligned with the vertical direction (since, in this case, ${}^i\textbf{R}_j{\vv}_j=\vv_j$). This latter situation is, in practice, very rare, but the former situation ($\vv_j\approx \0$) can frequently occur during the group  motion. Second, the estimation dynamics~(\ref{eq:hat_bearing_dynamics}) does not exploit knowledge of the angular velocity  $\omega_j$, which is instead an available quantity. Third, when robots $i$ and $j$ are in mutual visibility, the relative angle $\psi_{ij}$ can be algebraically computed from the two measured relative positions $\p_{ij}$ and $\p_{ji}$, see, e.g.,~\cite{franchi2012modeling}. The possible availability of a direct measurement of $\psi_{ij}$, although possibly noisy, is also not exploited by~(\ref{eq:hat_bearing_dynamics}).

To address these points, we propose an improved estimation strategy based on the previous one.
Let $\tilde \psi_{ij}$ be the state of the second estimator maintained by robot $i$. Since $\dot{\psi}_{ij}=\omega_j - \omega_i$, we choose the following estimation dynamics
\begin{equation}\label{eq:odometry_estimation}
    \dot{\tilde{\psi}}_{ij} = \omega_j - \omega_i + k_3f^*_{ij}f^*_{ji}\sin \left({\psi}_{ij} - \tilde{\psi}_{ij}\right) + k_2(\|\vv_j\|)(1-f^*_{ij}f^*_{ji}) \sin \left(\hat{\psi}_{ij} -  \tilde{\psi}_{ij}\right)
\end{equation}
where $k_2(\|\vv_j\|)=\bar{k}_2 w^0_1(\|\vv_j\|,\,v_{min},\,v_{max})$, $0<v_{min}<v_{max}$ are lower/upper threshold for the norm of $\vv_j$ and $\bar{k}_2>0$ and $k_3>0$ are estimation gains. The rationale behind~(\ref{eq:odometry_estimation}) is as follows: if robots $i$ and $j$ are in mutual visibility, then $f^*_{ij}f^*_{ji}=1$ and~(\ref{eq:odometry_estimation}) reduces to
\begin{equation}\label{eq:dot_psi_case_1}
  \dot{\tilde{\psi}}_{ij} = \omega_j - \omega_i + k_3\sin \left({\psi}_{ij} -  \tilde{\psi}_{ij}\right),
\end{equation}
i.e., a forward odometry propagation  with a correction term on the ``measured'' (but possibly noisy) relative angle $\psi_{ij}$ (the use of the $\sin(\cdot)$ allows dealing with possible jumps of $\pm 2\pi$ in $\psi_{ij}$). Note that, in this case, both robots $i$ and $j$ will implement their own instance of~(\ref{eq:dot_psi_case_1}): they can then exchange their estimated $\tilde{\psi}_{ij}$ and $\tilde{\psi}_{ji}$ for obtaining an average of the two values (exploiting the fact that ${\psi}_{ij}=-{\psi}_{ji}$).\\  If, instead, the two robots are not in mutual visibility but robot $i$ can measure robot $j$ (the other case being symmetrical), then $f^*_{ij}f^*_{ji}=0$ and~(\ref{eq:odometry_estimation}) reduces to
\begin{equation}
  \dot{\tilde{\psi}}_{ij} = \omega_j - \omega_i + k_2(\|\vv_j\|) \sin \left(\hat{\psi}_{ij} -  \tilde{\psi}_{ij}\right),
\end{equation}
which is again a forward odometry propagation but now using the estimated $\hat \psi_{ij}$ in the correction term. %
Since, as explained above, the reliability of the estimated $\hat \psi_{ij}$ decreases with the norm of $\vv_j$, the gain $k_2$ is designed so as to vanish for a $\|\vv_j\|\to v_{min}$ so that, when $\|\vv_j\|$ is too small,~(\ref{eq:odometry_estimation}) further reduces to $\dot{\tilde{\psi}}_{ij} = \omega_j - \omega_i$.

Robot $i$ can then communicate $\tilde{\psi}_{ji}=-\tilde{\psi}_{ij}$ to robot $j$, who needs this information for implementing the maintenance controller. Finally, in any intermediate case, i.e., when $0<f^*_{ij}f^*_{ji}<1$, the estimator~(\ref{eq:odometry_estimation}) will blend the two cases in a continuous way.

\section{Simulation results} \label{sec:conn_maint_results}

This section presents the results of a simulation in which $N=6$ UAVs were controlled by the connectivity-maintenance algorithm and obstacle avoidance actions discussed in the previous sections.\footnote{Experiments presented in this chapter were carried out thanks to a platform of the Robotex 2.0 French research infrastructure.} We implemented the proposed control and estimation algorithms using decentralized ROS~2 instances, a custom Unity scene to visualize the UAV behavior, and $robotpkg$ from LAAS' Openrobots project~\parencite{mallet2007reusable} to simulate their dynamics\footnote{Note that the employed simulation environment simulates the full 3D quadrotor dynamics {in presence of actuation noise} and not the simplified model~(\ref{eq:model}) used only for control design. Therefore, the reported results are also meant to validate the portability of the proposed approach to a more realistic case closer to real-world conditions.}.
The parameters used in the algorithm are reported in Table~\ref{tab:exp_params}.

\renewcommand*{\arraystretch}{1.5}
\begin{table}[ht]
    \centering
    \vspace{0.8cm}
    \begin{tabular}{|c|c|c|c|}
    \hline
        $d_{min} = 1~m$ & $a_{min} = 0.4$                      & $d^{obs}_{min} = 1.0~m$ & $k_1 = 10$\\
        $D_1 = 3~m$     &  $a_{max} = 0.95$                    & $d^{obs}_{max} = 3.5~m$ & $\bar{k}_2 = 10$\\
        $D_2 = 7~m$     &  $\mathbf{o}_C = \dfrac{[1, 1, 0]}{\sqrt{2}}$ & $\lambda^{min}_2 = 0.01$ & $k_3 = 10$\\
        $d_{max} = 9~m$ &                                      & $\lambda^{max}_2 = 3.01$ &\\
        \hline
    \end{tabular}
    \caption{Algorithm parameters used in our validation.}
    \label{tab:exp_params}
\end{table}

\begin{figure}[htp]
    \centering
    \includegraphics[width=0.8\linewidth]{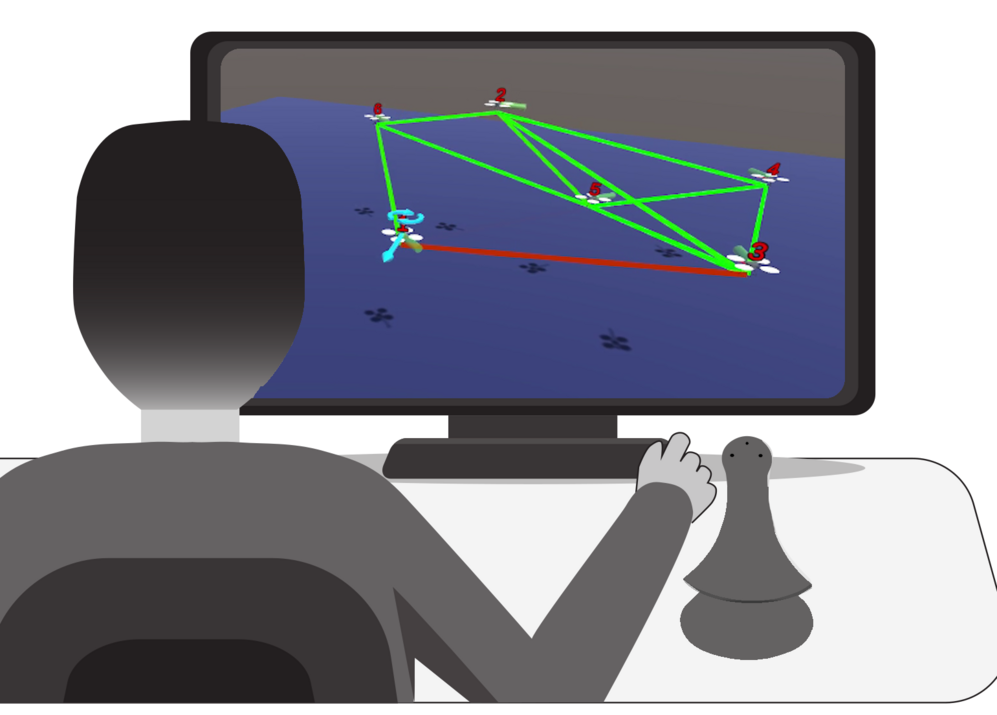}
    \caption{Experimental setup for simulation with Connectivity Maintenance.}
    \label{fig:exp_presentation}
\end{figure}

A human operator was asked to control the multi-robot team in a 3-dimensional space to reach a predefined target, employing the ``leader-follower'' paradigm presented in~\cite{robuffo2013passivity}.
Using an analog joystick, the human operator imparts an additional input velocity to one robot (the leader) for controlling its motion, see Fig.~\ref{fig:exp_presentation}.
As all the robots are subject to the connectivity-maintenance action of \eqref{eq:conn_ctrl}, the team follows the movement of the leader so as to keep the team connected.
Since the external human input is bounded, the connectivity maintenance action will always be able to overcome it and prevent disconnection from the group.
A discussion on the effectiveness of this approach in controlling multi-robot teams is presented in~\cite{aggravi2020connectivitymaintenance}.

The task begins with the team in a sparse random configuration. At the very beginning, the connectivity-maintenance algorithm makes the robots reaching a (local) maximum for the group connectivity by creating new links whenever two robots start sensing each other.
Then, as the human user imparts external commands to the robot leader $i=1$ to reach the target, the algorithm keeps adjusting the robot formation so as to satisfy the constraints described in Sect.~\ref{sec:conn_maint}, including the newly-introduced FOV weight.

\begin{figure}%
    \centering
    \subfloat[Side view of the drones trajectories.\label{fig:drones_traj}]{\includegraphics[width=0.48\linewidth]{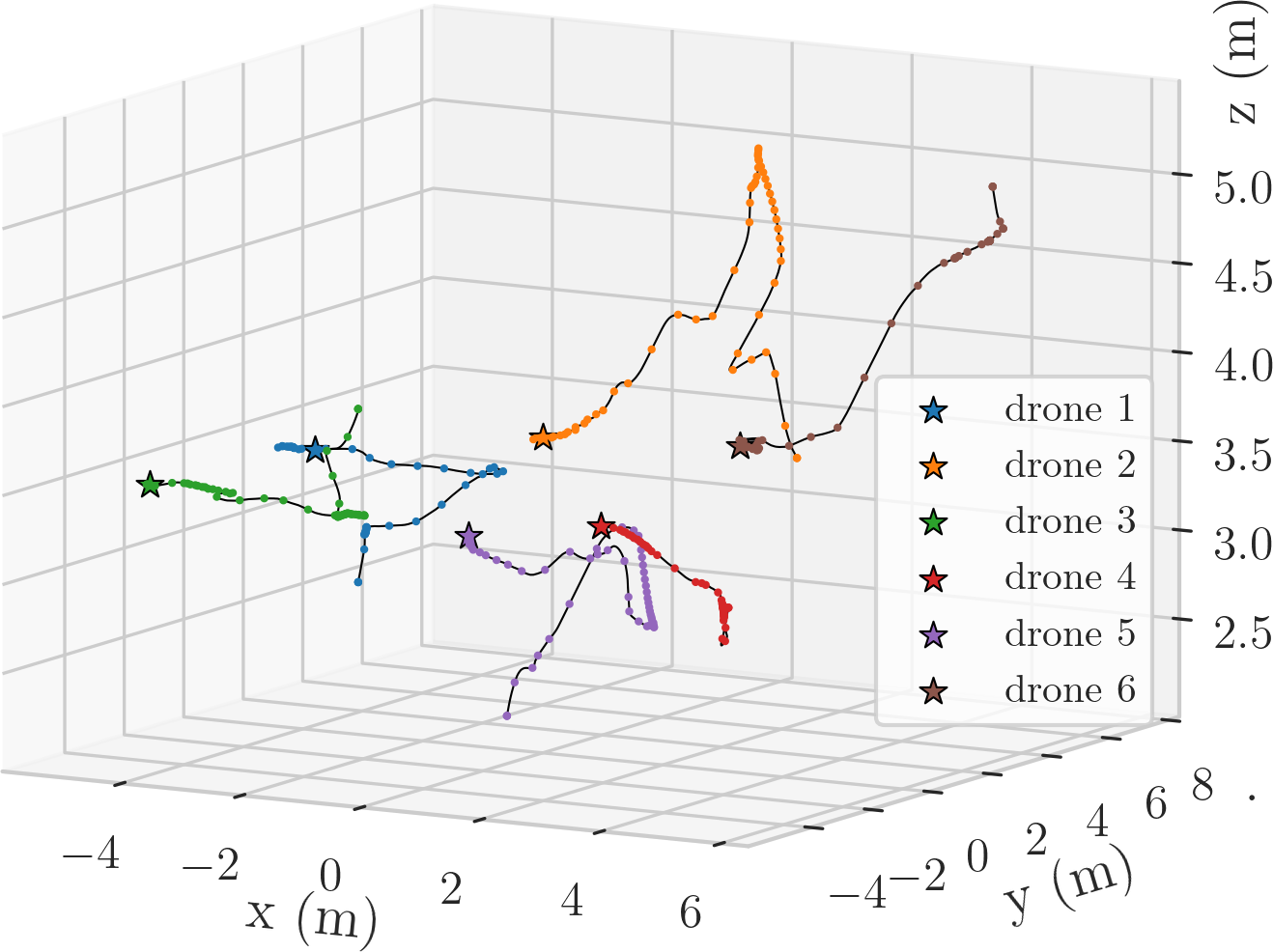}}
    \qquad
    \subfloat[Top view of the drones trajectories.\label{fig:drones_traj_top}]{\includegraphics[width=0.42\linewidth]{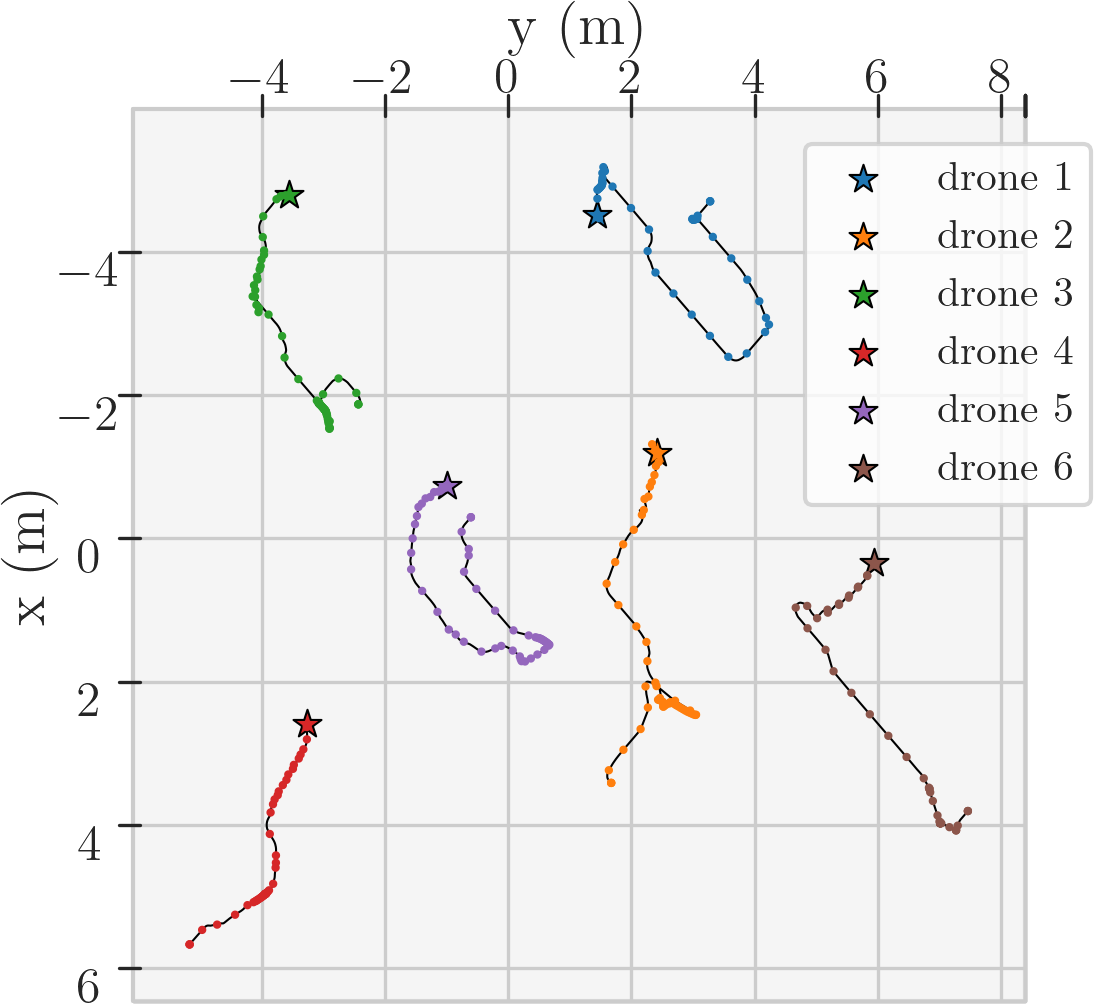}}
    \caption{Drones trajectories during the experiment. Drone 1 is the drone controlled by the human user. Each dot in the trajectory corresponds to the position of the drone at a 5 seconds interval.}%
\end{figure}

The video available at \url{https://youtu.be/CdnAmIrdBQU} shows various details of our approach.
Trajectories of the experiment present at 43 seconds in the online video (\url{https://youtu.be/CdnAmIrdBQU?t=43}) are shown in Fig.~\ref{fig:drones_traj} and Fig.~\ref{fig:drones_traj_top}.

\begin{figure}%
    \centering
    \includegraphics*[width=\linewidth]{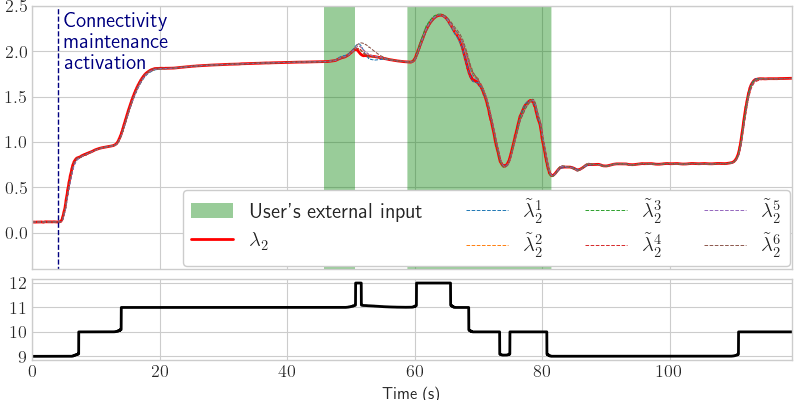}
    \caption{Validation in simulation. (Top) True $\lambda_2$ vs. the six estimations $\tilde{\lambda}^i_2$ calculated by the robots over time; the green phases indicate when the human user is providing an external input to the robot leader (robot $1$). (Bottom) Number of edges in the graph, which changes dynamically as the task evolves.}
    \label{fig:lambda2_est}
\end{figure}

Figure~\ref{fig:lambda2_est} (top) reports the behavior of the connectivity eigenvalue $\lambda_2(t)$ (solid red line) and the $N$ estimations $\tilde\lambda^i_2$ maintained by the $N$ robots during motion. One can note how the estimation error remains quite small despite the (sometimes) rapid changes in $\lambda_2(t)$ also due to the action of the human operator (green phases). Fig.~\ref{fig:lambda2_est} (bottom) reports the number of edges during motion which is not constant, as expected.

\begin{figure}%
    \centering
    \includegraphics*[width=\linewidth]{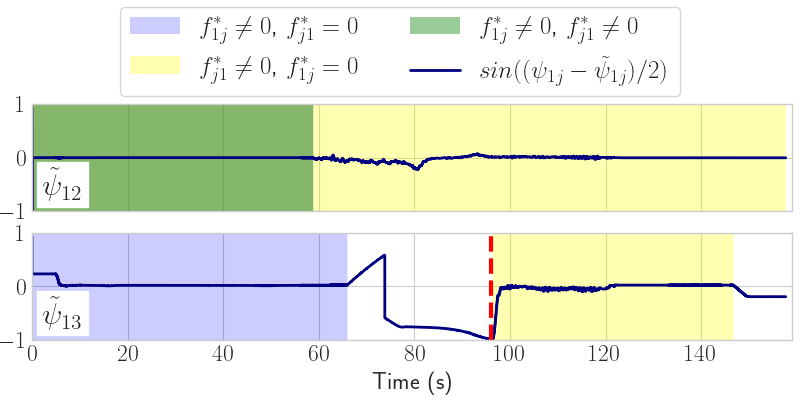}
    \caption{Estimations errors for $\tilde{\psi}_{12}$ and $\tilde{\psi}_{13}$. %
    Blue phases indicate when robot~$1$ is measuring robot~$j$ ($f^*_{1j} \neq 0$, see Sect.~\ref{sec:weight_design}) but robot~$j$ does not measure robot~$1$ ($f^*_{j1} = 0$); yellow phases indicate when robot~$j$ is measuring robot~$1$ but robot~$1$ does not measure robot~$j$; green phases indicate when both robot~$1$ and $j$ are measuring each other. Note how the estimation errors always remain very small apart from the white phase (bottom plot) since, in this case, the two robots are completely disconnected. The error however quickly converges back to zero as soon as one robot starts measuring the other one (vertical red dashed line).}
    \label{fig:psi_ij_est}
\end{figure}

Fig.~\ref{fig:psi_ij_est} reports the error in estimating the relative orientation among robot~$1$ and robot~$2$~and~$3$. For the reader's convenience, this estimation error is quantified by the quantity $\sin((\psi_{1j} - \tilde \psi_{1j})/2)$, which vanishes for $\psi_{1j} - \tilde \psi_{1j}=\pm2k\pi$. The plot also shows, with colored bands, the four possible visibility combinations among robot $1$ and the other robots (mutual visibility, one robot measuring the other but not being measured by it, the two robots not in visibility of each other). We can note that the estimation error is always limited. Moreover, when the two robots are in mutual visibility (green phase in Fig.~\ref{fig:psi_ij_est}) the error is close to zero as, in this case, a direct measurement of $\psi_{1j}$ is available. %
When the two robots are instead completely disconnected (white phases), no measurement is available. In this case, the estimations of $\psi_{1j}$ are not updated anymore, thus leading to an (unavoidable) increased estimation error, which is however quickly recovered when one of the robots (or both) become visible again (red dashed line). Estimation errors during blue and yellow phases can also temporarily increase when $\omega_i \neq 0$ or $\omega_j \neq 0$ but $v_j = 0$ (blue phase) or $v_i = 0$ (yellow phases) (see Sect.~\ref{sec:psi_estimation}).

\section{Conclusions}\label{sec:fov_conclusions}

We presented a decentralized connectivity-maintenance control algorithm for a team of quadrotor UAVs having limited FOV and not sharing a common reference frame for collectively expressing measurements and commands.
To the best of our knowledge, these contributions advanced the existing literature on this subject which, instead, has been based on the (simplifying) assumptions of omnidirectional sensing and presence of a common frame for all the robots in the group.
The proposed maintenance control action is therefore able to address a more realistic setting, closer to real-world conditions.
The approach has been successfully validated in realistic simulations involving $6$ quadrotor UAVs.

A limitation of the proposed algorithm lies in the model~(\ref{eq:model}), which assumes a simplified motion model that does not account for the tilting (roll/pitch) movements of quadrotors required to accelerate in space.
In scenarios where these tilting motions are significant, particularly outside of near-hovering regimes, the accuracy of the FOV weight may degrade, as the sensing capabilities of an onboard camera are influenced by the quadrotor's tilt angle. Addressing this limitation would require extending model~(\ref{eq:model}) to incorporate the effects of tilting on the field of view.

Another key challenge in vision-based relative localization is the identification of observed neighbors, which is not inherently resolved by the algorithm.
Several works from the literature, as reviewed in Sect.~\ref{sec:relativeSensing}, address this issue through different techniques.

\clearemptydoublepage
\graphicspath{{Chapitre5/figures/}}
\clearemptydoublepage
\chapter{Haptic shared control techniques for aerial multi-robot systems}\label{chap:chapitre5}
{\hypersetup{linkcolor=black}
\minitoc
}

\lettrine{T}{he} rapid advancement of robotic technologies has opened up new possibilities for efficient and effective task execution across various domains. Aerial robots, in particular, have emerged as versatile tools capable of addressing complex challenges in a broad range of environments. From urban planning and environmental monitoring to infrastructure inspection and entertainment, the applications of aerial robotics continue to expand, revolutionizing industries and enhancing human capabilities~\parencite{floreano2015science,marconi2012sherpa}.
The increasing sophistication of aerial robotic systems, however, brings forth the challenge of intuitive and efficient control mechanisms. As the number of robots in a team grows, traditional one-to-one control paradigms become impractical, necessitating innovative approaches to human-robot interaction~\parencite{tezza2019state,aggravi2020connectivitymaintenance}.

To meet the need for effective feedback in drone teleoperation, researchers have explored a variety of modalities. Haptic feedback has been a primary focus, with studies employing cable-driven systems~\parencite{rognon2018haptic}, vibrotactile armbands~\parencite{cacace2016multimodal}, and wearable tactile gloves~\parencite{tsykunov2019swarmtouch} to convey information about drone trajectory and formation states.
Grounded haptic interfaces have also been utilized, using force-feedback devices to control fleet leaders and provide team status updates~\parencite{lee2013semiautonomous,robuffo2011iros,franchi2012shared}. Some researchers have adopted hybrid approaches, integrating haptic feedback with visual or auditory cues to enhance situational awareness~\parencite{aggravi2020connectivitymaintenance}.

This chapter introduces a series of haptic-enabled shared control techniques for the teleoperation of multiple aerial robots.
We extend the foundational decentralized connectivity-maintenance algorithm for aerial robot teleoperation presented in the previous chapter by implementing versatile control architectures: \textit{(i)} a single leader-follower model, \textit{(ii)} a multiple leader-follower control approach, and \textit{(iii)} a formation centroid control.
With these extensions, we aim to enable diverse applications in complex environments, offering greater robustness and flexibility in maintaining formation and connectivity within the robotic team.
Our goal is to not only enhance the operator's situational awareness but also enable more intuitive and efficient management of multi-robot systems.

The key contributions of this work are:
\begin{enumerate}
    \item Design and development of new shared control techniques for human-swarm interaction, utilizing innovative haptic and visual interfaces for both control and feedback.
    \item Experimental evaluation via three human-subject studies conducted in a simulated environment, comparing the proposed feedback and control techniques in tasks of search and rescue, exploration, and surveillance.
\end{enumerate}

We address key limitations in current teleoperation approaches for multi-robot systems, notably the challenges of keeping desired team formations, enabling intuitive and responsive human control, and grounding the framework in practical sensing and communication assumptions suitable for real-world deployment.

To enhance human-robot interaction, we introduce advanced haptic feedback mechanisms, supported by user studies, which improve control precision and operator immersion. Additionally, we develop an improved visual interface to optimize operator situational awareness, facilitating the efficient management of multiple robots in complex environments while minimizing ground-to-remote communication requirements.

This work is under review for publication in the \textit{IEEE Transactions on Human-Machine Systems} journal.

\section{Automatic Leader Selection}
\label{subsec:automatic-leader-selection}
This section introduces a method for automatically selecting the leading robot(s) in a group to minimize the constraints imposed by the connectivity maintenance task.
Yaw control in multi-agent systems is particularly challenging when agents must balance field-of-view (FOV) constraints with the need to maintain connectivity.
These requirements can restrict the operator's ability to freely adjust an agent's orientation without compromising visibility constraints.
We propose a method to help the operator select the agent with the greatest rotational freedom.
The approach is based on assessing each agent's ability to \textit{look around}, which reflects how unconstrained the agent is in maintaining its neighbors within its FOV.

Recalling the definition of $f^*_{ij}$, from Section.~\ref{sec:weight_design}, as the function that quantifies the visibility of robot $j$ from robot $i$:
\begin{equation}\label{eq:f_ij^*_2}
    f^*_{ij}(a_{ij})=w^0_1(a_{ij},\,a_{min},\,a_{max})
\end{equation}
where $a_{ij}=\mathbf{o}^T_C\bbeta_{ij}$ represents the cosine of the angle between $\bbeta_{ij}$, the unit vector pointing from robot $i$ to robot $j$ (bearing vector), and $\mathbf{o}_C$, the fixed body-frame direction of the camera's optical axis.

We can quantify the \emph{yaw maneuverability} of robot $i$ based on the number of neighbors that can see it.
Specifically, if a robot is seen by many neighbors, we can infer that it faces fewer constraints on its yaw orientation, as it does not need to keep those neighbors within its FOV to maintain connectivity.
To quantify this, the quantity $\sum_{j\in\calN_i}f^*_{ji}$ is a good metric as it represents the visibility of robot $i$ from its neighbors, with $f^*_{ji}$ defined as in~(\ref{eq:f_ij^*_2}):
\begin{equation}
    f^*_{ji}(a_{ji})=w^0_1(a_{ji},\,a_{min},\,a_{max})
\end{equation}

Depending on the task, this metric can be used to select as a leader the agent $L$ with the greatest \emph{yaw maneuverability} in a set $\emph{S}$ as
\begin{equation}\label{eq:leader_selection}
    L = \argmax_{i\in\emph{S}}\sum_{j\in\calN_i}f^*_{ji},
\end{equation}
where $\emph{S}$ represents a suitable subset of the robots in the group, depending on the need of the task (see also Secs.~\ref{sec:haptic-guidance} and~\ref{sec:implementation}).

The next section will introduce the haptic shared control techniques and illustrate how the automatic leader selection is used to assist the operator.

\section{Haptic shared control for Connectivity Maintenance}\label{sec:haptic-guidance}
Haptic shared control enables a human user to control a complex robotic system, such as a group of aerial robots, while receiving force feedback on the status of the team and task.
This Section outlines how we enable users to control the robotic team via a haptic interface and how they are provided with suitable sensory feedback.

\subsection{Haptic interface}
\label{subsection:haptic-interface}

\begin{figure}[htp]
    \centering
    \subfloat[Virtuose device\label{fig:virtuose}]{\includegraphics[width=0.5\linewidth]{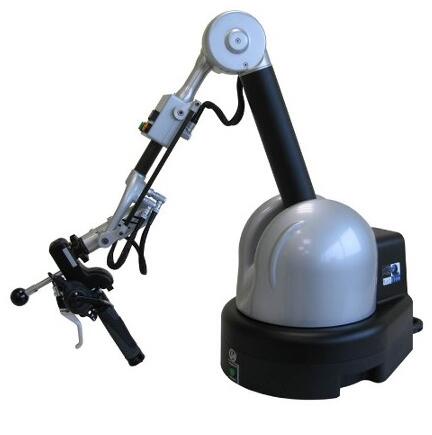}}
    \subfloat[Close-up of the handle\label{fig:virtuose_buttons}]{\includegraphics[width=0.5\linewidth]{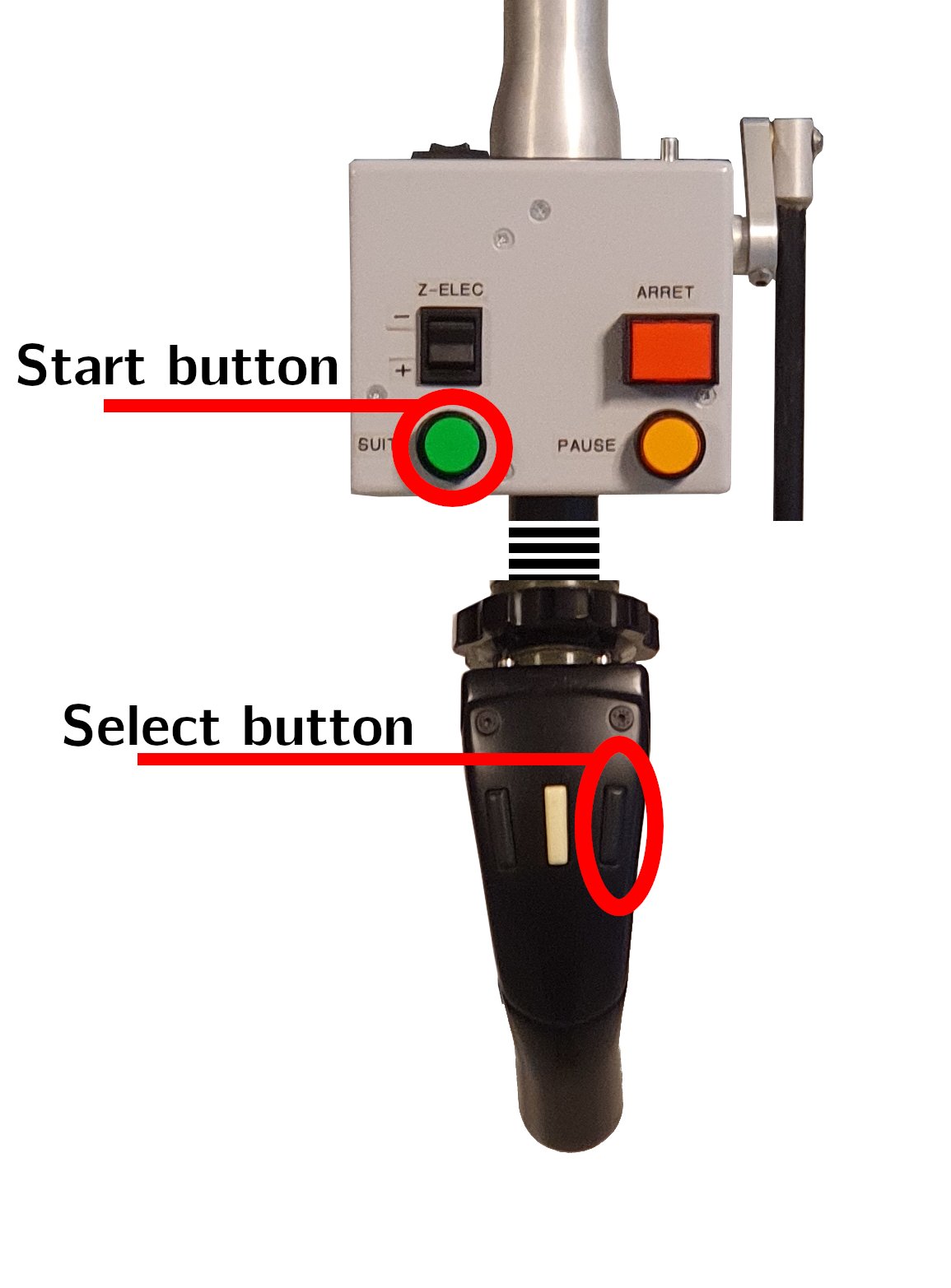}}
    \vspace{9pt}
    \caption{The Virtuose device is a high-end 6-DoF haptic interface able to provide 3-DoF translational and 3-DoF rotational feedback.}%
\end{figure}

We use a 6 active degrees-of-freedom haptic device (Virtuose, Haption), shown in Fig.~\ref{fig:virtuose}. It is able to provide force and torque feedback along all translational and rotational axes, allowing users to experience realistic interactions with virtual or remote objects across a wide workspace.
The device handle is equipped with a laser sensor that detects when the user grasps it.
This sensor serves two purposes:
\begin{enumerate}
    \item It acts as a security trigger, i.e., a dead man's switch, which disables both the control inputs and the applied force when the user is not holding the handle.
    \item It sets the origin of the user's workspace $\idlePos$ to the current position of the handle when the user grasp the handle for the first time, so as to enable clutching.
\end{enumerate}

Following~\cite{aggravi2021hapticenabled,aggravi2021decentralized}, we design three force contributions to achieve such a joystick-like behavior:
\begin{itemize}
    \item $\deadzoneForce$ ({deadzone force}): This force helps users stay at the center of the workspace, within a sphere with radius $\deadzone$ around $\idlePos$, whenever they do not want to provide any control input; within this deadzone, control inputs are disabled.
    \item $\centeringForce$ ({centering force}): This force brings users towards the center of the workspace $\idlePos$, making the deadzone easy and intuitive to find.
    \item $\limitForce$ ({workspace limit force}): This force keeps the user within the device workspace, in a sphere with radius $\workspaceRange$ centered on $\idlePos$.
\end{itemize}

For translations, let $\virtuosePos$ represent the position of the handle, with $\virtuoseInput = \virtuosePos - \idlePos$. We define
\begin{eqnarray}
    \label{eq:deadzoneForce}
    \deadzoneForce = &
    \left\{
    \begin{array}{c}
    -\deadzoneGain \virtuoseInput\\
    0
    \end{array}
    \right.
    &
    \begin{array}{c}
        \lVert \virtuoseInput \rVert \leq \deadzone  \\
        \lVert \virtuoseInput \rVert > \deadzone
    \end{array}
    \\
    \nonumber\\
    \label{eq:centeringForce}
    \centeringForce = &
    \begin{array}{c}
        - \centeringGain \virtuoseInput
    \end{array}
    &
    \begin{array}{c}
        \lVert \virtuoseInput \rVert \in \calbR
    \end{array}
    \\
    \nonumber\\
    \label{eq:limitForce}
    \limitForce = &
    \left\{
    \begin{array}{c}
    0\\
    -\limitGain \left( \workspaceRange\dfrac{\virtuoseInput}{\lVert \virtuoseInput \rVert} \right)
    \end{array}
    \right.
    &
    \begin{array}{c}
    \lVert \virtuoseInput \rVert \leq \workspaceRange  \\
    \lVert \virtuoseInput \rVert > \workspaceRange
    \end{array}
\end{eqnarray}
with $\deadzoneGain$, $\centeringGain$, and $\limitGain$ being the deadzone, the centering, and the workspace limit gains, respectively.

\begin{figure}[htp]
    \centering
    \includegraphics*[width=\linewidth]{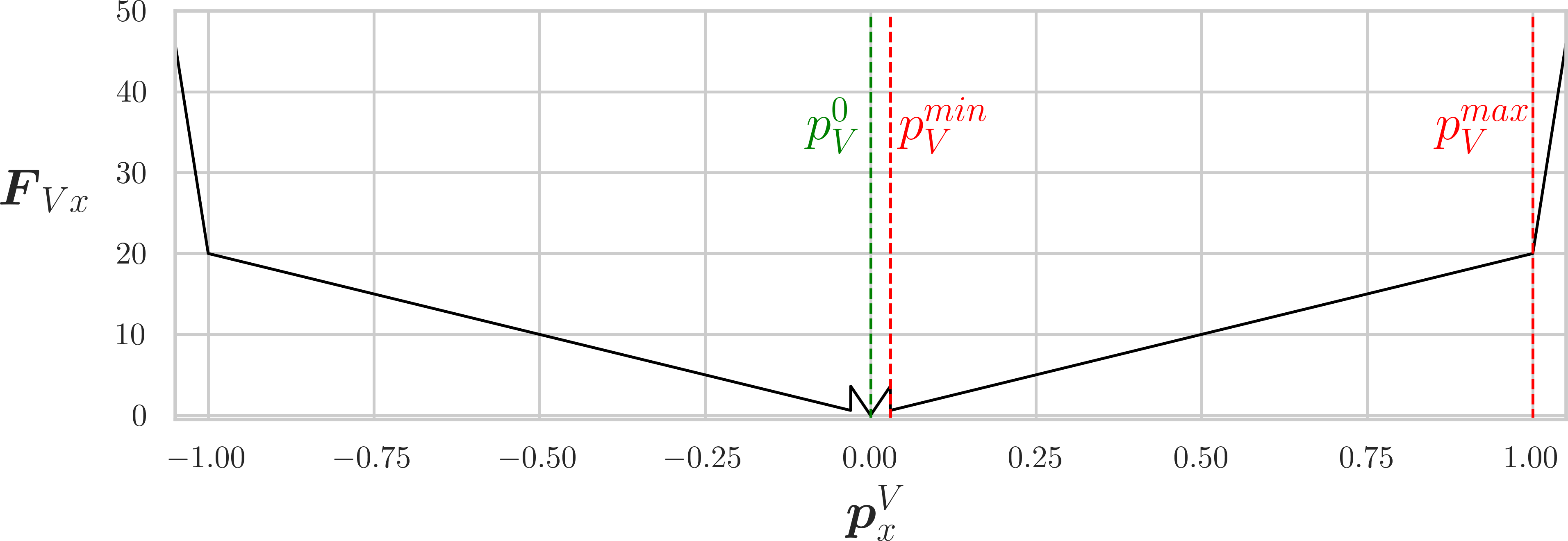}
    \caption{The shape of $\joystickForce$ on the $x$ axis with $\idlePos = 0$, $\deadzone = 0.03$, $\workspaceRange = 1$, $\deadzoneGain = 100$, $\centeringGain = 20$,
and $\limitGain = 500$. }
    \label{fig:virtuose_spring_force}
\end{figure}

From (\ref{eq:deadzoneForce}--\ref{eq:limitForce}), we define $\joystickForce$ as this base force applied to the handle:
\begin{equation}\label{eq:virtuose_spring_force}
    \joystickForce = \deadzoneForce + \centeringForce + \limitForce.
\end{equation}
Figure~\ref{fig:virtuose_spring_force} illustrates the shape of $\joystickForce$ on the $x$-axis with $\idlePos = 0$, $\deadzone = 0.03$, $\workspaceRange = 1$, $\deadzoneGain = 100$, $\centeringGain = 20$,
and $\limitGain = 500$.

For rotations, let $\idleRot$ represent the idle orientation of the handle and $\virtuoseRot$ its current orientation. The difference in orientation, $\virtuoseInputR$, is defined as the angular difference between $\virtuoseRot$ and $\idleRot$ in Euler angles.
Similarly as above, we can define $\deadzoneForceR$ as the deadzone torque, $\centeringForceR$ as the centering torque, and $\limitForceR$ as the workspace limit torque, expressed as follows:
\begin{eqnarray}
    \label{eq:deadzoneForceR}
    \deadzoneForceR = &
    \left\{
    \begin{array}{c}
    -\deadzoneGainR \virtuoseInputR\\
    0
    \end{array}
    \right.
    &
    \begin{array}{c}
        \lVert \virtuoseInputR \rVert \leq \deadzoneR  \\
        \lVert \virtuoseInputR \rVert > \deadzoneR
    \end{array}
    \\
    \nonumber\\
    \label{eq:centeringForceR}
    \centeringForceR = &
    \begin{array}{c}
        - \centeringGainR \virtuoseInputR
    \end{array}
    &
    \begin{array}{c}
        \lVert \virtuoseInputR \rVert \in \calbR
    \end{array}
    \\
    \nonumber\\
    \label{eq:limitForceR}
    \limitForceR = &
    \left\{
    \begin{array}{c}
    0\\
    -\limitGainR \left( \workspaceRangeR\dfrac{\virtuoseInputR}{\lVert \virtuoseInputR \rVert} \right)
    \end{array}
    \right.
    &
    \begin{array}{c}
    \lVert \virtuoseInputR \rVert \leq \workspaceRangeR  \\
    \lVert \virtuoseInputR \rVert > \workspaceRangeR
    \end{array}
\end{eqnarray}

Variables $\deadzoneGainR$, $\centeringGainR$, and $\limitGainR$ represent the deadzone gain, the centering gain, and the workspace limit gain for the rotational motion, respectively.
We define $\joystickForceR$ as this base torque applied to the handle:
\begin{equation}\label{eq:virtuose_spring_torque}
    \joystickForceR = \deadzoneForceR + \centeringForceR + \limitForceR.
\end{equation}

Forces $\joystickForce$ and $\joystickForceR$ are always active and are of course designed to be weaker than the other force contributions related to the status of the team and task, as described in Sec.~\ref{sec:implementation}.
This mapping approach is quite common and enables precise yet intuitive use of the haptic interface~\parencite{rahal2020caring,rahal2023haptic}.

\subsection{Control and feedback conditions}
\label{sec:conditions}
For a set of $N$ robots, let $\controlledAgents(t)$ denote the time-varying set of agents controlled by the user, where $0 \leq |\controlledAgents(t)| \leq N$.
{To facilitate user-agent interaction, we assume that a ground station has access to the position of each controlled agent $j \in \controlledAgents(t)$ within a common reference frame, as well as their connectivity maintenance inputs. The ground station hosts the user interfaces and communicates directly with the agents in the group.}
While the robots are implementing the connectivity maintenance control presented in Sect.~\ref{sec:conn_maint}, we consider three control methods that allow a single user to steer the whole group:

\begin{itemize}
    \item \textbf{Single Agent Control (S)}: The user controls one agent at a time, with $|\controlledAgents| = 1$.
    \item \textbf{Pinning Control (P)}: The user controls a virtual point that exerts a spring-like force to pull selected agents, with $|\controlledAgents| \in \{1,...,N\}$.
    \item \textbf{Centroid Control (C)}: The user controls the entire robot group simultaneously by translating and rotating it around the group's centroid,
    as well as scaling/expanding the formation, with $|\controlledAgents| = N$.
\end{itemize}

For each of these control methods, we consider two haptic feedback conditions:
\begin{itemize}
    \item \textbf{No haptic feedback (NH)}: the user receives no haptic feedback about the connectivity force applied on the robot(s) they are controlling. Only the joystick-like forces $\joystickForce$ and torques $\joystickForceR$ are applied to the Virtuose device, as introduced in Sect.~\ref{subsection:haptic-interface}.
    \item \textbf{Haptic feedback (H)}:  the user receives haptic feedback about the connectivity force applied on the robot(s) they are controlling. The forces and torques applied to the Virtuose device are the sums of the joystick-like and feedback contributions, i.e., $\joystickForce + \forceFeedback$ and $\joystickForceR + \torqueFeedback$.
\end{itemize}

According to the design of the connectivity maintenance control described in~Sec.~\ref{sec:conn_maint},
we define $\connectivityVelocityj$ and $\connectivityYawRatej$ as the linear and angular velocities of robot $j$ obtained from the connectivity maintenance algorithm.
Stability of the coupling between connectivity maintenance and external inputs can be shown by following the same arguments as in~\cite{lee2013semiautonomous}, i.e., for any bounded external action (in our case, a bounded external command from the human operator), the closed-loop system is stable besides improbable (zero-measure) situations of perfect (and unlikely) alignments among the robots.

In the case \textbf{NH} (no haptic feedback), the user-controlled agents $j \in \controlledAgents$ follow the velocity control input $(\connectivityVelocityj, \connectivityYawRate)$ from the connectivity maintenance algorithm, in addition to the user's inputs, i.e.,
\begin{equation} \label{eq:user-inputs}
    \left[ \begin{array}{c}
        {\vv}_j \\
        {\omega}_j
    \end{array} \right]
    =
    \left[ \begin{array}{ccc}
        \connectivityVelocityj & + &\externalInputLinear\\
        \connectivityYawRatej & + &\externalInputAngular
    \end{array} \right].
\end{equation}

In the case \textbf{H} (haptic feedback), the controlled agents $j \in \controlledAgents$ follow the user's inputs, supplemented by a residual connectivity maintenance input $(\residualconnectivityVelocityj, \residualconnectivityYawRatej)$.
This input ensures that connectivity is maintained if the haptic feedback is unable to provide enough force to guide the user (e.g., saturation of the haptic device force).
Details of this mechanism will depend on the control method at hand, as presented in subsections \ref{subsec:single-agent-control}, \ref{subsec:pinning-control}, and \ref{subsec:centroid-control}.
The velocity control inputs are then expressed as:
\begin{equation} \label{eq:user-inputs-h}
    \left[ \begin{array}{c}
        {\vv}_j \\
        {\omega}_j
    \end{array} \right]
    =
    \left[ \begin{array}{ccc}
        \residualconnectivityVelocityj & + & \externalInputLinear\\
        \residualconnectivityYawRatej & + & \externalInputAngular
    \end{array} \right].
\end{equation}
In this case, the connectivity maintenance control input
$(\connectivityVelocity, \connectivityYawRate)$ is not directly used as a control input, but it is provided to the users as haptic guidance, informing them about the connectivity force acting on the robot(s) they are controlling.
With $\forceFeedback$ representing the force feedback and $\torqueFeedback$ representing the torque feedback applied by the \textit{Virtuose} haptic device, we have:
\begin{equation} \label{eq:force-feedback-general}
    \left[ \begin{array}{c}
        {\forceFeedback} \\
        {\torqueFeedback}
    \end{array} \right]
    =
        \begin{array}{c}
            \left[f(\connectivityVelocity, \connectivityYawRate)\right].
    \end{array}
\end{equation}
As $(\connectivityVelocity, \connectivityYawRate)$ increase when $\lambda_2$ is low, the force and torque feedback $(\forceFeedback, \torqueFeedback)$ consequently become more intense.
To prevent overwhelming the user with excessive feedback, we implement a saturation mechanism that limits the feedback to maximum values of $\maxForceFeedback$ and $\maxTorqueFeedback$, to be adjusted according to the needs of the task and haptic system.

In the following subsections, we describe the three control methods and the associated haptic force feedback. A video, available at \url{https://youtu.be/gDR6Lky8_OI}, shows how the different control methods work.

\subsection{Single Agent Control (S)}
\label{subsec:single-agent-control}

This control mode enables the user to control one (leader) agent of the team. The rest of the agents follow the connectivity maintenance control presented in~Sec.~\ref{sec:conn_maint}. This control modality is similar to that used in the previous chapter and in~\cite{aggravi2020connectivitymaintenance,aggravi2021hapticenabled}.
\subsubsection{Control Inputs}
\label{subsubsec:single-agent-control-inputs}
The user controls a single agent $j$ with four degrees of freedom: $(\vv_j, \omega_j)$ where $\vv_j$ represents the linear velocities in the robot body frame and $\omega_j$ its yaw rate.
As showed in~(\ref{eq:user-inputs}), the user's inputs are applied to the selected agent as an external contribution $(\externalInputLinear, \externalInputAngular)$.

With $\virtuoseInput$ and $\virtuoseInputR = (\virtuoseInputRoll, \virtuoseInputPitch, \virtuoseInputYaw)$ representing the control inputs from the \textit{Virtuose} haptic device as described in Sec.~\ref{subsection:haptic-interface}, the control inputs are given by:
\begin{equation} \label{eq:single-agent-control-inputs}
    \left[ \begin{array}{c}
        \externalInputLinear \\
        \externalInputAngular
    \end{array} \right]
    =
    \left[ \begin{array}{c}
        k_{Ev}\virtuoseInput \\
        k_{E\omega}\virtuoseInputYaw
    \end{array} \right],
\end{equation}
with $k_{Ev}$ and $k_{E\omega}$ being the linear and angular velocity gains, respectively.

\subsubsection{Haptic Feedback (H)}
\label{subsubsec:single-agent-haptic-feedback}
With $\forceFeedback$ representing the force feedback and $\torqueFeedback$ the torque feedback guiding the user, we have
\begin{equation} \label{eq:SH-force-feedback}
    \left[ \begin{array}{c}
        {\forceFeedback} \\
        {\torqueFeedback}
    \end{array} \right]
    =
    \left[ \begin{array}{c}
        k_F\connectivityVelocityj\\
        k_T(0, 0, \connectivityYawRatej)
    \end{array} \right],
\end{equation}
with $k_F$ and $k_T$ being the force and torque feedback gains, respectively.

By pressing the ``select'' button, as shown in Figure~\ref{fig:virtuose_buttons}, the user can switch the control to a different agent (only one agent at a time).
The new agent to be controlled is selected using the algorithm presented in Sec.~\ref{subsec:automatic-leader-selection}.
When the controlled agent is changed, the previously controlled robot resumes following the
connectivity maintenance control $(\connectivityVelocity, \connectivityYawRate)$, while the newly selected agent receives only the user's inputs as discussed above.

\subsubsection{No Haptic Feedback (NH)}
\label{subsubsec:single-agent-no-haptic-feedback}
In this case, no additional haptic feedback is provided to the operator; only the joystick-like force $\joystickForce$ (eq.~\ref{eq:virtuose_spring_force}) and torque $\joystickForceR$ (eq.~\ref{eq:virtuose_spring_torque}) are provided. The user-controlled agent $j$ follows the velocity control input $(\connectivityVelocityj, \connectivityYawRate)$ from the connectivity maintenance algorithm, in addition to the user's inputs, as presented in eq.~(\ref{eq:user-inputs}).

\subsection{Pinning Control (P)}
\label{subsec:pinning-control}

\subsubsection{Control Inputs}
\label{subsubsec:pinning-control-inputs}
The user controls a virtual point that pulls selected agents using a spring-like force: $|\controlledAgents| \in \{1...N\}$.
Let $(\pinningPointPos, \pinningPointYaw)$ represent the position and yaw of the virtual point and $(\pinningPointVelocity, \pinningPointYawrate)$ its linear and angular velocities in the body frame.
The dynamics of the virtual pinning point are expressed as:
\begin{equation} \label{eq:pinning-point-dynamics}
    \left[ \begin{array}{c}
        \dotpinningPointPos \\
        \dotpinningPointYaw
    \end{array} \right]
    =
    \left[ \begin{array}{c}
        {\pinningPointVelocity} \\
        {\pinningPointYawrate}
    \end{array} \right]
    =
    \left[ \begin{array}{c}
        k_v\virtuoseInput \\
        k_{\omega}\virtuoseInputYaw
    \end{array} \right].
\end{equation}

\begin{figure}[htp]
    \centering
    \subfloat[Translational Pinning Force\label{fig:pinning-force-passive}]{\includegraphics[width=0.5\linewidth]{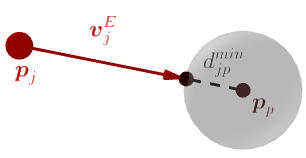}}
    \subfloat[Rotational Pinning Force\label{fig:pinning-force-rotational}]{\includegraphics[width=0.5\linewidth]{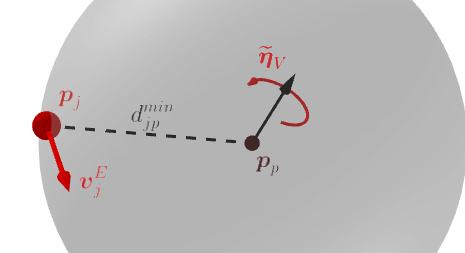}}
    \vspace{8pt}
    \caption{Pinning control. $\pinningPointPos$ and $\agentPosj$ are the position of the virtual point and the agent $j$, respectively. $\pinningMinDistj$ is the minimum distance between agent $j$ and the virtual point, $\virtuoseInputR$ is the user rotational input and $\externalInputLinear$ is the velocity control input controlling agent $j$.
    }
\end{figure}

The user can impart two types of motions to the controlled UAVs: translational motion by moving the virtual point to draw the agents towards it and rotational motion by orbiting around the virtual point.
For each controlled robot $j \in \controlledAgents$, the spring-like force applied is given by:
\begin{equation}\label{eq:pinning-control-inputs}
    \left\{
        \begin{array}{rl}
            \externalInputLinear =& k_{Ev}\RtoAgent{j} \left( \pinningVect \left(1-\dfrac{\pinningMinDistj}{\norm{\pinningVect}}\right) - \pinningVect \times \virtuoseInputR\right) \\
            \externalInputAngular =& 0
        \end{array}
    \right.
\end{equation}
where $\pinningVect = (\pinningPointPos-\agentPosj)$ and $\pinningMinDistj$ is the fixed minimum distance between the virtual point and the agent $j$.
Figure~\ref{fig:pinning-force-passive} illustrates the pinning force that pulls agent $j$ towards the orbit of the virtual point, while Figure~\ref{fig:pinning-force-rotational} shows the rotational pinning force, enabling the user to control the agent's orbit around the virtual point.

To provide visual feedback to the user about the pinning location, at least one agent should face the direction of motion of the group.
To achieve this, we can select an agent $l$ from the set of controlled UAVs $\controlledAgents$ using the algorithm outlined in Sec.~\ref{subsec:automatic-leader-selection}.
This agent is then controlled to minimize the angle between its onboard camera direction and the vector $\pinningVect$.
The yaw control input for the selected robot $l$ is given by:
\begin{equation}\label{eq:pinning-control-inputs-yaw}
    \externalInputAngular = k_{E\omega}\sin\left[\arctan(\pinningVect^y/\pinningVect^x) - \agentYaw{l}\right],
\end{equation}
where $\agentYaw{l}$ is the yaw of robot $l$.

In both feedback cases \textbf{NH} and \textbf{H}, the selected agent $l$ follows the pinning control yaw rate input in addition to its implemented connectivity maintenance yaw rate.
This is expressed as
\begin{equation}\label{eq:pinning-control-inputs-yawrate}
    \omega_l = \connectivityYawRatej + \externalInputAngular.
\end{equation}
To ensure that connectivity is maintained, the residual connectivity inputs $(\residualconnectivityVelocityj, \residualconnectivityYawRatej)$ are applied to the controlled agents $j \in \controlledAgents$ as follows:
\begin{equation}\label{eq:residual-connectivity-inputs-pinning}
    \left\{
        \begin{array}{rl}
            \residualconnectivityVelocityj =& k_r\connectivityVelocityj\\
            \residualconnectivityYawRatej =& k_r\connectivityYawRatej
        \end{array},
    \right.
\end{equation}
with $k_r$ being a scaling gain (set to $0.1$ in our experiments).
As the connectivity maintenance force becomes increasingly strong for low values of $\lambda_2$, these residual inputs ensure that connectivity is maintained if the haptic feedback proves insufficient or unable to fully guide the user, while otherwise allowing the user with maximum control freedom.

\subsubsection{Haptic Feedback (H)}
\label{subsubsec:pinning-haptic-feedback}
\begin{figure}[htp]
    \centering
    \includegraphics[width=0.4\linewidth]{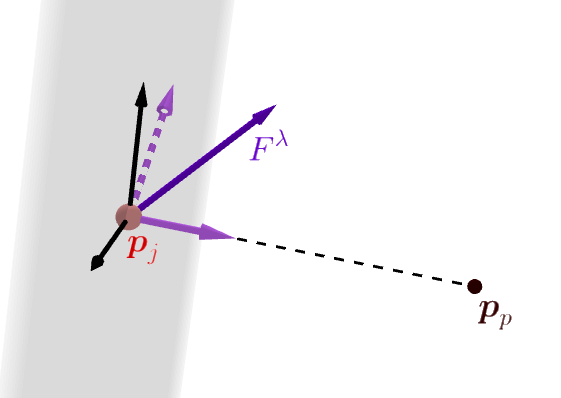}
    \caption{The projection of the linear elements of the connectivity maintenance force onto the vector $\pinningVect$ and the tangent plan containing $\pinningVect \times \virtuoseInputR$.}
    \label{fig:pinning_force_projection}
\end{figure}

In the set of controlled robots $\controlledAgents$, the user controls the virtual point rather than the individual agents directly.
Consequently, the force feedback guiding the user is the average of the projection of the connectivity maintenance forces
$\connectivityForcej, j \in \controlledAgents$ onto the vector $\pinningVect$ and the tangent plane containing $\pinningVect \times \virtuoseInputR$, as illustrated in Figure~\ref{fig:pinning_force_projection}.

Let $\boldsymbol{I}^3$ denote the $3\times3$ identity matrix, $J = |\controlledAgents|$ be the number of controlled agents, and $\pinningUnitVect = \pinningVect/\lVert \pinningVect \rVert$ represent the unit direction of the pinning force.
The force feedback is expressed as:
\begin{equation} \label{eq:force-feedback-pinning}
    \left[\begin{array}{c}
        {\forceFeedback} \\
        {\torqueFeedback}
    \end{array} \right]
    =
    \left[\begin{array}{c}
            \dfrac{k_F}{J}\sum_{j\in\controlledAgents} \left[ \pinningUnitVect\pinningUnitVect^T\right]\RfromAgentj\connectivityVelocityj \\
            \dfrac{k_T}{J}\sum_{j\in\controlledAgents} \left[\boldsymbol{I}^3 - \pinningUnitVect\pinningUnitVect^T\right]\RfromAgentj\connectivityVelocityj \\
    \end{array}\right]
\end{equation}
with $k_F$ and $k_T$ the force and torque feedback gains, respectively.

\subsubsection{No Haptic Feedback (NH)}
\label{subsubsec:pinning-no-haptic-feedback}
In this case, no additional haptic feedback is provided to the operator; only the joystick-like force $\joystickForce$ (eq.~\ref{eq:virtuose_spring_force}) and torque $\joystickForceR$ (eq.~\ref{eq:virtuose_spring_torque}) are provided.  The user-controlled agent $j$ follows the velocity control input $(\connectivityVelocityj, \connectivityYawRate)$ from the connectivity maintenance algorithm, in addition to the user's inputs, as presented in eq.~(\ref{eq:user-inputs}).

\subsection{Centroid Control (C)}
\label{subsec:centroid-control}

\subsubsection{Control Inputs}
\label{subsubsec:centroid-control-inputs}
In this configuration, the user controls the entire robot group with seven degrees of freedom: translations, orbiting around the centroid, and scaling of the group.
The centroid position of the group, $\centroidPos$, is not directly available to the individual agents and is calculated centrally by the ground station. This value is used to determine the velocity inputs sent to the drones. The centroid position is computed as the average of the positions of the controlled agents $j \in \controlledAgents$, given by:

\begin{equation}\label{eq:centroid-position}
    \centroidPos = \dfrac{1}{N}\sum_{j\in\controlledAgents} \agentPosj,
\end{equation}
with $N$ denoting the total number of robots in the group.
The orientation of the centroid $\centroidYaw$ is controlled by the user with the dynamics:
\begin{equation}\label{eq:centroid-yaw}
    \dot{\centroidYaw} = k_{\omega}\virtuoseInputYaw.
\end{equation}

Let $\centroidUserInputs = [\centroidUserInputsSub{x}, \centroidUserInputsSub{y}, \centroidUserInputsSub{z}, \centroidUserInputsSub{\phi}, \centroidUserInputsSub{\theta}, \centroidUserInputsSub{\psi}, \centroidUserInputsSub{\epsilon}]$ represent the user's inputs for the centroid control.
We define the control mapping matrix $\centroidMappingMatrix$ as follows:
\begin{equation}\label{eq:centroid-mapping-matrix}
    \centroidMappingMatrix =
    \begin{bmatrix}
        \begin{array}{c}
            \boldsymbol{I}^6\\
            \begin{array}{cccccc}
                0 & 0 & 0 & 0 & 1 & 0
            \end{array}
        \end{array}
    \end{bmatrix},
\end{equation}
with $\boldsymbol{K}_{\centroidUserInputs} = [K_{\centroidUserInputsSub{x}}, K_{\centroidUserInputsSub{y}}, K_{\centroidUserInputsSub{z}}, K_{\centroidUserInputsSub{\phi}}, K_{\centroidUserInputsSub{\theta}}, K_{\centroidUserInputsSub{\psi}}, K_{\centroidUserInputsSub{\epsilon}}]$ representing a vector of gains. The 6 DoF inputs from the \textit{Virtuose} device are then mapped to the centroid control inputs as follows:
\begin{equation}\label{eq:6dof-to-centroid}
    \begin{split}
        \centroidUserInputs & =
        \boldsymbol{K}_{\centroidUserInputs} \circ
        \left(
        \centroidMappingMatrix
        \begin{bmatrix}
            \virtuoseInput \\
            \virtuoseInputR
        \end{bmatrix}
        \right) \\
        & =
        \boldsymbol{K}_{\centroidUserInputs} \circ
        \begin{bmatrix}
            \virtuoseInputSub{x},
            \virtuoseInputSub{y},
            \virtuoseInputSub{z},
            \virtuoseInputRoll,
            \virtuoseInputPitch,
            \virtuoseInputYaw,
            \virtuoseInputPitch
        \end{bmatrix}^T
    \end{split},
\end{equation}
with $\circ$ representing the Hadamard product.

Let $\centroidVect = (\centroidVectSub{x}, \centroidVectSub{y}, \centroidVectSub{z}) = \RtoAgent{j}(\agentPosj - \centroidPos)$ denote the relative position of each robot $j$ with respect to the centroid of the group $\centroidPos$ in the body frame of the agent $j$.
We define $\centroidVectUnit = \centroidVect/\norm{\centroidVect}$ as the normalized vector of $\centroidVect$.
Using $b_p \in \{0,1\}$ and $\overline{b_p} = (1-b_p)$ to switch between \textit{pan-tilt} control ($b_p=1$) and \textit{scaling} control ($\overline{b_p}=1$),
we define the Centroid Control matrix $\centroidControlMatrixj$ as follows:
\begin{equation}\label{eq:centroid-control-matrix-j}
    \centroidControlMatrixj =
    \begin{bmatrix}
        \begin{array}{c@{}c@{}c}
            \begin{array}{c}
                \\
                \RtoAgent{j} \\
                \\
                \begin{array}{c}
                    \mathbf{0}
                \end{array}
            \end{array}
            & \begin{array}{cc}
                            0 & 0\\
                            0 & 0\\
                            b_p\centroidVectSub{x} & -b_p\centroidVectSub{y} \\
                            0 & 0
                        \end{array} & \begin{array}{cc}
                                        -\centroidVectSub{y} & \overline{b_p}\centroidVectUnitSub{x}\\
                                        \centroidVectSub{x} & \overline{b_p}\centroidVectUnitSub{y}\\
                                        0 & 0\\
                                        1 & 0
                                    \end{array}
        \end{array}\\
    \end{bmatrix}.
\end{equation}

From~(\ref{eq:centroid-control-matrix-j}), we construct the global matrix $\centroidControlMatrix$ as follows:
\begin{equation}\label{eq:centroid-control-matrix}
    \centroidControlMatrix =
    \begin{bmatrix}
        \vdots\\
        \centroidControlMatrixj\\
        \vdots
    \end{bmatrix}_{j = 1..N}
\end{equation}
The value of $b_p$ can be changed by pressing a button on the \textit{Virtuose} device, allowing the user to switch between \textit{pan-tilt} and \textit{scaling} control.

From~(\ref{eq:centroid-control-matrix}), the control inputs for an individual agent $j \in \controlledAgents$ can be expressed as:
\begin{equation} \label{eq:centroid-drones-inputs}
    \left[ \begin{array}{c}
        \externalInputLinear \\
        \externalInputAngular
    \end{array} \right]
    =
    \centroidControlMatrixj \centroidUserInputs,
\end{equation} and the overall control inputs for the entire robot group is then given by:
\begin{equation} \label{eq:centroid-all-drones-inputs}
    \left[ \begin{array}{c}
        \externalInputLinearAll \\
        \externalInputAngularAll
    \end{array} \right]
    =
    \centroidControlMatrix \centroidUserInputs.
\end{equation}

\subsubsection{Haptic Feedback (H)}
\label{subsubsec:centroid-haptic-feedback}
As the user does not control all the degrees of motion of the agents but rather their position around the centroid, we decompose the connectivity force $\connectivityForcej$ implemented by each agent $j \in \controlledAgents$ into two components:
\begin{itemize}
    \item $\connectivityForceParallell$: The part of the connectivity force the user can control through the inputs presented in Sect.~\ref{subsubsec:centroid-control-inputs};
    \item $\connectivityForceNullSpace$: The part of the connectivity force the user cannot directly control, e.g., the yaw of individual agents.
\end{itemize}
We propose to guide the user with the component $\connectivityForceParallell$ through haptic feedback, while the remaining component $\connectivityForceNullSpace$
will be added to the robot inputs as part of the group autonomy through $\residualconnectivityVelocityj$ and $\residualconnectivityYawRatej$.

From~(\ref{eq:centroid-control-matrix}) we define $\connectivityForceParallell$ as:
\begin{equation}\label{eq:centroid-force-parallel}
    \connectivityForceParallell
    =
    \centroidControlMatrix^\dagger
    \begin{bmatrix}
        \begin{array}{c}
            \connectivityVelocity\\
            \connectivityYawRate
        \end{array}
    \end{bmatrix},
\end{equation}
where $\externalInputLinear$ and $\externalInputAngular$ represent the aggregated linear and angular velocities for the entire group, respectively, and $\centroidControlMatrix^\dagger$ represents the pseudo-inverse of $\centroidControlMatrix$.
Similarly, $\connectivityForceNullSpace$ is defined as:
\begin{equation}\label{eq:centroid-force-nullspace}
    \connectivityForceNullSpace
    =
    (\left[\boldsymbol{I}^{4}\right]^{N} - \centroidControlMatrix\centroidControlMatrix^\dagger)
    \begin{bmatrix}
        \begin{array}{c}
            \connectivityVelocity\\
            \connectivityYawRate
        \end{array}
    \end{bmatrix}.
\end{equation}

Using the control mapping matrix $\centroidMappingMatrix$ from~(\ref{eq:centroid-mapping-matrix}), the force feedback guiding the user is then:
\begin{equation} \label{eq:force-feedback-centroid}
    \left[\begin{array}{c}
        {\forceFeedback} \\
        {\torqueFeedback}
    \end{array} \right]
    =
    \centroidMappingMatrix^\dagger \connectivityForceParallell,
\end{equation}
and the overall residual connectivity inputs $(\residualconnectivityVelocity, \residualconnectivityYawRate)$ are given by:
\begin{equation} \label{eq:centroid-residual-connectivity-inputs}
    \left[ \begin{array}{c}
        \residualconnectivityVelocity \\
        \residualconnectivityYawRate
    \end{array} \right]
    =
    \connectivityForceNullSpace,
\end{equation}
such as
\begin{equation} \label{eq:centroid-full-drones-inputs}
    \left[ \begin{array}{c}
        {\vv} \\
        {\omega}
    \end{array} \right]
    =
    \begin{array}{c}
        \centroidControlMatrix \centroidUserInputs + \connectivityForceNullSpace
    \end{array}.
\end{equation}

\subsubsection{No Haptic Feedback (NH)}
\label{subsubsec:centroid-no-haptic-feedback}
In this case, no additional haptic feedback is provided to the operator; only the joystick-like force $\joystickForce$ (eq.~\ref{eq:virtuose_spring_force}) and torque $\joystickForceR$ (eq.~\ref{eq:virtuose_spring_torque}) are provided.  The user-controlled agent $j$ follows the velocity control input $(\connectivityVelocityj, \connectivityYawRate)$ from the connectivity maintenance algorithm, in addition to the user's inputs, as presented in eq.~(\ref{eq:user-inputs}).

\section{Software Implementation and User Interface}\label{sec:implementation}
\begin{figure}[htp]
    \centering
    \includegraphics*[width=0.95\linewidth]{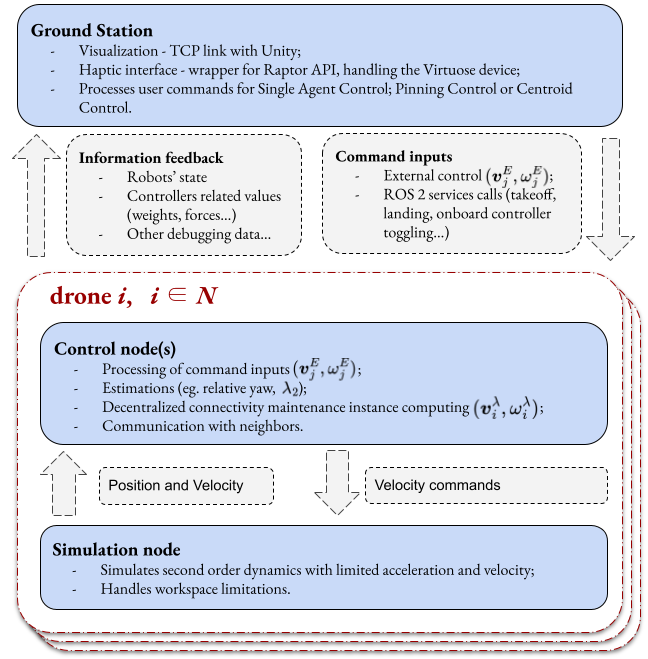}
    \caption{Software implementation. For each simulated drone, one instance of the control and simulation nodes is executed.}
    \label{fig:implementation}
\end{figure}

We use ROS~2 to interconnect two main components: the ground station and the robot group. The functions and integration of the different components are summarized in Figure~\ref{fig:implementation} and detailed in the following subsections. An even more detailed description of the software architecture can be found in Appendix~\ref{app:software}.
We use ROS~2 to interconnect the ground station and the robot group, facilitating communication and coordination between the system's components.
The overall structure is summarized in Fig.~\ref{fig:implementation} and further explained in the following subsections.
For additional implementation details, please refer to Appendix~\ref{app:software}.

\subsection{Ground station}\label{subsec:ground_station}
Given the control methods presented in Sect.~\ref{sec:haptic-guidance}, a centralized ROS~2 node functions as the ground station and serves as a bridge between the robots and the user interface.
On one hand, it interfaces with the Virtuose device to process user inputs and provide connectivity-based haptic feedback, if applicable.
On the other hand, it forwards external control commands to the agents and receives their states for display in the Unity 3D visualization.

The ground station runs several other ROS~2 nodes responsible for operating other functions such as takeoff, emergency stop or landing, logging data for debugging purposes or emulating sensors, such as a limited FOV sensor providing relative positions w.r.t. their neighbors to the robots.

\subsection{Robots control and simulation}
\label{subsec:robots_simulation}
Each robot operates as two individual ROS~2 nodes, as shown in Figure~\ref{fig:implementation}:
\begin{enumerate}
    \item \textbf{Simulation Node}: It implements the dynamics of a second-order system, with a maximum acceleration of 5 m$\cdot$s\textsuperscript{2}.
    It receives control commands as input and outputs the robot's state, including position, velocity and orientation.
    This node runs at 100 Hz.
    \item \textbf{Control Node}: It implements the decentralized connectivity maintenance algorithm and handles the necessary estimations and communication protocols.
    It also integrates external control inputs from the user with the connectivity-maintenance control.
    This node runs at 30 Hz.
\end{enumerate}

\subsection{Visualization}
\label{subsec:visualization}
We use the \emph{ROS TCP Connector} Unity plugin~\cite{unity2022rostcpconnector} and the \emph{ROS TCP Endpoint} ROS~2 package~\parencite{unity2022rostcpendpoint} to connect the ROS~2 control and simulation nodes with the Unity visualization through a TCP bridge.

\begin{figure}[htp]
    \centering
    \includegraphics*[width=0.95\linewidth]{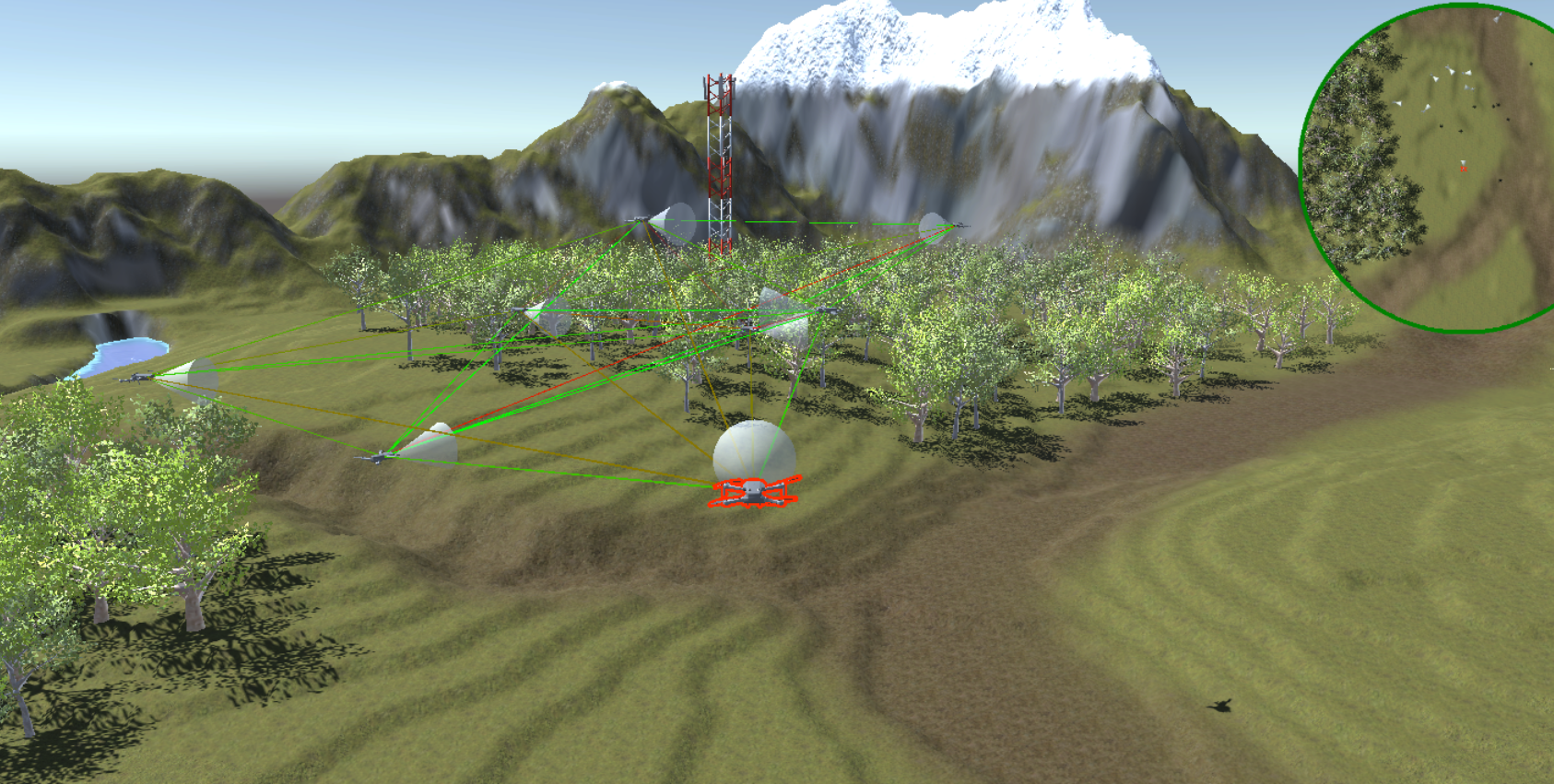}
    \caption{``Plains'' environment. Unity 3D visualization of the robots when using the Single Agent Control method. The leader robot is highlighted.}
    \label{fig:visualization_plains}
\end{figure}

\begin{figure}[htp]
    \centering
    \includegraphics*[width=0.95\linewidth]{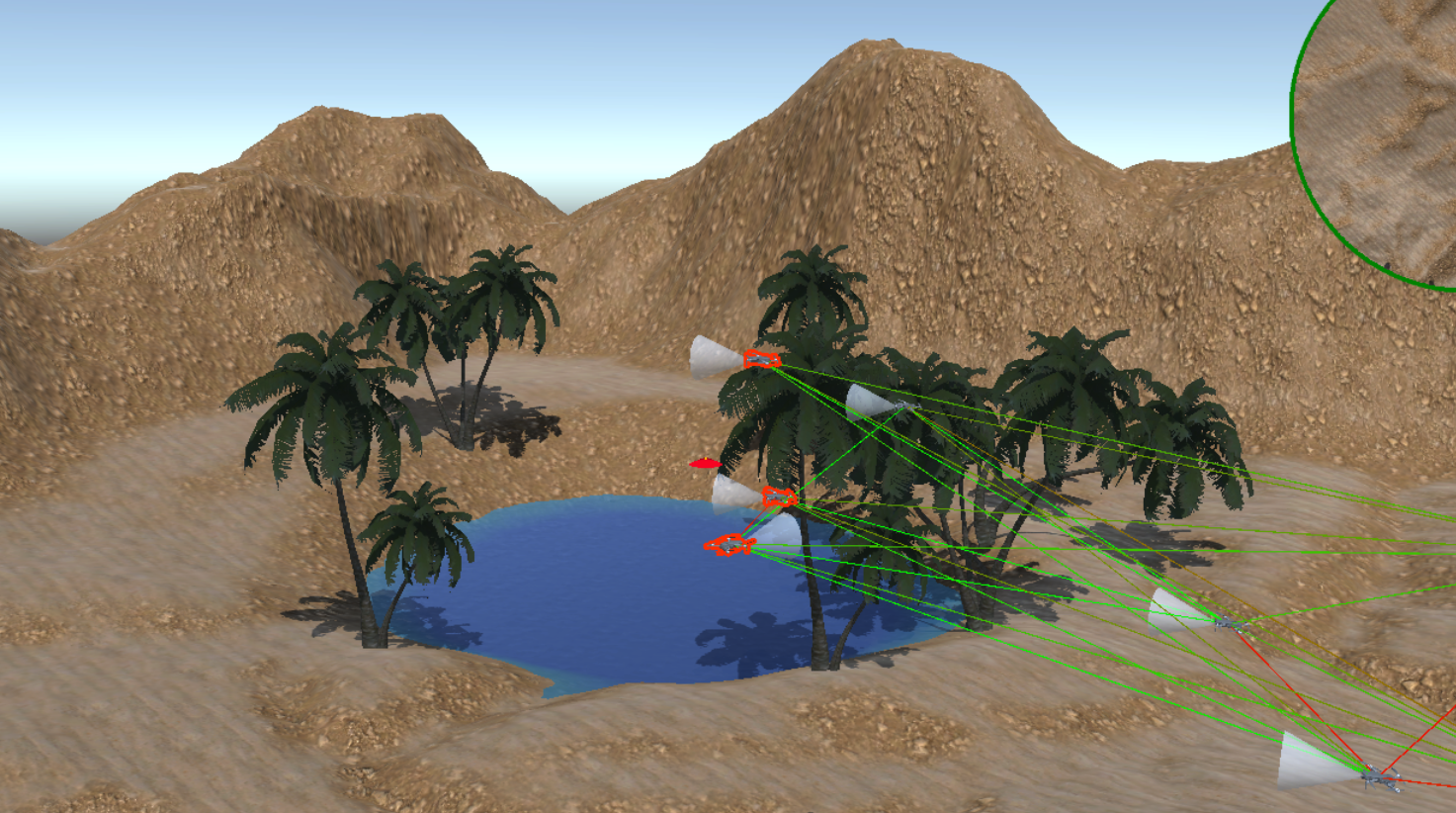}
    \caption{``Desert'' environment. Unity 3D visualization of the robots when using the Pinning Control method. The pinned robots are highlighted.}
    \label{fig:visualization_desert}
\end{figure}

\begin{figure}[htp]
    \centering
    \includegraphics*[width=0.95\linewidth]{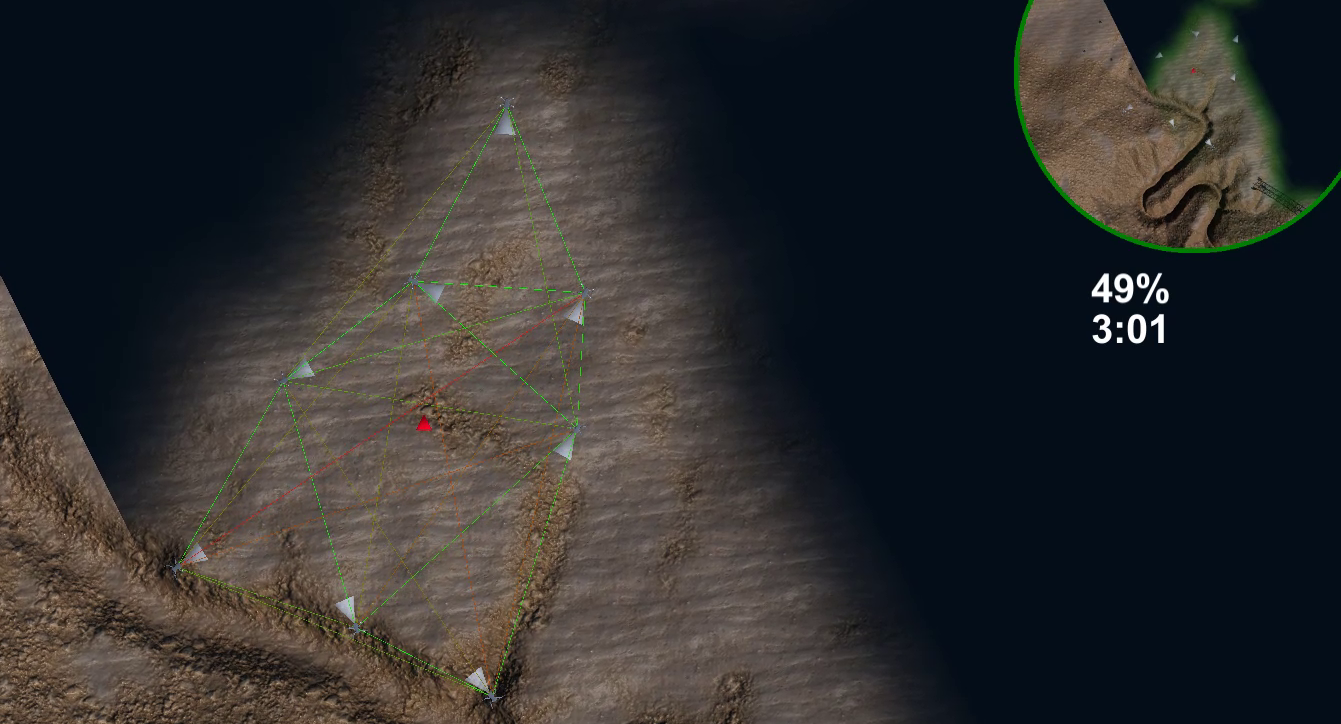}
    \caption{Unity 3D visualization of the robots when using the Centroid Control method. The centroid of the group is depicted by the red arrow.}
    \label{fig:visualization_centroid}
\end{figure}

In a virtual Unity 3D environment, each robot is represented by a UAV 3D model, as shown in Figures~\ref{fig:visualization_plains}, \ref{fig:visualization_desert} and ~\ref{fig:visualization_centroid}.

In addition to providing visual feedback, the Unity interface also allow for user interaction with the robots.
For example, during Pinning Control, the user can select or deselect \emph{pinned} agents by clicking on them within the Unity interface.
Depending on the control method, the third-person camera follows either a single agent, the virtual pinning point, or the centroid of the group.
At all times, the user can rotate the camera around the tracked object and zoom in or out using the mouse.

\subsection{Experimental Scenarios}
\label{subsec:experimental-scenarios}

This Section describes how we leverage the visualized simulated environment to test the different control methods. We have implemented three distinct scenarios, each presented in the video available at \url{https://youtu.be/gDR6Lky8_OI}:
\begin{enumerate}
    \item \textbf{Single Agent Control} (\emph{S}, see Sect.~\ref{subsec:single-agent-control}): Each robot is equipped with one of three types of sensors, distinguished by different colors. There are six target locations, and reaching one reveals the next in sequence. Each target must be visited by a robot equipped with a specific sensor type. The user controls the appropriate robot to reach the designated target location, viewing the scene through a third-person camera attached to the controlled robot. Video: \url{https://youtu.be/gDR6Lky8_OI?t=34}.
    \item \textbf{Pinning Control} (\emph{P}, see Sect.~\ref{subsec:pinning-control}): Similarly as before, each robot is equipped with one of three types of sensors, distinguished by different colors. Here, each objective requires two robots, each with a specific and distinct sensor type, to reach the target. The user controls the pinning point to make the two designated robots reach the current target location, viewing the scene through a third-person camera attached to the virtual pinning point. Video: \url{https://youtu.be/gDR6Lky8_OI?t=177}.
    \item \textbf{Centroid Control} (\emph{C}, see Sect.~\ref{subsec:centroid-control}): At the start of the task, the environment is hidden by a dark fog that gradually clears as the robotic team moves across the map. Certain areas are marked in red and must be avoided. The user controls the robotic team to uncover as much of the environment as possible, viewing the scene from a top-down perspective with an orthographic projection. Video: \url{https://youtu.be/gDR6Lky8_OI?t=334}.
\end{enumerate}

To investigate the effect of haptic feedback on user performance depending on the control strategy, users repeat the given mission on each scenario twice: once with (\emph{H}) and once without (\emph{NH}) haptic feedback.
To mitigate learning effects from repeating the same mission in identical visual settings, we have developed two visually distinct environments, which we called ``Plains''  (see Fig.~\ref{fig:visualization_plains}) and ``Desert'' (see Fig.~\ref{fig:visualization_desert}).
The two environments are visually different, but the target objectives to visit are placed in the same locations.

\section{User study}
\label{sec:user-study}
To evaluate the effectiveness and usability of the proposed control and feedback methods, we carried out a user study enrolling 20 participants.

\subsection{Methods}
\label{subsec:methods}

\subsubsection{Setup and Task}

\begin{table}[htp]
     \renewcommand*{\arraystretch}{1.2}
     \centering
        \begin{tabular}{|c|l|c|c|c|}
        \hhline{~~---|}
        \multicolumn{1}{l}{}
        & & \multicolumn{3}{c|}{{\cellcolor[rgb]{1,1,0.78}}\textbf{Control conditions}}\\
        \hhline{~~---|}
        \multicolumn{1}{l}{} & \textbf{}   & \multicolumn{1}{l|}{{\cellcolor[rgb]{1,1,0.78}}\textbf{Single Agent}} & \multicolumn{1}{l|}{{\cellcolor[rgb]{1,1,0.78}}\textbf{Pinning}} & \multicolumn{1}{l|}{{\cellcolor[rgb]{1,1,0.78}}\textbf{Centroid}}  \\
        \hline
        {\cellcolor[rgb]{0.855,0.91,0.988}} & {\cellcolor[rgb]{0.855,0.91,0.988}}\begin{tabular}[c]{@{}>{\cellcolor[rgb]{0.855,0.91,0.988}}l@{}}\textbf{without}\\\textbf{Haptic}\\\textbf{Feedback}\end{tabular} & SNH & PNH & CNH \\
        \hhline{|>{\arrayrulecolor[rgb]{0.855,0.91,0.988}}->{\arrayrulecolor{black}}----|}
        \multirow{-2}{*}{{\cellcolor[rgb]{0.855,0.91,0.988}}\rotcell{\begin{tabular}[c]{@{}>{\cellcolor[rgb]{0.855,0.91,0.988}}c@{}}\textbf{Feedback}\\\textbf{conditions}\end{tabular}}} & {\cellcolor[rgb]{0.855,0.91,0.988}}\begin{tabular}[c]{@{}>{\cellcolor[rgb]{0.855,0.91,0.988}}l@{}}\textbf{with}\\\textbf{Haptic}\\\textbf{Feedback}\end{tabular} & SH & PH & PNH \\
        \hline
        \end{tabular}
     \medskip
     \caption{Name of the different control and feedback conditions.}
     \label{tab:experimental-scenarios}
\end{table}

\begin{figure}[htp]
    \centering
    \includegraphics[width=0.95\linewidth]{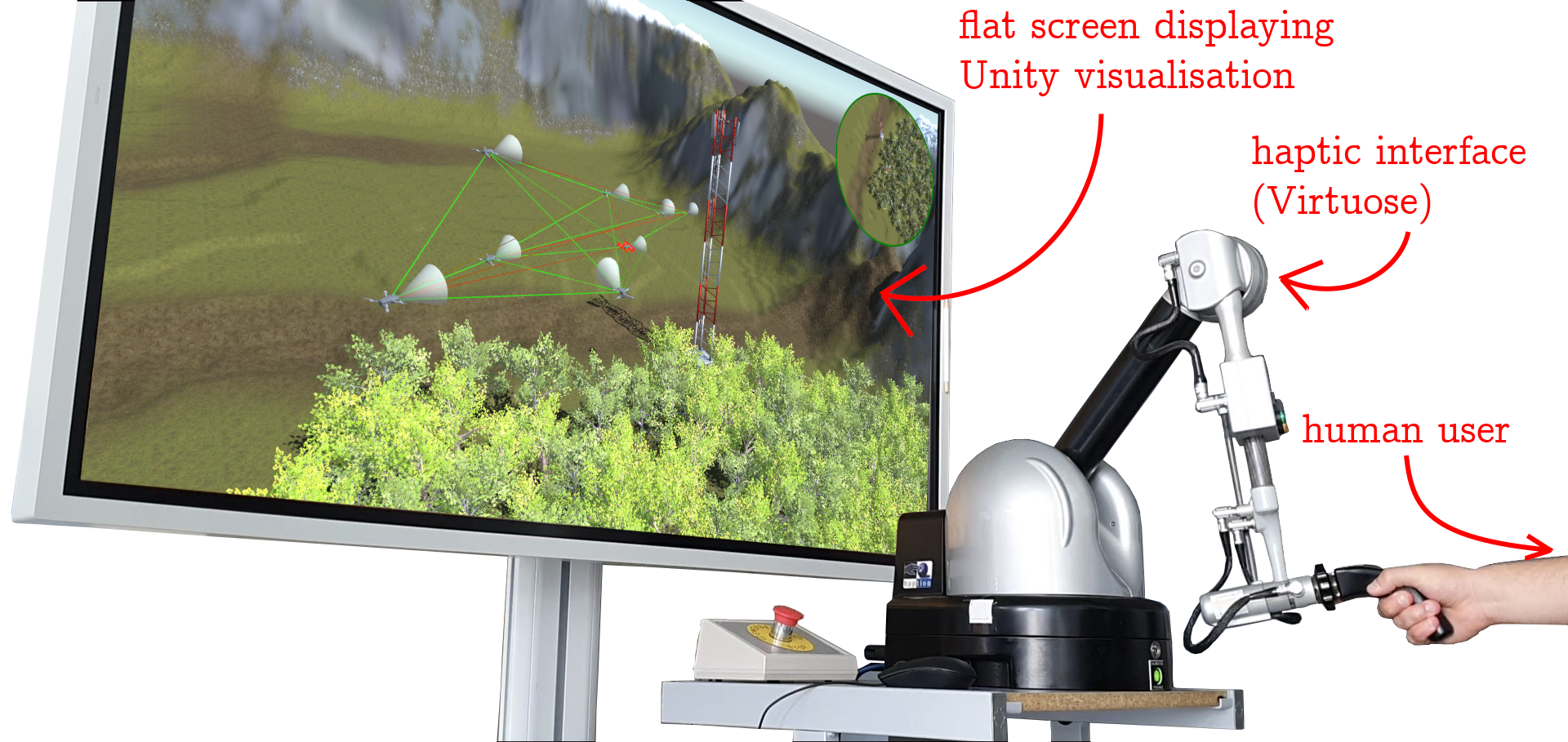}
    \caption{Experimental setup for the user study. The environment is displayed on a large screen. The user controls the agents using the Virtuose haptic interface.}
    \label{fig:experimental_setup}
\end{figure}

Participants are asked to seat in front of a large TV screen and hold the end-effector of the Virtuose 6D device to control the multi-robot team (see Fig.~\ref{fig:experimental_setup}).

The experiment is divided into three blocks, one for each of the experimental scenarios described in Sec.~\ref{subsec:experimental-scenarios}, presented in the same sequence for all users: Single Agent Control (S), Pinning Control (P), and Centroid Control (C).
Each scenario is then repeated once with haptic feedback (\emph{H}) and once without haptic feedback (\emph{NH}).
The order of presentation of the haptic feedback (\emph{H} or~\emph{NH}) and visual environment (Desert and Plains) was counterbalanced among participants.

At the beginning of each block, we carried out a training session in an empty environment, so as to make users familiar with the considered control method and haptic feedback.

\subsubsection{Collected Data}

We recorded the user inputs and the force feedback provided.
In scenarios~\emph{S} and~\emph{P}, we measured the task completion time; shorter times were interpreted as indicative of a more effective control method.
In scenario~\emph{C}, we measured the area explored by the user; a larger discovered area was interpreted as indicative of a more effective control method.
Prior to the experiment, participants completed a demographic questionnaire.

At the end of each block, participants filled out a subjective questionnaire to evaluate their experience with the control method and the haptic feedback system.
This evaluation included two aspects: perceived effectiveness in achieving the task and the intuitiveness of the control method.

\subsubsection{Population}
The study involved 20 participants (16 M, 4 F) aged 18 to 35 years (m: 25.5y, sd: 4.5y).
Of these participants, 18 were right-handed and 2 were left-handed.
Half of the participants were experienced with haptics, while the other half were not.
Similarly, half had experience with multi-robot systems, while the other half did not.
Additionally, two participants had very limited experience with virtual 3D environments (such as video games), whereas the remaining 18 participants had extensive experience.

\subsection{Results}
\label{subsec:haptic_results}

\subsubsection{Single Agent Control}

\begin{figure}[htp]
  \centering
  \subfloat[Questionnaire results\label{fig:poll_single}]{\includegraphics[width=0.55\linewidth]{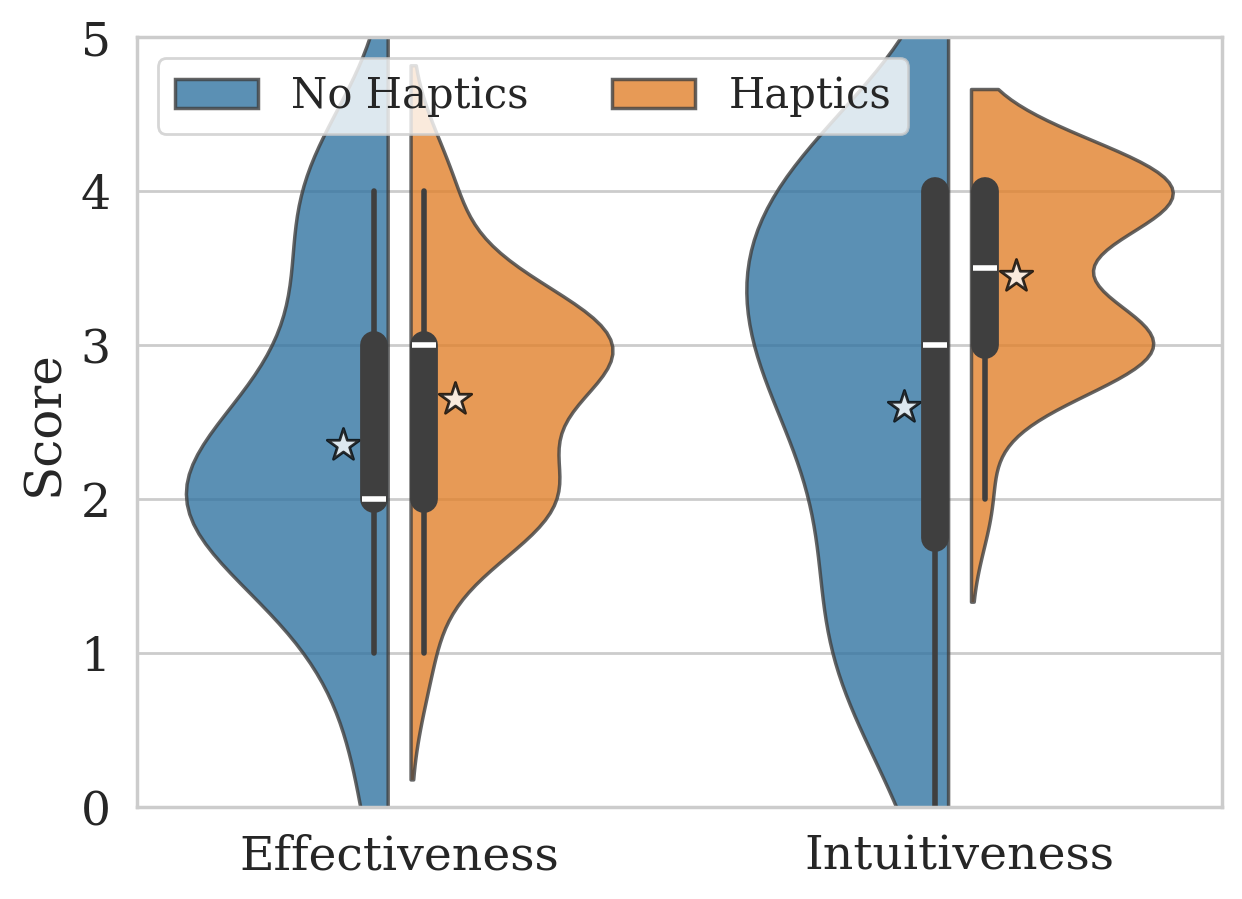}}
  \subfloat[Completion time in seconds.\label{fig:ct_single}]{\includegraphics[width=0.45\linewidth]{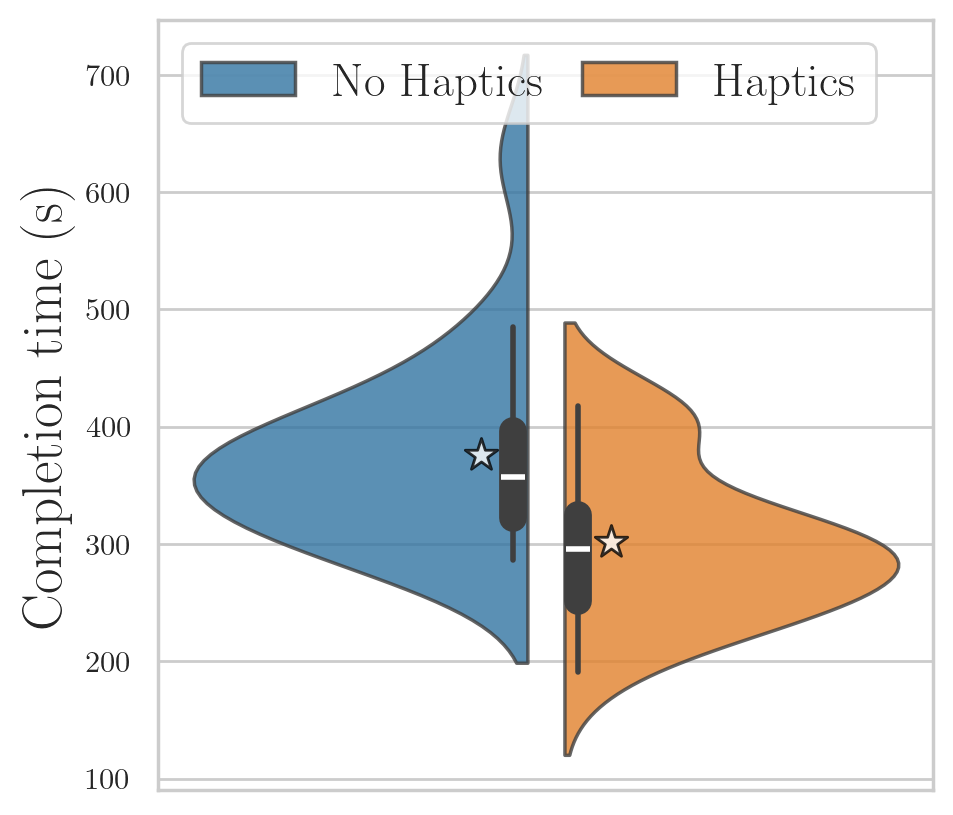}}
  \vspace{9pt}
  \caption{Experimental results. (a) User perceived effectiveness and intuitiveness, and (b) Measured completion time for the Single Agent Control scenario.}%
\end{figure}

Figure~\ref{fig:poll_single} presents the results of user-rated effectiveness and intuitiveness for this scenario (higher scores indicate better user experience). A paired-samples t-test determined that the registered effectiveness differed statistically significantly between feedback conditions (\emph{NH} vs. \emph{H}, t(19) = -2.342, p = 0.030, \emph{H} is better, confidence interval of 95\%). No statistically significantly difference was registered for the intuitiveness metrics.
Figure~\ref{fig:ct_single} shows the measured completion time (lower is better).
A paired-samples t-test determined that the completion time differed statistically significantly between feedback conditions (\emph{NH} vs. \emph{H}, t(19) = -4.296, p $<$ 0.001, \emph{H} is better, confidence interval of 95\%).

As discussed in Sec.~\ref{sec:haptic-guidance}, in the~\emph{NH} condition, the controlled agent follows the connectivity maintenance control command with a reduced gain compared to the other agents, providing the user with greater control flexibility.
In contrast, in the~\emph{H} condition, the controlled agent exclusively follows the user's input, which likely accounts for the improved completion time.

\subsubsection{Pinning Control}

\begin{figure}[htp]
  \centering
  \subfloat[Poll results.\label{fig:poll_pinning}]{\includegraphics[width=0.55\linewidth]{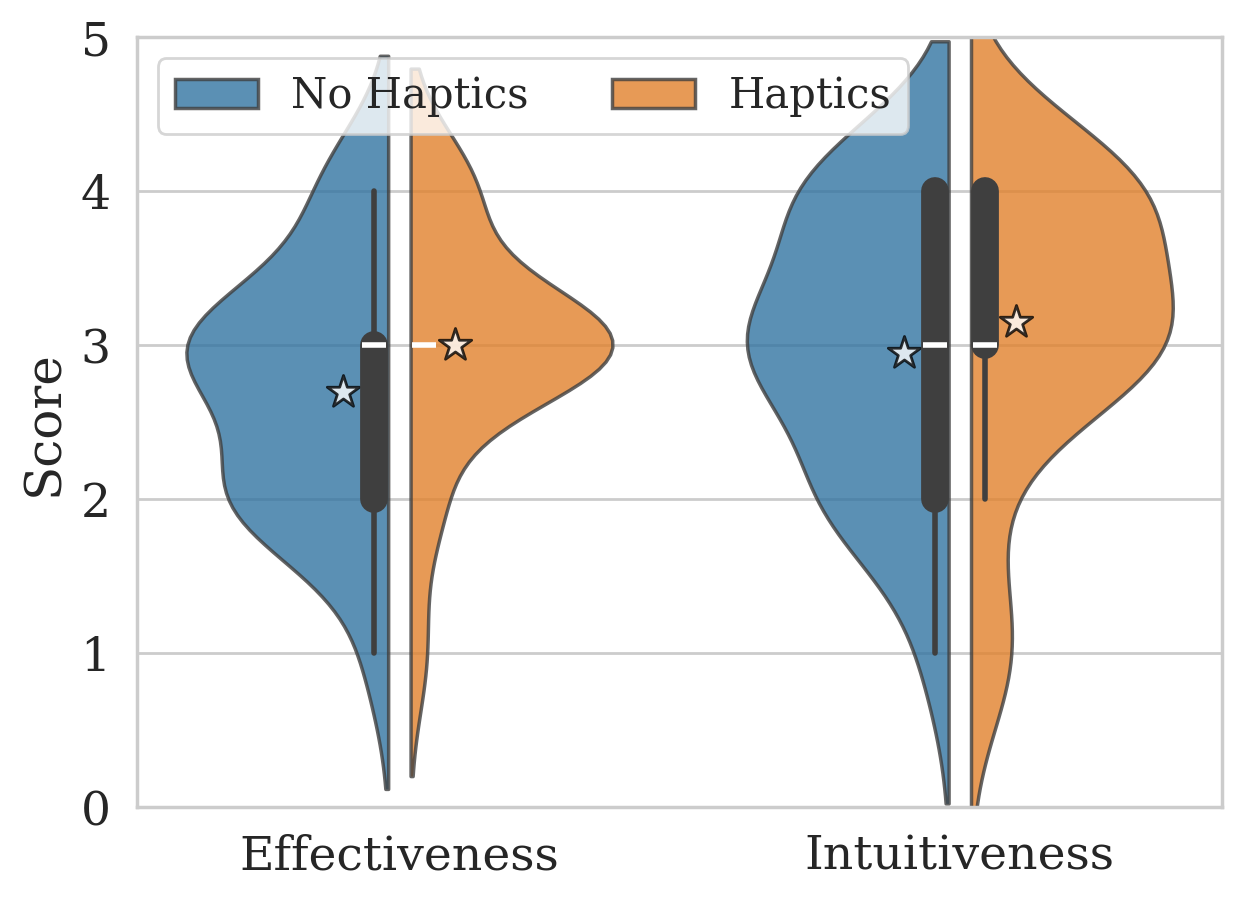}}
  \subfloat[Completion time in seconds.\label{fig:ct_pinning}]{\includegraphics[width=0.45\linewidth]{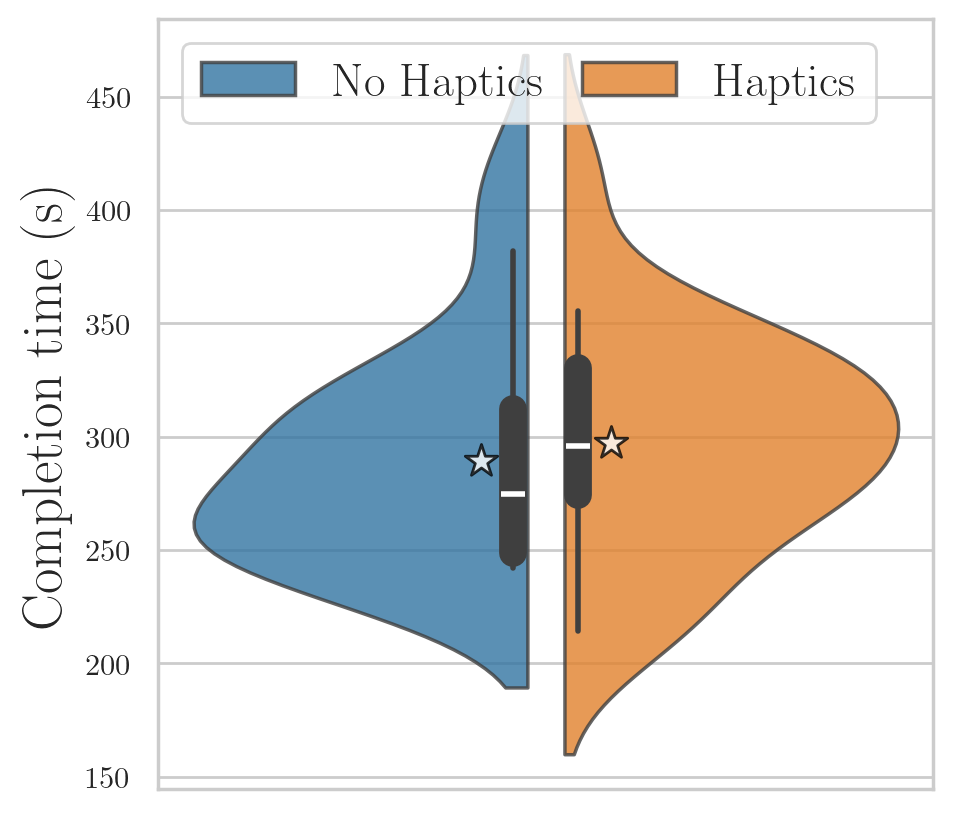}}
  \vspace{9pt}
  \caption{User Perceived Effectiveness and Intuitiveness, and Measured Completion Time for Pinning Control Experiment.}%
\end{figure}

Figure~\ref{fig:poll_pinning} presents the results of user-rated effectiveness and intuitiveness for this scenario (higher scores indicate better user experience). A paired-samples t-test determined that the registered effectiveness differed statistically significantly between feedback conditions (\emph{NH} vs. \emph{H}, t(19) = -2.179, p = 0.042, \emph{H} is better, confidence interval of 95\%). No statistically significantly difference was registered for the intuitiveness metrics.
Figure~\ref{fig:ct_pinning} shows the measured completion time (lower is better). No statistically significantly difference was registered for this metrics.

\subsubsection{Centroid Control}

\begin{figure}[htp]
  \centering
  \subfloat[Poll results.\label{fig:poll_centroid}]{\includegraphics[width=0.55\linewidth]{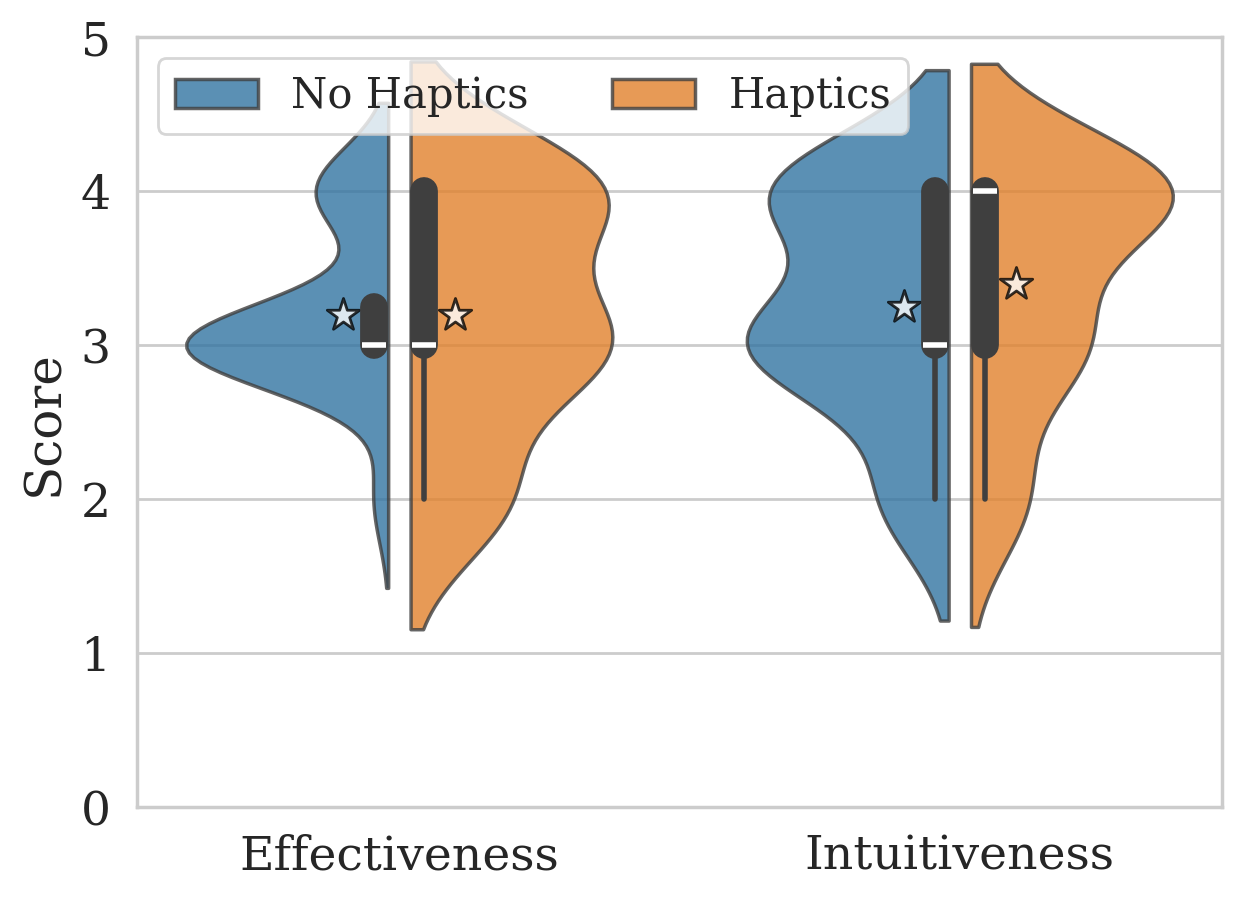}}
  \subfloat[Proportion of Discovered Environment.\label{fig:ct_centroid}]{\includegraphics[width=0.45\linewidth]{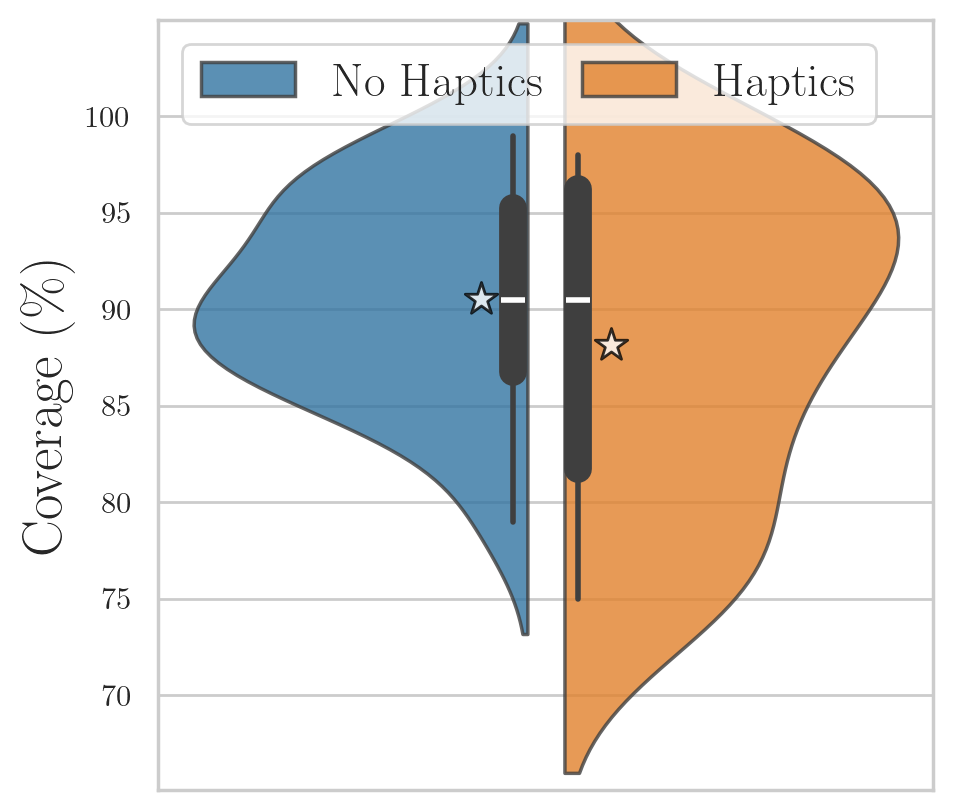}}
  \vspace{9pt}
  \caption{User Perceived Effectiveness and Intuitiveness, and Proportion of Discovered Environment for Centroid Control Experiment.}%
\end{figure}

Figure~\ref{fig:poll_centroid} presents the results of user-rated effectiveness and intuitiveness for this scenario (higher scores indicate better user experience). A paired-samples t-test determined that the registered intuitiveness differed statistically significantly between feedback conditions (\emph{NH} vs. \emph{H}, t(19) = -2.773, p = 0.012, \emph{H} is better, confidence interval of 95\%). No statistically significantly difference was registered for the effectiveness metrics.
Figure~\ref{fig:ct_centroid} shows the measured completion time (lower is better).
A paired-samples t-test determined that the discovered environment differed statistically significantly between feedback conditions (\emph{NH} vs. \emph{H}, t(19) = -2.255, p = 0.036, \emph{H} is better, confidence interval of 95\%).

\section{Use case}
\label{sec:use-case}
We show the robotic team carrying out a more complex recognition task, seamlessly switching from one control modality to the other. The video of this use case is available at \url{https://youtu.be/gDR6Lky8_OI?t=463}.

A human operator is tasked with controlling the multi-robot team to reach multiple objectives in a large map using different control methods.
The mission begins with the operator using the Centroid Control method to quickly guide the entire robot group to the target area.
Upon arrival, the operator switches to Single Agent Control to identify the areas to visit.
Once the objectives are located, the operator employs Pinning Control to direct multiple agents to the target locations and execute the tasks.
After visiting all the objectives, the operator returns to Centroid Control to quickly guide the group back to the home base.

\begin{figure}[htp]
    \centering
    \includegraphics*[width=0.6\linewidth]{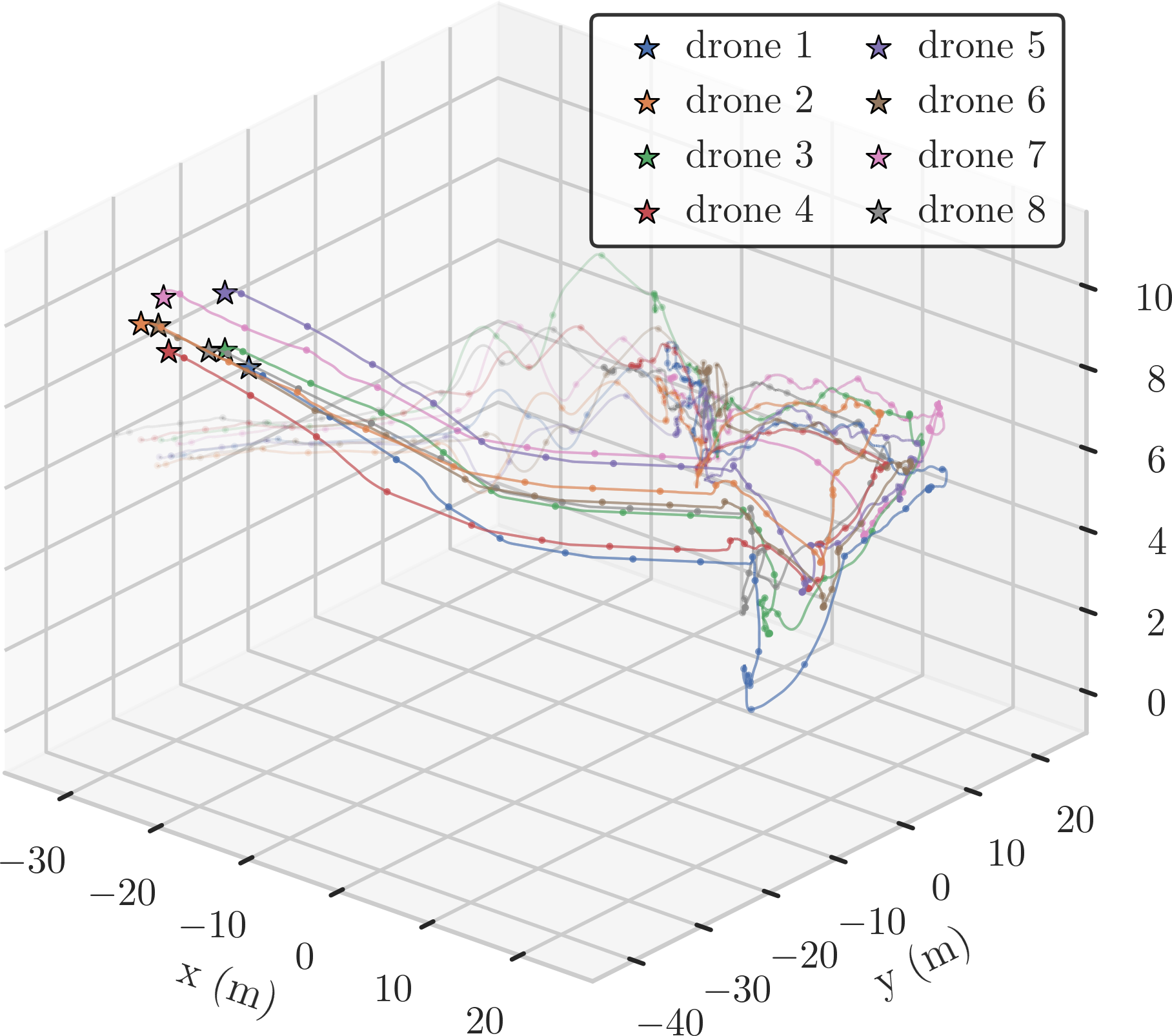}
    \caption{Agents trajectories throughout the experiment. Each dot in the trajectory corresponds to the position of an agent at a 5 seconds interval.}
    \label{fig:useCaseTrajectorySide}
    \includegraphics*[width=0.6\linewidth]{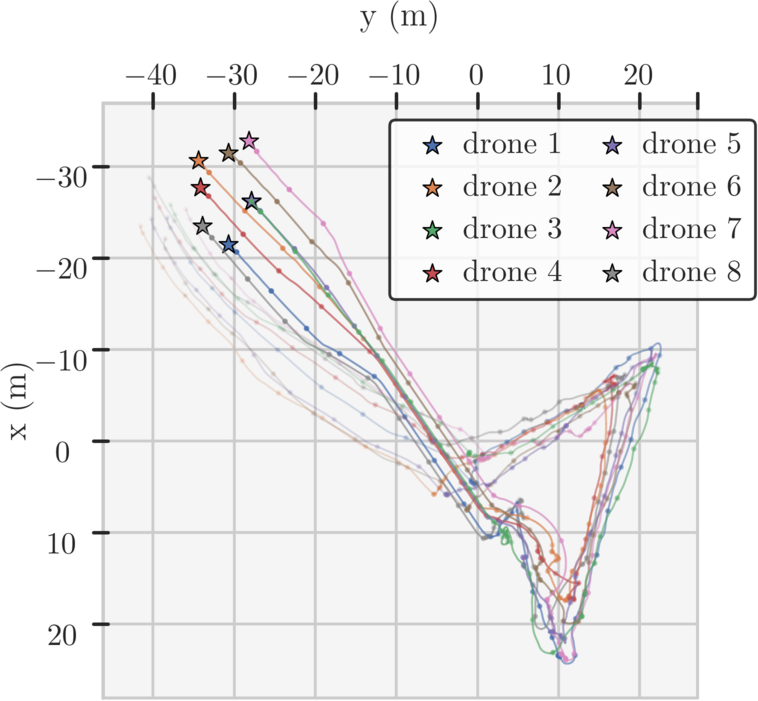}
    \caption{Top view of the agents trajectories throughout the experiment.}
    \label{fig:useCaseTrajectoryTop}
\end{figure}

Figures~\ref{fig:useCaseTrajectorySide} and \ref{fig:useCaseTrajectoryTop} shows the trajectories of the eight drones throughout the task.
Star markers indicate the final position of each agent, while the fading lines represent their trajectory, with older positions appearing more transparent.
Each dot along the trajectory corresponds to the agent's position at 5-second intervals.

\begin{figure}[h]
    \centering
    \includegraphics*[width=0.9\linewidth]{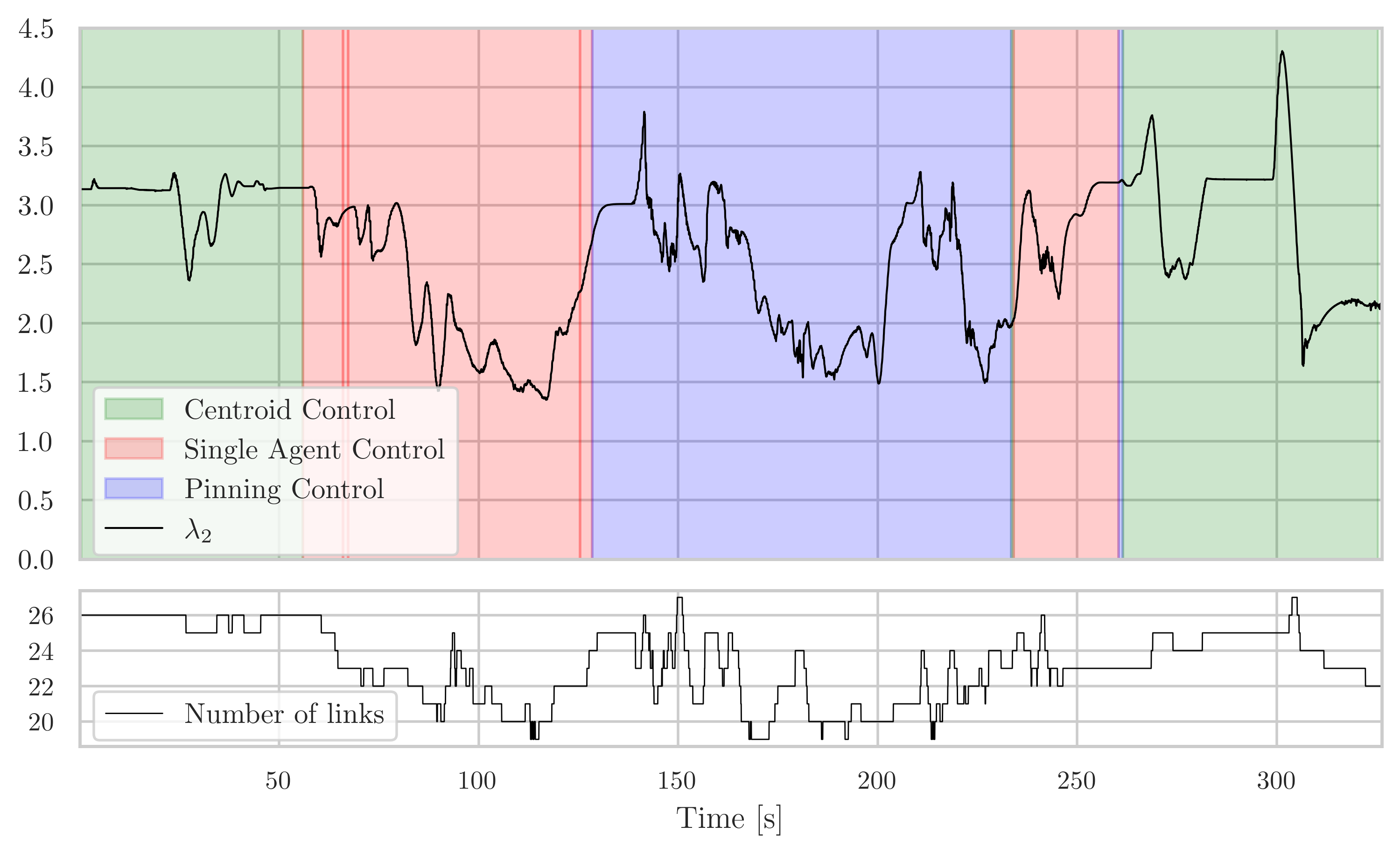}
    \caption{The shape of $\lambda_2$ throughout the experiment and the number of connected agents.}
    \label{fig:useCaseLambda2}
\end{figure}

Figure~\ref{fig:useCaseLambda2} (top) illustrates the evolution of the second eigenvalue of the Laplacian matrix $\lambda_2(t)$ throughout the experiment, colored patches indicate changes in the used control method.
Figure~\ref{fig:useCaseLambda2} (bottom) presents the evolution of the number of connected agents.
As depicted in the figures, the connectivity maintenance algorithm successfully maintains the connectivity of the multi-robot team throughout the mission, even when the control methods change.
Additionally, we observe the dynamic removal and creation of edges, demonstrating the flexibility and robustness of the connectivity maintenance algorithm.

\section{Conclusions}
In this chapter, we proposed and evaluated a series of haptic shared control techniques for the teleoperation of multi-robot aerial systems.
Specifically, we implemented and assessed three control paradigms: \textit{Single Agent Control}, \textit{Pinning Control}, and \textit{Centroid Control}, each designed to address different aspects of multi-robot teleoperation. These approaches featured innovative haptic guidance mechanisms to inform the operator about the team's connectivity state and task dynamics.

The experimental results from the user study involving 20 participants demonstrated the overall benefit of employing haptic feedback. For instance, in the Single Agent Control scenario, the use of haptic guidance significantly reduced task completion times, highlighting the efficacy of haptic cues in providing situational updates.
Subjective feedback indicated an improvement in perceived effectiveness and intuitiveness when haptic feedback was utilized. This was particularly evident in the Centroid Control mode, where participants rated the haptic-enabled interface as significantly more intuitive compared to the non-haptic variant.
Overall, the designed control methods effectively maintained global swarm connectivity under varying conditions, including transitions between control modes. The use case demonstrated the seamless transition between control modalities, showcasing the framework's adaptability to mission requirements. This flexibility supports a wide range of applications, from search and rescue to environmental exploration.

The presented framework addresses some important challenges in human-swarm interaction, particularly the need for intuitive and responsive control mechanisms for large-scale robotic teams. By leveraging haptic feedback, operators can maintain high levels of situational awareness while managing complex tasks. The findings validate the hypothesis that haptic feedback not only enhances task efficiency but also reduces cognitive load by providing interpretable cues about system states.

\clearemptydoublepage
\backmatter
\part{Conclusions and Future Work}
\label{part:conclusions}
\graphicspath{{Conclusion/figures/}}

\lettrine{I}{n} this final part, we aim to summarize the main theoretical and experimental contributions of this thesis, highlighting the key results achieved. Additionally, we will discuss the limitations and unresolved challenges encountered during the course of this work. For these open issues, we propose potential directions for future investigation and research.

\section*{Summary and contributions}
\addstarredsection{Summary and contributions}

The primary objective of this thesis was to explore haptic shared control for multi-robot systems, emphasizing the development of human-robot interfaces that harness the complementary strengths of humans and robots. We sought to address challenges arising from sensing constraints and incomplete relative measurements, while relaxing the strict requirement of rigidity of the formation, thereby granting the robots greater freedom of movement.

Building on the concept of formation control under global constraints (Section~\ref{section:connectivityMaintenanceIntro}), this thesis introduced an extension to connectivity maintenance that addresses field-of-view limitations and body-frame measurements (Section~\ref{sec:conn_maint}). Additionally, a novel method for estimating relative orientations among sensed agents was proposed (Section~\ref{sec:psi_estimation}). This approach enables robots to autonomously adjust pair-wise distances and orientations to maintain a globally connected communication and sensing graph, as demonstrated through simulated experiments (Section~\ref{sec:conn_maint_results}).

After establishing the feasibility of the connectivity maintenance approach, it was integrated into a haptic shared control framework, allowing human operators to guide robots via three distinct control architectures: single-leader control (Section~\ref{subsec:single-agent-control}), multiple-leader control (Section~\ref{subsec:pinning-control}), and formation centroid control (Section~\ref{subsec:centroid-control}). The effectiveness of this framework was validated through user studies comparing the shared control interface with haptic feedback to a teleoperation interface without haptic feedback (Section~\ref{sec:user-study}). These studies highlighted increased efficiency when controlling a single leader and greater intuitiveness when managing the formation centroid (Section~\ref{subsec:haptic_results}). Furthermore, a use case demonstrated seamless transitions between control modalities, enabling operators to adapt dynamically to mission requirements (Section~\ref{sec:use-case}).

\section*{Limitations and future perspectives}
\addstarredsection{Limitations and future perspectives}

The results of this work highlight the potential of haptic shared control in enhancing the operation of multi-robot systems in dynamic environments with realistic assumptions on sensor capabilities and communication constraints.
However, several limitations and challenges remain.

The current implementation of haptic feedback is limited to force feedback, overlooking the potential advantages of incorporating tactile feedback. Enhancing the interface with additional sensory modalities could significantly improve the operator's situational awareness and task performance.
For instance, the \textit{DORNELL} project (\url{https://www.inria.fr/fr/dornell}) aims to design a multi-actuator vibrotactile handle intended to guide visually impaired individuals.
Such handle could be integrated with the Virtuose device, greatly improving the system's capability to convey information to the operator \parencite{cabaret2024design, cabaret2023does}.

Furthermore, the user study was conducted in a controlled environment with a relatively small number of participants, which may not adequately capture the variability in user preferences and behaviors.
Conducting a broader study with a larger and more diverse participant pool, as well as investigating long-term learning effects, would offer a more comprehensive evaluation of the proposed haptic shared control framework.
Additionally, exploring collaborative control by multiple operators using haptic interfaces could expand the framework's applicability to larger and more complex missions.

Several aspects of the algorithms developed in this thesis could be further investigated to enhance their performance and robustness.
The connectivity maintenance algorithm presented in Sect.~\ref{sec:conn_maint} relies on a simplified motion model that does not account for the roll and pitch motions of quadrotors required for acceleration.
In scenarios involving agile or aggressive maneuvers, such tilting can cause onboard cameras to lose sight of neighboring robots, preventing relative position measurements.
Overcoming this limitation would require extending the model to account for the impact of tilting on the field of view.

Moreover, the connectivity maintenance algorithm cannot recover from a temporary loss of global connectivity.
If the second smallest eigenvalue of the Laplacian matrix becomes zero due to external perturbations, latencies or other factors, the formation splits into disconnected components.
In such cases, the algorithm cannot autonomously restore a globally connected graph without external intervention.
A recent work by~\cite{restrepo2024distributed} proposes a solution to address this limitation, enabling the addition or removal of agents within the connected formation.

To account for field-of-view constraints, the connectivity algorithm presented in this thesis does not include line-of-sight occlusion and obstacle avoidance capabilities, which were included in the algorithm originally proposed by \cite{robuffo2013passivity}.
Future work could focus on integrating these features into the connectivity maintenance algorithm to enhance the robots' safety and improve the system's robustness in cluttered environments.

However, due to its design, the algorithm can account for heterogeneous field-of-view among agents. This feature was tested, although not formally documented, in a scenario where one agent of the team had a null field-of-view, rendering it incapable of measuring neighbors' relative positions. Nevertheless, by leveraging communication and the visual perception of other agents, the blind agent maintained connectivity with the team. This capability is particularly relevant for heterogeneous teams of robots, where UAVs equipped with advanced sensors can supervise ground agents while maintaining connectivity to ensure communication and coordination in areas where direct communication with a remote operator is not possible. Such fully autonomous systems offer significant potential for applications in search and rescue missions, where heterogeneous robot teams can maintain communication and sensing capabilities while efficiently exploring unknown environments.

In scenarios where remote operator intervention is required, the connectivity maintenance algorithm can still play a crucial role in ensuring local communication and sensing capabilities among the robots.
The following remarks offer insights into the practical feasibility of remotely controlling a group of robots in realistic scenarios using the proposed framework.

\begin{remark} \label{remark:one}
    When robot $i$ is connected to robot $j$ as per the weights defined in Section~\ref{sec:conn_maint}, direct communication between them is feasible. Robot $i$ can measure or estimate the relative state $(\RtoAgent{i}\agentPos_{ij}, \agentYaw{ij})$ in a decentralized manner, and the same applies symmetrically for robot $j$.
\end{remark}

\begin{remark} \label{remark:two}
    Maintaining global connectivity facilitates the establishment of an ad-hoc communication network among the robots. This ensures that any agent can communicate with any other agent in the group via a multi-hop network, resulting in a fully connected communication graph.
\end{remark}

\begin{remark} \label{remark:three}
    Cooperative localization algorithms, such as the adaptive observer described in~\cite{decarli2024adaptive}, allow the robots' states to be estimated in a shared frame through the local network. Consequently, a ground station with access to any agent in the group can generate and display a 3D visualization of the entire team.
\end{remark}

\begin{remark} \label{remark:four}
    If the environment is pre-mapped or constructed collaboratively using decentralized SLAM algorithms, the resulting map can be transmitted to the ground station and visualized by the user.
\end{remark}

\begin{remark} \label{remark:five}
    Leveraging the above remarks and assuming reasonable communication delays, the ground station can limit direct communication to a subset of agents, such as the one(s) actively controlled by the user, while still maintaining a comprehensive view of the group's state and environment.

    \emph{Extension:} In scenarios where only one agent has the capability to communicate with the ground station (e.g., due to advanced long-range communication capabilities), the previous remarks still hold. However, this setup imposes a heavier communication load on the ad-hoc network, which may require optimization strategies to maintain system efficiency.
\end{remark}

These considerations pave the way for sophisticated real-world experiments, where the robotic team operates with full autonomy.
The human operator oversees the mission through a user-friendly interface, gaining access to the robots' state and environmental data without requiring direct communication with each individual robot.

Connectivity maintenance is, in our opinion, a key feature for decentralized multi-robot systems, as, with appropriately designed weights modeling inter-agent constraints, it ensures that robots maintain physical requirements.
The algorithm proposed in this thesis guarantees that both the communication and sensing graphs remain globally connected, enabling the integration of other decentralized, task-specific algorithms that rely on global connectivity such as coverage, team state estimation or exploration and mapping.
In advanced scenarios requiring human intervention to resolve deadlocks, make decisions, or adapt to unforeseen events, the haptic shared control framework can assist the user in comprehending the situation and guiding robots efficiently and intuitively.

\part*{Appendix}
\label{part:appendix}
\addcontentsline{toc}{part}{\nameref{part:appendix}}
\appendix
\clearemptydoublepage
\clearemptydoublepage
\graphicspath{{Chapitre98/figures/}}
\chapter{Decentralized Software Implementation}\label{app:software}
{\hypersetup{linkcolor=black}
\minitoc
}

\lettrine{I}{n} this appendix, we present the software architecture and implementation details of the decentralized multi-agent human interaction system developed for this thesis. The system is designed to operate in a simulated environment, with minimal modifications required for deployment in real-world experiments.

Built on the Robot Operating System 2 (ROS~2) middleware, the software provides a flexible and modular framework for robotic applications. It consists of three main components: the onboard software running on each agent, the ground station, and a Unity-based graphical interface. The onboard software handles the autonomous control of the agents, while the ground station acts as the central hub for human-multi-robot interaction. The Unity environment offers a user-friendly interface for monitoring and controlling the agents.

Designed with modularity in mind, the system allows for seamless integration of new controllers, sensors, and communication protocols, making it adaptable to various applications. However, in this appendix, we focus solely on the specific use case showcased in this thesis.

\section{Onboard software}

\begin{figure}[ht]
    \centering
    \def\svgwidth{\linewidth}
    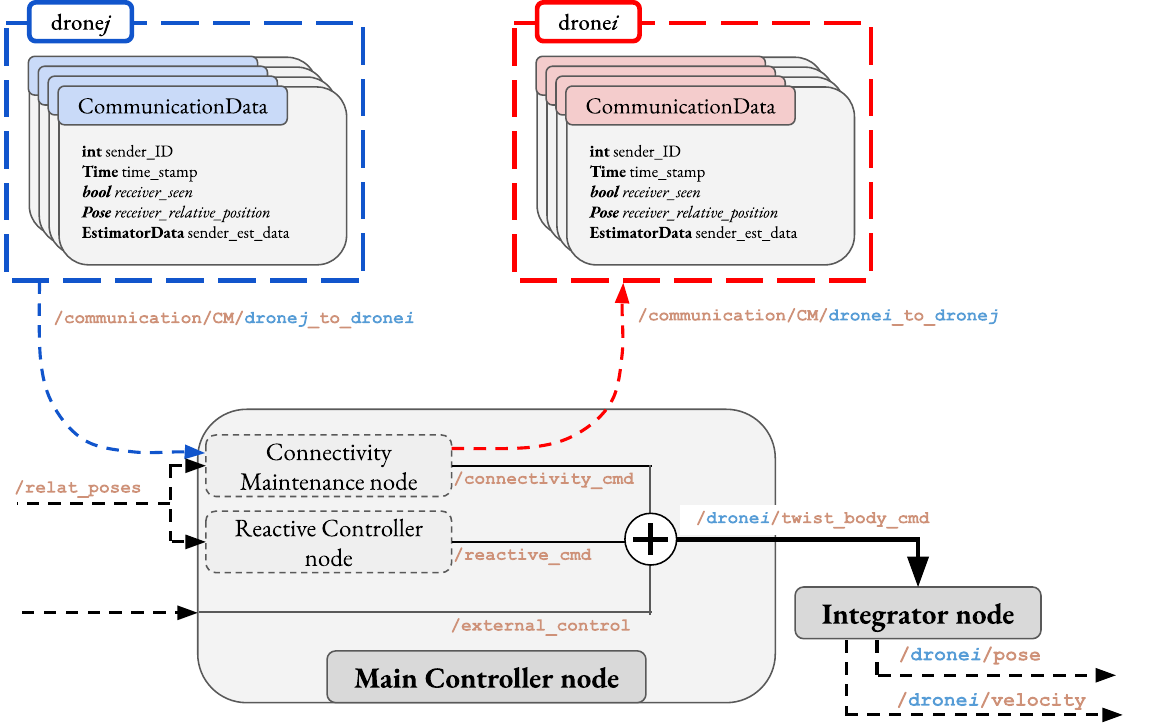
    \caption{\label{fig:onboardNodes} ROS~2 nodes running onboard each agent under the same namespace (drone$i$) and their input/output topics.}%
\end{figure}

The onboard software is responsible for controlling the drone. While all nodes run on the same machine in a simulated experiment, the onboard nodes are isolated under a dedicated ROS~2 namespace for each agent, effectively simulating their physical separation from the rest of the system. The main nodes composing the onboard software of a given agent $i$, as illustrated in Figure~\ref{fig:onboardNodes}, are:
\begin{itemize}
    \item[$\bullet$] \textbf{Integrator node}: In a simulated environment, drone dynamics are modeled using a double integrator with limited acceleration and velocity. This node receives velocity commands in the body frame of drone $i$ and outputs the corresponding velocity and pose of the agent.
    \item[$\bullet$] \textbf{Main controller node}: This node loads and executes high-level controllers, such as the connectivity maintenance controller and the reactive controller. It combines their outputs with external inputs to generate the final velocity command for the integrator node.
    \begin{itemize}
        \item \textbf{Connectivity maintenance controller}: Implements the decentralized connectivity maintenance algorithm presented in this thesis. It processes the relative positions of sensed neighbors along with communicated data (PI-ACE values and relative position measurements) to compute the connectivity maintenance force and generate the desired velocity command. Additionally, it transmits relevant data to sensed neighbors using peer-to-peer dedicated topics. This communication design presents a scalability limitation, as the number of topics increases exponentially with the number of agents. In a real-world experiment, inter-agent communication would be managed by a dedicated communication module, mitigating this limitation.
        \item \textbf{Reactive controller}: Implements a simple reactive avoidance algorithm. It receives the relative positions of neighboring agents and computes a repulsive force to prevent collisions.
    \end{itemize}
\end{itemize}

\section{Groundstation}

The ground station serves as the core of the human-multi-robot interaction system. Its primary component is the user control interface, which manages initialization, monitoring, and external control of the agents. Additionally, the ground station simulates onboard sensors, such as a relative position sensor, and provides force feedback to the user through a haptic device. Finally, the ground station handles scenario-specific tasks, including managing objectives, logging data, and debugging.

\subsection{User inputs}
Leveraging standardized ROS~2 topics, users can interact with the system through various input devices. Commands can be sent to the agents using a standard gamepad running the default \textit{joy node} provided by ROS~2 or via the \textit{Virtuose} from Haption, a 6-DOF haptic device.

\begin{figure}[ht]
    \centering
    \def\svgwidth{\linewidth}
    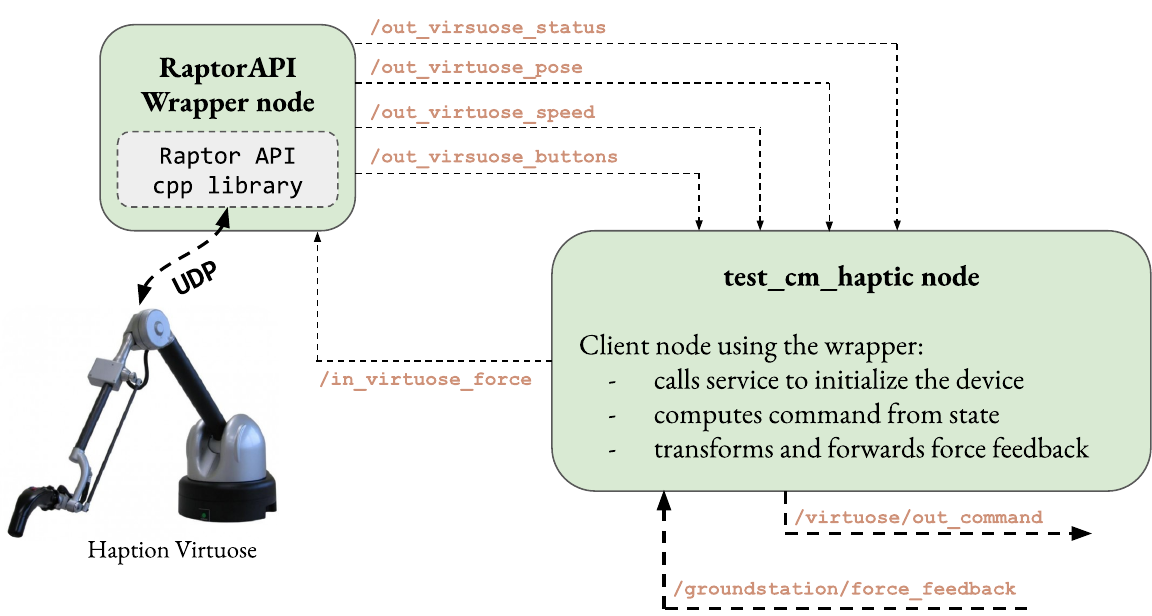
    \caption{\label{fig:raptorApi} Software implementation for the haptic interface.}%
\end{figure}

A custom ROS~2 node (\textit{test\_cm\_haptic}), implemented in \CC\ for high execution speed, outputs a \textit{sensor\_msgs/joy} message, mimicking a regular gamepad.
Unlike a gamepad, however, the haptic device requires force feedback from the ground station, which is transmitted via a \textit{geometry\_msgs/WrenchStamped} message to the\linebreak \textit{test\_cm\_haptic} node on the \textit{/groundstation/force\_feedback} topic.
This topic is published by a dedicated node (\textit{force\_feedback\_node}) that computes the force feedback based on the current control modality, while the \textit{test\_cm\_haptic} node handles the joystick-like forces. Both forces are detailed in Chapter~\ref{chap:chapitre5}.

To interface with the \textit{Virtuose} device, we utilize the \textit{Raptor API} from Haption, embedded within a dedicated wrapper node provided by Haption and slightly modified to fit our requirements. This wrapper node initializes the haptic device and serves as an interface for a client node (\textit{test\_cm\_haptic} in our case). It exposes several readable topics, such as the device’s position, velocity, and status, as well as a writable topic for desired force feedback.

Figure~\ref{fig:raptorApi} provides a graphical representation of the haptic device implementation within our system.

\subsection{User control interface}

The \textit{centralized\_controller\_node} serves as the primary interface between the user and the system. It enables the user to send commands to the agents, such as takeoff, landing, or switching high-level controllers. Additionally, it allows dynamic selection of the controlled agent and adjustment of the control modality (single-agent, pinning, or centroid).

\begin{figure}[ht]
    \centering
    \def\svgwidth{\linewidth}
    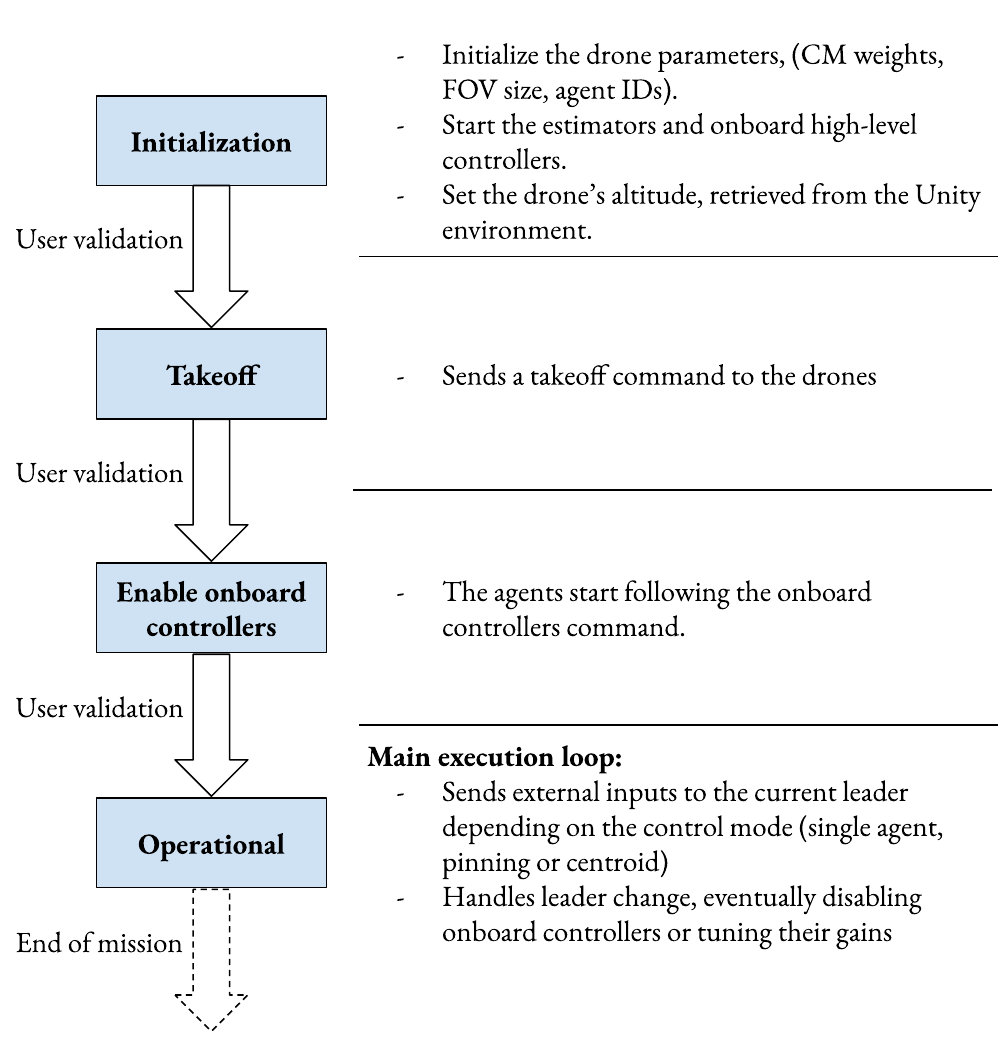
    \caption{\label{fig:stateMachine} State machine of the \textit{centralized\_controller\_node}.}%
\end{figure}

The node operates based on a state machine, illustrated in Figure~\ref{fig:stateMachine}, which transitions through various initialization states before entering the main execution loop.

To support the various control modalities presented in this thesis, the \linebreak\textit{centralized\_controller\_node} either directly interacts with a single agent, forwarding velocity commands from the user input system, or delegates control to a dedicated node responsible for implementing pinning or centroid control strategies. This design enables seamless switching between control modalities, as demonstrated in the video available at \url{https://youtu.be/gDR6Lky8_OI}.

\paragraph{Pinning control}
The \textit{pinning\_control\_node} emulates a virtual agent that is directly controlled by the user. Its state is sent to the Unity environment for visualization and is used to compute external commands for the pinned agents, as detailed in Chapter~\ref{chap:chapitre5}. Additionally, the node computes force feedback and publishes it to the \textit{/pinning\_control/feedback\_force} topic, which is accessed by the \textit{force\_feedback\_node}.

\paragraph{Centroid control}
The \textit{centroid\_control\_node} calculates the centroid of the agents group and transmits its position to the Unity environment for visualization. It also determines the external commands for the agents, as described in Chapter~\ref{chap:chapitre5}. The position of the centroid is not directly controller by the user but rather by the movement induced by the user onto the drones. The node is also responsible for computing the force feedback, publishing it onto a dedicated \textit{/centroid\_control/feedback\_force} topic, accessed by the \textit{force\_feedback\_node}.

\subsection{Sensor simulation}
A central node (\textit{relat\_pose\_node}) simulates an onboard sensor that provides relative position measurements of identified neighbors within a limited field of view (FOV). It computes the relative position of all neighboring agents in the body frame of the observing agent, considering FOV constraints, and publishes it to the \textit{/drone}i\textit{/relat\_poses} topic, which serves as input for the connectivity maintenance controller. Additionally, it supplies the reactive controller with relative position data of neighboring agents without FOV restrictions, effectively simulating an omnidirectional close-range distance sensor.

\subsection{Miscellaneous nodes}
Several optional nodes operate within the system, handling scenario-specific tasks, data logging, or debugging. These nodes include:
\begin{itemize}
    \item[$\bullet$] \textbf{Objective node}: This node manages objectives in the single-leader and pinning scenarios. It retrieves the list of objectives from the Unity environment and verifies whether the designated drone has reached the currently highlighted objective. If so, it either progresses to the next objective or concludes the mission.
    \item[$\bullet$] \textbf{Task timer node}: This node provides two services to start and stop a timer, enabling the measurement of mission completion time. It continuously publishes time data on a dedicated topic throughout the mission in order to save it in a ROS bag.
    \item[$\bullet$] \textbf{Debug node}: Initially designed to compare estimated $\lambda_2$ values with actual values, this node can also supply agents with the real centrally computed $\lambda_2$ value for improved stability and efficiency. Additionally, it transmits inter-agent weight values to the Unity environment for visualization.
    \item[$\bullet$] \textbf{Leader suggestion node}: Implementing the leader suggestion strategy described in Chapter~\ref{chap:chapitre5}, this node provides a service to recommend the most suitable leader from a given subset of agents.
    \item[$\bullet$] \textbf{Data grabber node}: Previously used for logging experimental data into CSV files for offline analysis, this node has been deprecated. Data logging is now managed via a ROS bag recorded from a separate computer connected to the same network. This physical separation ensures that the logging process does not interfere with the system's real-time performance.
\end{itemize}

\section{Graphical interface}
This section provides a brief overview of the functions implemented in the Unity environment, which serves as the graphical interface for the user. For a more comprehensive understanding of the system, we strongly encourage the reader to refer to the full video available at \url{https://youtu.be/gDR6Lky8_OI}, as presented in Chapter~\ref{chap:chapitre5}.

The Unity-based visual interface offers a user-friendly menu for selecting various display options, including:
\begin{itemize}
    \item Visualization of topological links between agents, color-coded according to their connectivity weights.
    \item Selection of the observer camera type (onboard the current leader or external 3rd person view).
    \item Display or concealment of onboard camera views.
    \item Choice of environment (desert or plains)...
\end{itemize}

The Unity environment interfaces with the ROS~2 system via TCP, utilizing the {\textit{default\_server\_endpoint}} node from the \textit{ROS TCP Endpoint} package \parencite{unity2022rostcpendpoint}. This TCP connection enables the Unity visualization to run on a separate device from the ROS~2 system, reducing the load on a single machine and allowing for the use of a Windows-based platform, which facilitates integration with a Virtual Reality headset.
The Unity environment subscribes to multiple topics such as robot positions, inter-agent connectivity weights, and the positions of the virtual pinning agent or centroid (if applicable), along with other scenario-specific data. Additionally, it implements and calls ROS~2 services while monitoring environment-specific metrics depending on the scenario.

For instance:
\begin{itemize}
    \item \textbf{Single-leader or pinning scenarios}: The Unity environment provides a service called by the \textit{objective\_node}, which supplies the list and types of environment-specific objectives. It is also used to set which objective is currently highlighted for the user, using a colored cylinder and a moving arrow, as shown in the videos.
    \item \textbf{Centroid control scenario}:  The Unity environment detects when an agent enters a restricted (red) area and reports it to the ROS~2 system. It also manages the progressive discovery of new areas by dynamically removing the fog of war. At the mission's conclusion, the Unity environment saves the discovered regions as a bitmap image.
\end{itemize}

Finally, the Unity environment provides the ROS system with the altitude of the drones in the simulated environment through a service call, allowing the system to adjust the drones' starting altitude and avoiding setting a virtual limit to the drones' vertical movement.

\section{Launch files}

This project uses the \textit{simple launch} package available on Github at \url{https://github.com/oKermorgant/simple_launch}. This package provides a Python class to help writing ROS 2 launch files.
The way ROS~2 launch files are handled allows for an easily configurable system, allowing to change parameters and arguments without modifying the code nor compiling the project. This feature is particularly useful for the system presented in this thesis, as it allows for quick changes in the system's configuration, such as the number of agents, the control modality, or the simulation environment.

Here is an example of a launch file for the system presented in this thesis:
\begin{codebox}
\input{Chapitre98/launch_file}
\end{codebox}

This launch file is used to start the system with a given number of agents, a specific leader, and various configuration options. It launches the ground station and all the agents, passing the necessary arguments to each launch file. The launch files for the ground station and the agents are written in a similar fashion, allowing for easy configuration of the system.

\medskip

This software is part of a project named \textit{Olive}, which aims to facilitate the deployment of multi-agent systems in both simulated and real environments while remaining agnostic to the underlying hardware. Various high-level controllers, such as formation control and flocking, have been explored, and integration with our motion capture system has been achieved, though not presented in this thesis.

\clearemptydoublepage
\phantomsection
\addcontentsline{toc}{chapter}{Bibliography}

\printbibliography[]

\clearemptydoublepage
\cleartoevenpage[\thispagestyle{empty}]
\markboth{}{}
\newgeometry{inner=30mm,outer=20mm,top=40mm,bottom=20mm}

\backcoverheader

\selectfontbackcover{ %

\titleFR{Contrôle Partagé avec Assistance Haptique pour Systèmes Multi-Robots}

\keywordsFR{Systèmes multi-agents, Contrôle partagé, Retour haptique}

\abstractFR{Les systèmes multi-agents offrent des avantages en termes de flexibilité et de redondance, permettant la répartition efficace des tâches complexes entre plusieurs robots.
Cependant, leur déploiement exige des stratégies avancées de contrôle afin d'assurer une coopération optimale entre agents autonomes et opérateurs humains.
Nous proposons une modification de l'algorithme de maintenance de connectivité globale afin d'intégrer des mesures réalisées dans le repère propre de chaque agent, avec un capteur à champ de vision limité.
Cet algorithme, assurant le contrôle autonome du système multi-robots, est combiné à différentes stratégies de supervision adaptative, offrant à l'utilisateur un contrôle à échelle variable sur l'ensemble du groupe:
\textit{(i)} Contrôle d'agent unique (S) : l'opérateur pilote un robot spécifique,
\textit{(ii)} Contrôle par ancrage (P) : l'opérateur impose des contraintes à plusieurs robots pour influencer le groupe,
\textit{(iii)} Contrôle par barycentre (C) : l'opérateur agit sur le centre de masse du système multi-robots pour guider la flotte.
Associées à l'algorithme de maintenance de connectivité, ces méthodes garantissent le respect des contraintes inter-agents tout en fournissant un retour haptique à l'utilisateur sur l'état de connectivité du groupe. L'efficacité de ces approches a été évaluée dans une étude utilisateur comparant les performances avec et sans retour haptique à travers plusieurs scénarios adaptés aux différents modes de contrôle.}

\titleEN{Haptic Shared Control for Multi-Robot Systems}

\keywordsEN{Multi-agent systems, Shared control, Haptic feedback}

\abstractEN{Multi-agent systems offer significant advantages in terms of flexibility and redundancy, enabling the efficient distribution of complex tasks among multiple robots.
However, their deployment requires advanced control strategies to ensure optimal cooperation between autonomous agents and human operators.
We propose a modification of the global connectivity maintenance algorithm to incorporate measurements taken in agent's local frame, using sensors with limited fields of view.
This algorithm, responsible for the autonomous behavior of the multi-robot system, is combined with different adaptive supervision strategies, allowing the user to exert control at varying scales over the entire group.
\textit{(i)} Single Agent Control (S): The operator directly controls a single robot,
\textit{(ii)} Pinning Control (P): The operator imposes constraints on multiple robots to influence the group,
\textit{(iii)} Centroid Control (C): The operator acts on the system's center of mass to steer the entire fleet.
In conjunction with the connectivity maintenance algorithm, these methods ensure compliance with inter-agent constraints while providing haptic feedback to the user regarding the system's connectivity state. The effectiveness of these approaches was evaluated through a user study, comparing performance with and without haptic feedback across multiple scenarios tailored to the different control modes.}

}

\restoregeometry

\end{document}